\documentclass[11pt]{article}

\usepackage{iclr2026_conference,times}

\usepackage[T1]{fontenc}
\usepackage[utf8]{inputenc}
\usepackage{microtype}
\usepackage{inconsolata}
\usepackage{graphicx}
\usepackage[table]{xcolor}
\usepackage{booktabs}
\usepackage{framed}
\usepackage{multirow}
\usepackage{array}
\usepackage{tabularx}
\usepackage{longtable}
\usepackage{amsmath}
\usepackage{amssymb}
\usepackage{enumitem}
\usepackage{placeins}
\usepackage{float}
\usepackage{fancyvrb}
\usepackage{algorithm}
\usepackage{algpseudocode}
\usepackage{tikz}
\usepackage{hyperref}
\usepackage{url}
\hypersetup{hidelinks}
\algrenewcommand\algorithmiccomment[1]{\hfill \(\triangleright\) #1}


\newcommand{\method}{MAPLE}
\newcommand{\methodours}{\method{} (ours)}

\newcommand{\methodlogo}{\includegraphics[height=2.85em]{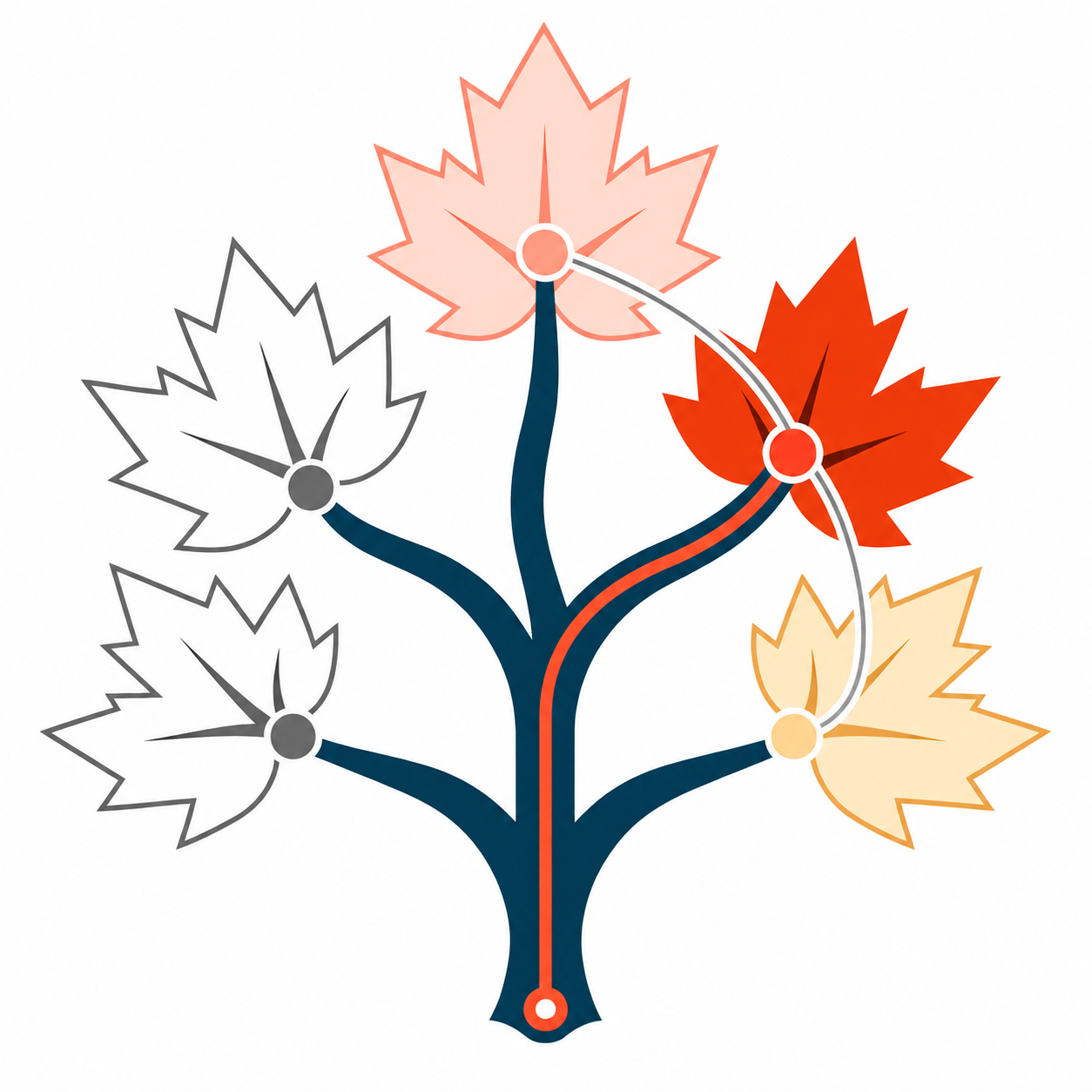}}
\newcommand{\githubicon}{\raisebox{-0.16em}{\includegraphics[height=1.05em]{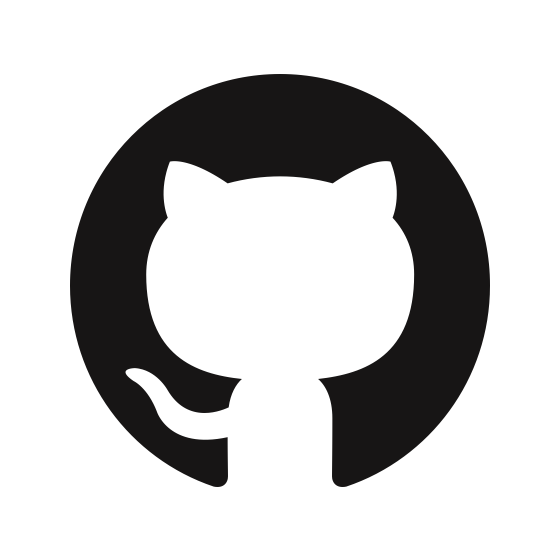}}}
\newcommand{\demoicon}{\raisebox{-0.22em}{\includegraphics[height=1.30em]{figures/maple-logo.png}}}
\newcommand{\MAPLEResearchURL}{https://github.com/xin8coder/MAPLE}
\newcommand{\MAPLEHarnessURL}{https://github.com/xin8coder/MAPLE-harness-core}
\newcommand{\MAPLEProjectURL}{https://xin8coder.github.io/MAPLE-harness/}
\newcommand{\methodtitleblock}{%
		\resizebox{0.94\textwidth}{!}{%
		\begin{tabular}{@{}c@{\hspace{0.55em}}l@{}}
			\raisebox{-0.42\height}{\methodlogo} &
			\begin{tabular}[c]{@{}l@{}}
				MAPLE: Memory-Augmented Planning\\[-0.04em]
				with Language and Evolution
			\end{tabular}\\[0.18em]
			\multicolumn{2}{c}{%
				\normalfont\scriptsize
				\githubicon\enspace\textbf{Research}\enspace
				\href{\MAPLEResearchURL}{\nolinkurl{https://github.com/xin8coder/MAPLE}}\hphantom{\nolinkurl{MM}}
				\hspace{1.2em}
				\githubicon\enspace\textbf{Application}\enspace
				\href{\MAPLEHarnessURL}{\nolinkurl{https://github.com/xin8coder/MAPLE-harness-core}}\hphantom{\nolinkurl{MM}}%
			}\\[-0.02em]
			\multicolumn{2}{c}{%
				\normalfont\scriptsize
				\demoicon\enspace\textbf{Live demo}\enspace
				\href{\MAPLEProjectURL}{\nolinkurl{https://xin8coder.github.io/MAPLE-harness/}}%
			}
		\end{tabular}}%
}

\definecolor{LiveOptNavy}{HTML}{334B60}
\definecolor{LiveOptTeal}{HTML}{39857D}
\definecolor{LiveOptSky}{HTML}{A7C8DC}
\definecolor{LiveOptPale}{HTML}{F0F6F3}
\definecolor{LiveOptRowAccent}{HTML}{E6EFEB}
\definecolor{LiveOptGold}{HTML}{947443}
\definecolor{LiveOptPurple}{HTML}{6F6484}
\definecolor{LiveOptGray}{HTML}{666666}
\definecolor{LiveOptStaticRow}{HTML}{F0F3F6}
\definecolor{LiveOptDynamicRow}{HTML}{F0F5F2}
\definecolor{LiveOptParetoRow}{HTML}{F3F1F6}
\definecolor{LiveOptHardRow}{HTML}{F6F2EE}
\definecolor{LiveOptBaselineRowA}{HTML}{FFFFFF}
\definecolor{LiveOptBaselineRowB}{HTML}{F4F6F7}
\definecolor{LiveOptTableHeader}{HTML}{E8EDF0}
\definecolor{LiveOptTableRule}{HTML}{75838C}
\definecolor{LiveOptTableLight}{HTML}{F8FAFC}
\definecolor{LiveOptTablePending}{HTML}{7A7A7A}
\definecolor{LiveOptTypeSummary}{HTML}{E6EFEB}
\definecolor{LiveOptTimeSummary}{HTML}{F3EEDF}
\newcommand{\MainTextTableSetup}{%
	\small
	\setlength{\tabcolsep}{4.5pt}%
\renewcommand{\arraystretch}{1.25}%
	\arrayrulecolor{LiveOptTableRule}%
}
\newcommand{\MainResultTableSetup}{%
	\footnotesize
	\setlength{\tabcolsep}{4.0pt}%
\renewcommand{\arraystretch}{1.25}%
	\arrayrulecolor{LiveOptTableRule}%
}
\newcommand{\ResultTableSetup}{%
	\footnotesize
	\setlength{\tabcolsep}{3.5pt}%
\renewcommand{\arraystretch}{1.22}%
	\arrayrulecolor{LiveOptTableRule}%
}
\newcommand{\DenseResultTableSetup}{%
	\scriptsize
	\setlength{\tabcolsep}{3.0pt}%
\renewcommand{\arraystretch}{1.20}%
	\arrayrulecolor{LiveOptTableRule}%
}
\newcommand{\TableHead}{\rowcolor{LiveOptTableHeader}}

\newcolumntype{C}[1]{>{\centering\arraybackslash}p{#1}}
\newcolumntype{L}[1]{>{\raggedright\arraybackslash}p{#1}}
\newcolumntype{Y}{>{\centering\arraybackslash}X}
\newcolumntype{Z}{>{\raggedright\arraybackslash}X}
\newcommand{\TYes}{\textcolor{LiveOptTeal}{\bfseries Yes}}
\newcommand{\TPart}{\textcolor{LiveOptGold}{\bfseries Part.}}

\newcommand{\TX}{\textcolor{LiveOptGray}{No}}
\newcommand{\Adapted}{\textsuperscript{\textcolor{LiveOptTablePending}{\scriptsize *}}}

\newcommand{\TightStack}[1]{%
	\begingroup
	\renewcommand{\arraystretch}{0.82}%
	\begin{tabular}[c]{@{}c@{}}#1\end{tabular}%
	\endgroup
}

\newsavebox{\LiveOptPairLeftBox}
\newsavebox{\LiveOptPairRightBox}
\newlength{\LiveOptPairHeight}
\newcommand{\LiveOptEqualHeightPair}[4]{%
	\begingroup
	\sbox{\LiveOptPairLeftBox}{\begin{minipage}[t]{#1}\vspace{0pt}#3\end{minipage}}%
	\sbox{\LiveOptPairRightBox}{\begin{minipage}[t]{#2}\vspace{0pt}#4\end{minipage}}%
	\setlength{\LiveOptPairHeight}{\dimexpr\ht\LiveOptPairLeftBox+\dp\LiveOptPairLeftBox\relax}%
	\ifdim\dimexpr\ht\LiveOptPairRightBox+\dp\LiveOptPairRightBox\relax>\LiveOptPairHeight
		\setlength{\LiveOptPairHeight}{\dimexpr\ht\LiveOptPairRightBox+\dp\LiveOptPairRightBox\relax}%
	\fi
	\noindent
	\begin{minipage}[t][\LiveOptPairHeight][t]{#1}\vspace{0pt}\usebox{\LiveOptPairLeftBox}\end{minipage}%
	\hfill
	\begin{minipage}[t][\LiveOptPairHeight][s]{#2}\vspace{0pt}#4\par\end{minipage}%
	\endgroup
}
\newsavebox{\LiveOptLeftCaptionBox}
\newsavebox{\LiveOptRightCaptionBox}
\newlength{\MapleCaptionedHeight}
\newcommand{\LiveOptCaptionedPair}[6]{%
  \begingroup
  \sbox{\LiveOptPairLeftBox}{\begin{minipage}[t]{#1}\vspace{0pt}#3\end{minipage}}%
  \sbox{\LiveOptPairRightBox}{\begin{minipage}[t]{#2}\vspace{0pt}#5\end{minipage}}%
  \sbox{\LiveOptLeftCaptionBox}{\begin{minipage}[b]{#1}\raggedright\small#4\par\end{minipage}}%
  \sbox{\LiveOptRightCaptionBox}{\begin{minipage}[b]{#2}\raggedright\small#6\par\end{minipage}}%
  \setlength{\MapleCaptionedHeight}{\dimexpr\ht\LiveOptPairLeftBox+\dp\LiveOptPairLeftBox+\ht\LiveOptLeftCaptionBox+\dp\LiveOptLeftCaptionBox+3pt\relax}%
  \ifdim\dimexpr\ht\LiveOptPairRightBox+\dp\LiveOptPairRightBox+\ht\LiveOptRightCaptionBox+\dp\LiveOptRightCaptionBox+3pt\relax>\MapleCaptionedHeight
    \setlength{\MapleCaptionedHeight}{\dimexpr\ht\LiveOptPairRightBox+\dp\LiveOptPairRightBox+\ht\LiveOptRightCaptionBox+\dp\LiveOptRightCaptionBox+3pt\relax}%
  \fi
  \noindent\begin{minipage}[t][\MapleCaptionedHeight][s]{#1}\vspace{0pt}%
    \usebox{\LiveOptPairLeftBox}\par\vspace{3pt}\vfill\usebox{\LiveOptLeftCaptionBox}%
  \end{minipage}\hfill
  \begin{minipage}[t][\MapleCaptionedHeight][s]{#2}\vspace{0pt}%
    \usebox{\LiveOptPairRightBox}\par\vspace{3pt}\vfill\usebox{\LiveOptRightCaptionBox}%
  \end{minipage}%
  \endgroup
}
\newcommand{\MapleDiagram}[3]{%
  \noindent\begin{minipage}{\linewidth}
    \centering\includegraphics[width=\linewidth,keepaspectratio]{#1}\par
    \caption{#2}\label{#3}%
  \end{minipage}%
}

\newcommand{\NLDOEpisodeHead}[1]{%
	\par\addvspace{0.85em}%
	\noindent{\small\bfseries\scshape #1}%
	\par\nobreak\vspace{1.5pt}%
	\noindent{\color{LiveOptTeal}\rule{\linewidth}{0.5pt}}%
	\par\nobreak\vspace{0.10em}%
}
\newcommand{\NLDOUpdateTag}[1]{%
	\begingroup\setlength{\fboxsep}{1.4pt}%
	\colorbox{LiveOptPale}{\textcolor{LiveOptTeal!75!black}{\scriptsize #1}}%
	\endgroup%
}
\newlist{NLDOUpdateList}{itemize}{1}
\setlist[NLDOUpdateList]{leftmargin=18.2mm,labelwidth=18.2mm,labelsep=0.6mm,align=left,itemsep=0.15pt,topsep=0.8pt,parsep=0pt,partopsep=0pt}
\newcommand{\NLDOUpdateLabel}[2]{\texttt{#1} \NLDOUpdateTag{#2}}

    \title{\texorpdfstring{\methodtitleblock}{MAPLE: Memory-Augmented Planning with Language and Evolution}}

\author{%
Kesheng Chen \quad
Yamin Hu \quad
Wenjian Luo\\
Harbin Institute of Technology, Shenzhen, China%
}
	\renewenvironment{abstract}
		{\vskip.075in\centerline{\large\sc Abstract}\vspace{0.5ex}\par\noindent\ignorespaces}
		{\par\vskip 1ex}
    \iclrfinalcopy
\newcommand{\LiveOptPublicDMHV}{0.875}
\newcommand{\PersistentPublicDMHV}{0.042}
\newcommand{\LiveOptPublicSMHV}{0.832}
\newcommand{\PersistentPublicSMHV}{0.259}

\newcommand{\WarmPublicDMHV}{0.866}
\newcommand{\FullPublicDMHV}{0.825}
\newcommand{\LiveOptPublicUpdateHV}{0.879}
\newcommand{\WarmPublicUpdateHV}{0.868}
\newcommand{\FullPublicUpdateHV}{0.824}
\newcommand{\NoTSSPublicUpdateHV}{0.665}
\newcommand{\NoTSSPublicConditionalHV}{0.826}

\newcommand{\PublicWarmDelta}{-0.010}

\newcommand{\PublicFullDelta}{-0.054}
\newcommand{\PersistentPublicUpdateHV}{0.024}
\newcommand{\ReactPublicSMHV}{0.300}
\newcommand{\ReactPublicUpdateHV}{0.000}
\newcommand{\OptimusPublicSMHV}{0.417}
\newcommand{\OptimusPublicUpdateHV}{0.056}
\newcommand{\ORLMPublicSMHV}{0.000}
\newcommand{\ORLMPublicUpdateHV}{0.000}
\newcommand{\ORAgentPublicSMHV}{0.440}
\newcommand{\ORAgentPublicUpdateHV}{0.013}
\newcommand{\OptimAIPublicSMHV}{0.377}
\newcommand{\OptimAIPublicUpdateHV}{0.024}

\begin{document}

		\vspace*{-0.20in}

	\maketitle

		\lhead{Preprint}
		\vspace{-0.29in}

		\begin{abstract}
Domain practitioners understand their business constraints but may lack operations-research expertise or dedicated support. LLM-based optimization agents translate natural-language requirements into models or solver programs that established optimization tools can execute. This progress makes optimization more accessible, but real-world operations are dynamic: changing demand, resources, and priorities require updates to data, constraints, and objectives. Methods centered on isolated requests offer limited support for rapid adaptation that preserves earlier decisions and reuses useful search results. We introduce MAPLE (Memory-Augmented Planning with Language and Evolution), an agent for maintaining optimization problems through successive natural-language requests. MAPLE combines language-based problem construction with mathematical programming and evolutionary search. It retains the optimization program, accepted plans, earlier updates, and candidate solutions for subsequent requests. We introduce NLDO, a benchmark of 15 trajectories and 180 updates spanning selection, scheduling, rostering, routing, and cloud-resource placement. In the main evaluation, MAPLE completes all trajectories and achieves online scalar quality of 0.951 and a Pareto hypervolume ratio of 0.875. Controlled comparisons further show that maintaining executable state improves update validity and can preserve useful search information across substantial revisions.

\end{abstract}

		\begin{figure}[!htbp]
			\centering
			\hfuzz=1.5pt
\includegraphics[width=\linewidth]{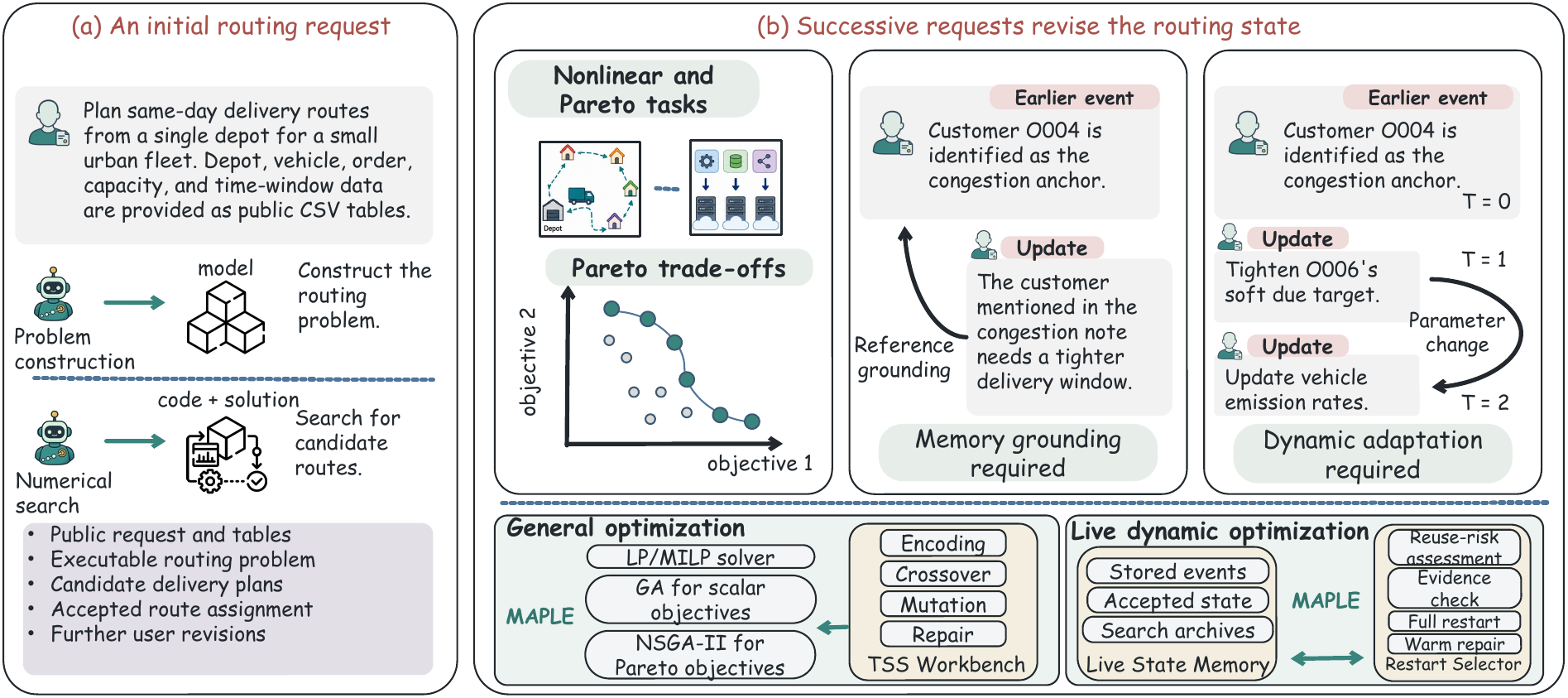}
				\vspace{0.12cm}
			\begin{minipage}[b]{0.440\linewidth}
				\vspace{0pt}
				\centering
				\begingroup
					\fontsize{5.9pt}{7.3pt}\selectfont
					\setlength{\tabcolsep}{0.15pt}
				\renewcommand{\arraystretch}{1.22}
\textbf{Native capabilities and evaluated adaptations}
				\vspace{0.06em}
\begin{tabularx}{0.995\linewidth}{@{}l*{5}{Y}@{}}
					\toprule
					\TableHead
					\TightStack{\textbf{Method}} &
					\TightStack{\textbf{Natural}\\\textbf{language}} &
					\TightStack{\textbf{Solver}\\\textbf{output}} &
					\TightStack{\textbf{Retained}\\\textbf{Pareto set}} &
					\TightStack{\textbf{Live}\\\textbf{updates}} &
						\TightStack{\textbf{Solver}\\\textbf{state reuse}} \\
					\midrule
					\rowcolor{LiveOptBaselineRowA}
					ReAct & \TYes & \TYes & \TX & \TPart\Adapted & \TX \\
					\rowcolor{LiveOptBaselineRowB}
OR-LLM-Agent & \TYes & \TYes & \TX & \TPart\Adapted & \TX \\
					\rowcolor{LiveOptBaselineRowA}
					ORLM & \TYes & \TYes & \TX & \TPart\Adapted & \TX \\
					\rowcolor{LiveOptBaselineRowB}
					OptiMUS & \TYes & \TYes & \TX & \TX & \TX \\
					\rowcolor{LiveOptBaselineRowA}
					OptimAI & \TYes & \TYes & \TX & \TX & \TX \\
					\rowcolor{LiveOptBaselineRowB}
Persistent ReAct & \TYes & \TYes & \TX & \TYes\Adapted & \TPart\Adapted \\
					\rowcolor{LiveOptRowAccent}
					\textbf{\method{}} (ours) & \TYes & \TYes & \TYes & \TYes & \TYes \\
					\bottomrule
				\end{tabularx}
				\endgroup
			\end{minipage}\hfill
			\begin{minipage}[b]{0.552\linewidth}
				\vspace{0pt}
				\centering
				\begingroup
					\fontsize{5.9pt}{7.3pt}\selectfont
					\setlength{\tabcolsep}{0.15pt}
				\renewcommand{\arraystretch}{1.22}
				\textbf{Primary results by static/dynamic setting}
				\vspace{0.06em}
				\let\OriginalTableAdapted\Adapted
				\renewcommand{\Adapted}{\rlap{\OriginalTableAdapted}}
				\begin{tabularx}{0.990\linewidth}{@{}l*{6}{Y}@{}}
					\toprule
					\TableHead
					\textbf{Method} & \TightStack{\textbf{NLP4LP-}\\\textbf{hard}} & \textbf{BWOR20} & \TightStack{\textbf{NLDO}\\\textbf{SS}} & \TightStack{\textbf{NLDO}\\\textbf{DS}} & \TightStack{\textbf{NLDO}\\\textbf{SM}} & \TightStack{\textbf{NLDO}\\\textbf{DM}} \\
					\midrule
					\rowcolor{LiveOptBaselineRowA}
					ReAct & 57.6\% & 70.0\% & 0.889\Adapted & 0.333\Adapted & \ReactPublicSMHV\Adapted & 0.023\Adapted \\
					\rowcolor{LiveOptBaselineRowB}
					OR-LLM-Agent & 50.8\% & 65.0\% & 0.778\Adapted & 0.111\Adapted & \ORAgentPublicSMHV\Adapted & 0.046\Adapted \\
					\rowcolor{LiveOptBaselineRowA}
					ORLM & 23.7\% & 45.0\% & \textbf{1.000}\Adapted & 0.120\Adapted & \ORLMPublicSMHV\Adapted & 0.000\Adapted \\
					\rowcolor{LiveOptBaselineRowB}
					OptiMUS & \textbf{69.5}\% & 65.0\% & 0.889\Adapted & 0.281\Adapted & \OptimusPublicSMHV\Adapted & 0.084\Adapted \\
					\rowcolor{LiveOptBaselineRowA}
					OptimAI & 61.0\% & 55.0\% & 0.556\Adapted & 0.077\Adapted & \OptimAIPublicSMHV\Adapted & 0.051\Adapted \\
					\rowcolor{LiveOptBaselineRowB}
					Persistent ReAct & 62.7\% & 70.0\% & 0.889\Adapted & 0.501\Adapted & \PersistentPublicSMHV\Adapted & \PersistentPublicDMHV\Adapted \\
					\rowcolor{LiveOptRowAccent}
					\textbf{\method{}} (ours) & 59.3\% & \textbf{80.0}\% & \textbf{1.000} & \textbf{0.951} & \textbf{\LiveOptPublicSMHV} & \textbf{\LiveOptPublicDMHV} \\
					\bottomrule
					\end{tabularx}
					\endgroup
			\end{minipage}
\caption{\textbf{MAPLE maintains plans through successive requests.} The routing example shows problem revision, memory of earlier decisions, and reuse of search results. The tables compare capabilities and performance. An asterisk marks adapted baselines.}
				\label{fig:first-page-overview}
				\vspace{-0.45em}
			\end{figure}

				\vspace{-0.45em}

		\section{Introduction}

Domain practitioners understand operational requirements but may lack operations research training or specialist support. Production planners, dispatchers, workforce coordinators, and infrastructure operators must translate these requirements into optimization models and maintain the software that solves them. This expertise barrier limits everyday adoption of optimization methods \citep{ahmadi2024optimus,zhang2025orllmagent}. LLM-based optimization agents translate natural-language descriptions into models or solver code and execute them with established optimization packages. Some systems use execution errors to revise the generated model or program \citep{ahmadi2024optimus,huang2025orlm,thind2025optimai,zhang2025orllmagent}. More recent systems ask users to clarify the planning problem and build optimization models from business data \citep{xie2026orpilot}. These systems connect domain practitioners to optimization tools.

Real-world operations are dynamic: new orders arrive, resources become unavailable, and priorities change. Model generation alone cannot ensure rapid, consistent adaptation. Keeping a previously discussed customer's vehicle assignment requires the agent to revise the program, recover that decision, and assess whether earlier candidates remain useful.

Related work addresses parts of this setting from different directions. Multi-turn planning benchmarks study how plans change as new constraints arrive \citep{oh2025flextravelplanner}. Other LLM-based optimization methods address nonlinear problems and multi-objective search \citep{thind2025optimai,schwanke2026mohollm}. Dynamic optimization, meanwhile, has long studied population repair, transfer, and restart under changing objectives and constraints \citep{branke2002dynamic,nguyen2012dynamic}. We address a practical gap in real-world operations: adapting executable optimization programs to evolving natural-language requirements while preserving accepted decisions and reusing relevant search state. We call this setting live dynamic optimization.

MAPLE maintains an executable optimization program (the Workbench), input data, accepted plans, update records, and candidate solutions. At initialization, Typed Search-Space Scaffolding (TSS) lets the model define decisions and evaluation, while the system supplies search operations and solvers. It supports mathematical programming for linear formulations, evolutionary search for nonlinear or combinatorial objectives, and Pareto search for competing objectives. For subsequent requests, Live Problem Decomposition (LPD) identifies the input data or Workbench functions affected by the revision. Live State Memory (LSM) retrieves information from earlier updates and accepted plans. When a request asks to keep an earlier assignment, the system enforces that assignment during the next search. For evolutionary search, a restart selector reads a summary of the changes and assesses whether earlier candidates remain useful. Fixed checks then select repaired-history initialization (Warm) or fresh initialization (Full). The revised state is committed only after executable and output checks pass.

We evaluate MAPLE on NLDO, a controlled benchmark containing 15 trajectories and 180 natural-language updates across selection, scheduling, rostering, routing, and cloud-resource placement. NLDO provides both single- and multi-objective tasks and evaluates trajectory continuity separately from the quality of the resulting plans. In the main evaluation, MAPLE completes all trajectories, reaching online scalar quality of 0.951 and Pareto hypervolume ratio of 0.875. The component and restart controls further examine how executable-state maintenance affects validity, historical grounding, and reuse across revisions.

Our contributions are threefold. First, we formulate live dynamic optimization as optimization under cumulative natural-language revisions and introduce NLDO, with static and dynamic views of single- and multi-objective tasks. Second, we develop MAPLE, an executable-state architecture that integrates problem construction, accepted decisions, historical grounding, and reusable numerical search state. Third, we evaluate complete agent workflows and controlled variants, showing that persistent executable state improves continuity and validity while allowing useful search information to survive substantial revisions.

		\linespread{1.00}\selectfont
		\section{Task and Benchmark}
\label{sec:task}
\label{sec:benchmark}

\noindent\textbf{From requests to optimization problems.}\quad
NLDO evaluates a sequence of planning requests. Each episode begins with a request and input tables $(x_0,D_0)$, followed by revisions $u_{1:T}$.
Together they define, at time $t$, a feasible set $\mathcal{F}_t$ and objective vector $\mathbf{f}_t=(f_{t,1},\ldots,f_{t,k_t})$.
For $k_t=1$, the desired output is one feasible plan with a small objective value.
For $k_t>1$, it is a set of feasible, mutually nondominated plans that represents the trade-offs.
Objectives are converted to minimization for evaluation.
The public request distinguishes optimization objectives, hard constraints, and auxiliary diagnostics.
Every method receives the initial request, public tables, cumulative public updates, required outputs, and public validation feedback.
Each method retains its own dialogue and state through a separate execution history.
Section~\ref{sec:experiments} specifies the different history interfaces used in the comparison.
MAPLE stores the input tables $D_t$, optimization program $W_t$, accepted output $A_t$, and candidate solutions $P_t$. These form the state $S_t=(D_t,W_t,A_t,P_t)$. It also retains the earlier update records $E_t$.
For $t\geq1$, the update procedure receives the current request, preceding optimization state, and event history:
\[
S_t=\Phi(x_0,u_{1:t},S_{t-1},E_{t-1}).
\]
An accepted update appends its event record to $E_{t-1}$.
A rejected revision leaves the preceding committed state and event history unchanged.
For Pareto tasks, $A_t$ stores alternative plans and the representative used for historical assignments (Section~\ref{sec:method}).

\paragraph{Continuity and quality.}
We evaluate recorded protocol continuity and saved-plan quality separately.
Let $a_t$ record whether the original output passes both its output-protocol checks and its recorded validity check.
The strict trajectory protocol keeps the prefix $r_t=\prod_{j=0}^{t}a_j$.
Let $v_t$ indicate whether the saved output contains a feasible plan under the evaluation formulas, and let $q_t$ measure its quality, with $q_t=0$ when $v_t=0$.
We report
\[
  \mathrm{SolveRate}=\frac{1}{T+1}\sum_{t=0}^{T}r_tv_t,
  \qquad
  \mathrm{OnlineQuality}=\frac{1}{T+1}\sum_{t=0}^{T}r_tq_t.
\]
Here $q_t$ is normalized scalar objective quality or a hypervolume (HV) ratio for Pareto outputs.
Both summaries include the initial state.
The first recorded rejection ends the eligible prefix, and all remaining states receive zero, including unattempted states.
We separately report output-protocol failures, infeasible plans, and feasibility among attempted updates.
Appendix~\ref{app:metric-definitions} defines normalization, IGD, and reference construction.

\paragraph{NLDO coverage.}
NLDO has five families with three sequences each: 15 trajectories, 195 states, and 180 updates. Each sequence contains an initial request and twelve updates.
\textbf{NLDO-SS} and \textbf{NLDO-SM} evaluate the single- and multi-objective initial states.
\textbf{NLDO-DS} and \textbf{NLDO-DM} evaluate complete trajectories from t00 to t12, including initialization.
Update-only analyses explicitly use t01--t12.
Numerical seeds repeat the search within each model-generated trajectory.
The main episodes retain their initial table schemas.
Rostering, routing, and cloud placement include disruptive replacements at t11--t12.
Some updates change only diagnostics or positively rescale an objective, preserving Pareto dominance. Here, public data means input data available to the agent.
Separate controls test historical bindings and candidate reuse beyond the main trajectories (Appendix~\ref{app:component-details}).

Appendix~\ref{app:benchmark-details} explains how updates affect feasibility, objective ordering, rescaling, and diagnostics.
		\section{Method}
\label{sec:method}

\begin{figure}[!t]
\centering
\includegraphics[width=0.99\linewidth]{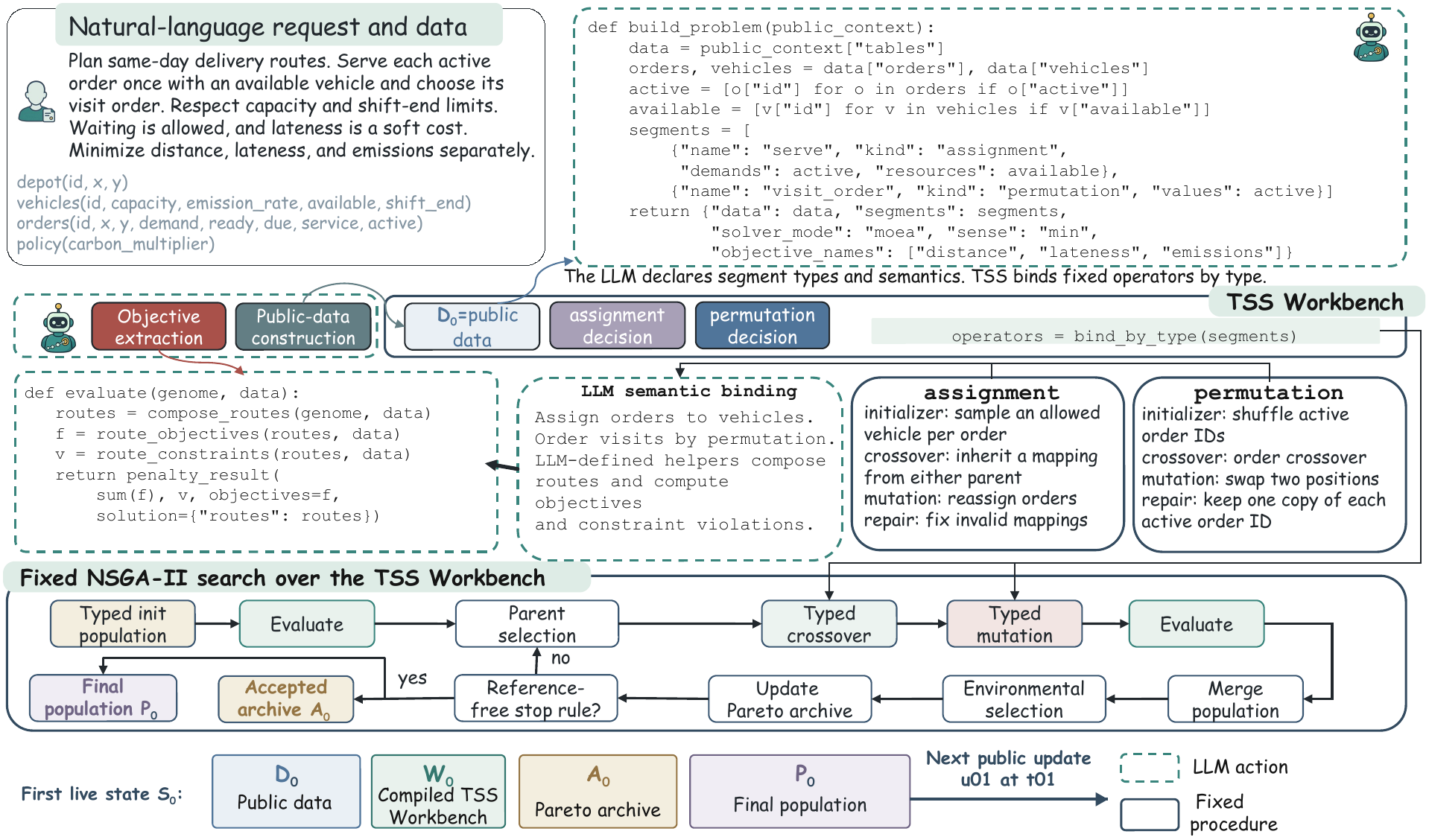}
\caption{\textbf{Constructing the initial state.} The model builds $W_0$ from $D_0$; TSS and NSGA-II produce $P_0$ and $A_0$. Dashed teal marks model-authored code; solid blue-gray marks fixed execution.}
\label{fig:method-initialization}
\end{figure}

MAPLE operates in two stages: constructing an initial executable state and revising it as requests arrive. TSS connects the model's decision declarations to fixed search operations and solvers during construction (Figure~\ref{fig:method-initialization}). For later requests, LPD locates the required edits and LSM resolves references to accepted history. Restart selection determines whether evolutionary search reuses earlier candidates or starts afresh. The fixed runtime checks the program and solver output before committing the updated state (Figure~\ref{fig:method-live-update}).

\begingroup\interlinepenalty=10000
\paragraph{TSS: bind decision types to problem semantics.}
The model writes \texttt{build\_problem(public\_context)} to read public tables, construct evaluation data, and declare solver mode, objectives, and decision segments. A segment contains binary, categorical, bounded integer, bounded real, permutation, or assignment decisions; segments compose a typed candidate. \texttt{evaluate(genome, data)} decodes the candidate into a plan and returns objective values and constraint violations.
\par\endgroup

\begingroup\interlinepenalty=10000
Declaring a type binds its fixed initialization, crossover, mutation, and domain-repair operators.
Real vectors use simulated binary crossover and polynomial mutation \citep{deb2002nsga}.
Permutations use order crossover and swap mutation \citep{davis1985applying}, while discrete vectors and assignments use gene-wise parent choice and type-valid mutation \citep{holland1975adaptation,syswerda1989uniform}.
\par\endgroup

All four operators preserve the declared representation space $\mathcal{X}(W_t,D_t)$; \texttt{evaluate()} checks problem constraints.
The declared formulation selects the solver. For linear and mixed-integer linear problems, the fixed interface builds a model from variables, bounds, objective coefficients, and constraints, then extracts the requested plan from returned variable values \citep{dantzig1963linear,nemhauser1988integer}. Scalar nonlinear or combinatorial tasks use GA \citep{holland1975adaptation}, and separate objectives use NSGA-II \citep{deb2002nsga}. The evolutionary loop evaluates candidates, selects parents, applies variation and repair, and updates the population and Pareto set; stopping uses Workbench search progress.

\textit{Routing example (Figure~\ref{fig:method-initialization}).}
\texttt{build\_problem()} reads the depot, order, vehicle, and policy tables, keeps active orders and available vehicles, and declares two decision segments. An assignment $a_i$ selects each order's vehicle; a permutation $\pi$ specifies visit order. Assignment initialization and mutation sample allowed vehicles, while crossover inherits each order's vehicle from a parent. Permutations use shuffling, order crossover, and swaps, retaining each active order exactly once. Assignment repair corrects invalid vehicle choices.

\texttt{evaluate()} groups orders by $a_i$ in $\pi$ order into one depot-to-depot route per used vehicle. It accumulates Euclidean distance, simulates travel, waiting, and service for lateness, and computes emissions from distance, the vehicle emission rate, and carbon multiplier. Capacity excess, late shift return, and missing orders produce constraint violations. It returns routes and three objectives to the fixed NSGA-II loop.

\paragraph{LPD: locate the requested revision.}
LPD reads the public update, decision declarations, stored events, and accepted result to identify changes to data, decisions, evaluation logic, or historical bindings. The system applies supplied table edits directly. When a change is described only in language, the model translates it into edits to the input tables. The editor receives the requested Workbench functions and preserves the remaining functions. A price change can therefore update the tables alone, while a new constraint may require an evaluation edit.

\begingroup\interlinepenalty=10000
\paragraph{LSM: bind requests to accepted history.}
LSM stores events, accepted decisions, and search candidates. For a request to preserve the assignment of a previously mentioned customer, the event identifies the customer and the accepted plan supplies the vehicle. A typed binding query retrieves this relation and checks it against the current decision domain. The runtime enforces the resolved assignment during search. Events are added to the committed history after an update is accepted.
Requests about earlier assignments use one saved plan from the preceding accepted state. The full Pareto set is stored separately. In the recorded runs, the saved plan is the first candidate returned by final NSGA-II selection.
\par\endgroup

\begin{figure}[!htb]
\MapleDiagram{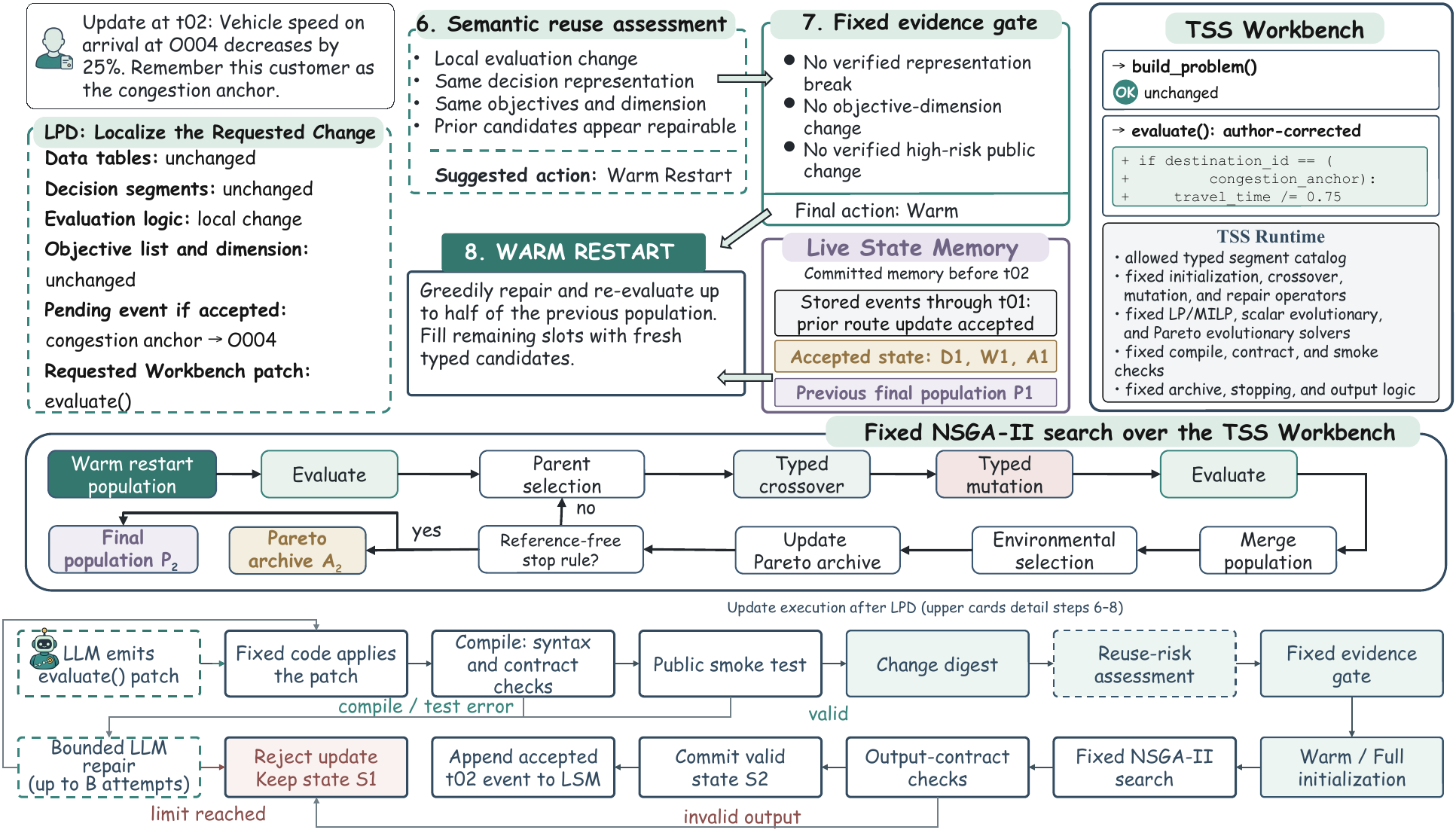}{\textbf{A localized routing update.}
LPD locates edits and LSM resolves historical references. Compilation and public-test errors trigger bounded repair; invalid search outputs reject the update. Valid outputs commit the state and history.}{fig:method-live-update}
\end{figure}

\paragraph{Restart selector: estimate reuse risk.}
After compilation and a small input check, the runtime summarizes changes to public data, decision types, and objectives. The model uses this summary and accepted history to assess reuse risk and recommend Warm or Full. Its explanation cites changed fields or declarations, such as replaced active orders, reversed objective preferences, or changed machine capacities and energy costs. Algorithm~\ref{alg:restart-gate} verifies the cited changes and structural support for each proposed cause.

\begin{algorithm}[!htbp]
\caption{Fixed gate for semantic restart selection}
\label{alg:restart-gate}
\small
\begin{algorithmic}[1]
\Require public-change summary $\Delta_t$, model assessment $(r,M,E,v)$
\Ensure restart action Warm or Full
\State $V \gets M \cap \operatorname{SupportedMechanisms}(\Delta_t)$ \Comment{check structural support}
\State $C \gets E \cap \operatorname{ChangedFields}(\Delta_t)$ \Comment{verify the cited changes}
\State \Return Full if $r=\texttt{high}$ and $v=\texttt{true}$ and $V\ne\varnothing$ and $C\ne\varnothing$;
\Statex \hspace{\algorithmicindent}\textbf{otherwise} \Return Warm.
\end{algorithmic}
\end{algorithm}

\begingroup\interlinepenalty=10000
Here $r$ is the assessed reuse risk, $M$ contains the model's proposed causes of that risk, $E$ identifies the public changes cited in its explanation, and $v$ is true when it recommends initializing the whole population afresh.
\par\endgroup
$\operatorname{SupportedMechanisms}(\Delta_t)$ checks which of the five allowed causes have matching structural changes. For example, changing the declared decision segments supports replacement of decision choices; changes to multiple resource attributes, such as CPU capacity and energy cost, support a change in the relative usefulness of resources.
$\operatorname{ChangedFields}(\Delta_t)$ lists the public data and program declarations that actually changed, including table values and membership, decision segments, objectives, and the two editable functions. Thus $V$ retains supported causes and $C$ retains verified cited changes. Appendix~\ref{app:restart-gate-definition} specifies the five causes and their structural tests.

Warm maps the preceding population and archive to the revised decisions, then uses local typed repair to prioritize feasibility and lower penalized scalar fitness. Repaired candidates fill at most half the population; fresh candidates fill the rest. Full initializes every candidate afresh. Both use the same subsequent solver and stopping rule; NSGA-II compares objectives separately. Warm additionally evaluates candidates during migration and repair.

\paragraph{Validation and commitment.}
The runtime compiles each edit, checks the required function interfaces, and executes a small input example. Errors in these checks trigger at most $B$ model repair attempts. The fixed solver then runs and checks the required output fields and structure. A failure rejects the revision and preserves the preceding state. A passing result commits the revised data, Workbench, accepted output, and search state, then appends the accepted event. Held-out evaluation scores the saved solution after execution.
Appendix~\ref{app:architecture-details} gives the control flow, and Appendix~\ref{app:skill-catalog} records the editable surfaces and available choices.
			\section{Experiments}
\label{sec:experiments}
\newcommand{\LiveOptMatchedStageHVPHeader}{%
 & \textbf{P010} & \textbf{P011} & \textbf{P012} & \textbf{P013} & \textbf{P014} & \textbf{P015} \\
\rowcolor{LiveOptTableHeader}
\scriptsize\textbf{Order} & \AblationHVOrder & \AblationHVOrder & \AblationHVOrder & \AblationHVOrder & \AblationHVOrder & \AblationHVOrder \\
}
\newcommand{\LiveOptMatchedStageHVSummaryRows}{%
\rowcolor{LiveOptTypeSummary}
\textbf{Type} & \multicolumn{3}{c}{\textbf{Routing}\quad \AblationHVCell{\textbf{0.757}}{0.391}{0.733}{0.656}} & \multicolumn{3}{c}{\textbf{Cloud}\quad \AblationHVCell{0.801}{\textbf{0.821}}{0.799}{0.805}} \\
}
\newcommand{\LiveOptMatchedStageHVRows}{%
\textbf{All-t} & \AblationHVCell{\textbf{0.752}}{0.000}{0.724}{0.663} & \AblationHVCell{\textbf{0.736}}{0.327}{0.712}{0.662} & \AblationHVCell{0.785}{\textbf{0.845}}{0.764}{0.643} & \AblationHVCell{\textbf{0.854}}{0.837}{0.853}{0.851} & \AblationHVCell{0.763}{\textbf{0.870}}{0.761}{0.773} & \AblationHVCell{0.785}{0.755}{0.784}{\textbf{0.791}} \\
t01 & \AblationHVCell{\textbf{0.746}}{0.000}{\textbf{0.746}}{0.710} & \AblationHVCell{\textbf{0.731}}{0.379}{\textbf{0.731}}{0.663} & \AblationHVCell{0.847}{\textbf{0.995}}{0.847}{0.684} & \AblationHVCell{\textbf{0.915}}{0.813}{\textbf{0.915}}{0.873} & \AblationHVCell{0.815}{\textbf{0.905}}{0.815}{0.833} & \AblationHVCell{0.826}{\textbf{0.883}}{0.826}{0.816} \\
\rowcolor{LiveOptBaselineRowB}
t02 & \AblationHVCell{\textbf{0.774}}{0.000}{\textbf{0.774}}{0.657} & \AblationHVCell{\textbf{0.776}}{0.492}{\textbf{0.776}}{0.726} & \AblationHVCell{\textbf{0.872}}{0.838}{\textbf{0.872}}{0.744} & \AblationHVCell{0.917}{0.901}{0.917}{\textbf{0.929}} & \AblationHVCell{0.743}{\textbf{0.880}}{0.743}{0.763} & \AblationHVCell{0.821}{\textbf{0.896}}{0.821}{0.836} \\
t03 & \AblationHVCell{\textbf{0.791}}{0.000}{\textbf{0.791}}{0.769} & \AblationHVCell{\textbf{0.778}}{0.440}{\textbf{0.778}}{0.730} & \AblationHVCell{0.815}{\textbf{0.880}}{0.815}{0.642} & \AblationHVCell{0.899}{0.874}{0.899}{\textbf{0.901}} & \AblationHVCell{0.782}{\textbf{0.881}}{0.782}{0.764} & \AblationHVCell{0.846}{\textbf{0.915}}{0.846}{0.859} \\
\rowcolor{LiveOptBaselineRowB}
t04 & \AblationHVCell{\textbf{0.778}}{0.000}{\textbf{0.778}}{0.696} & \AblationHVCell{\textbf{0.768}}{0.450}{\textbf{0.768}}{0.704} & \AblationHVCell{0.832}{\textbf{0.931}}{0.832}{0.659} & \AblationHVCell{0.879}{0.878}{0.879}{\textbf{0.898}} & \AblationHVCell{0.770}{\textbf{0.892}}{0.770}{0.795} & \AblationHVCell{0.833}{\textbf{0.900}}{0.833}{0.835} \\
t05 & \AblationHVCell{\textbf{0.805}}{0.000}{\textbf{0.805}}{0.668} & \AblationHVCell{\textbf{0.797}}{0.397}{\textbf{0.797}}{0.692} & \AblationHVCell{0.838}{\textbf{0.869}}{0.838}{0.640} & \AblationHVCell{0.911}{0.902}{0.911}{\textbf{0.913}} & \AblationHVCell{0.823}{\textbf{0.927}}{0.823}{0.840} & \AblationHVCell{0.834}{\textbf{0.905}}{0.834}{0.854} \\
\rowcolor{LiveOptBaselineRowB}
t06 & \AblationHVCell{\textbf{0.849}}{0.000}{\textbf{0.849}}{0.751} & \AblationHVCell{\textbf{0.842}}{0.241}{\textbf{0.842}}{0.772} & \AblationHVCell{0.872}{\textbf{0.895}}{0.872}{0.723} & \AblationHVCell{0.898}{0.888}{0.898}{\textbf{0.909}} & \AblationHVCell{0.843}{\textbf{0.927}}{0.843}{0.860} & \AblationHVCell{0.846}{\textbf{0.908}}{0.846}{0.859} \\
t07 & \AblationHVCell{\textbf{0.850}}{0.000}{\textbf{0.850}}{0.733} & \AblationHVCell{\textbf{0.846}}{0.289}{\textbf{0.846}}{0.736} & \AblationHVCell{0.876}{\textbf{0.940}}{0.876}{0.688} & \AblationHVCell{0.924}{0.914}{0.924}{\textbf{0.924}} & \AblationHVCell{0.748}{\textbf{0.907}}{0.748}{0.786} & \AblationHVCell{0.813}{\textbf{0.900}}{0.813}{0.840} \\
\rowcolor{LiveOptBaselineRowB}
t08 & \AblationHVCell{\textbf{0.862}}{0.000}{\textbf{0.862}}{0.764} & \AblationHVCell{\textbf{0.784}}{0.463}{\textbf{0.784}}{0.629} & \AblationHVCell{0.905}{\textbf{0.931}}{0.905}{0.699} & \AblationHVCell{\textbf{0.887}}{0.880}{\textbf{0.887}}{0.885} & \AblationHVCell{0.749}{\textbf{0.897}}{0.749}{0.781} & \AblationHVCell{0.845}{\textbf{0.915}}{0.845}{0.837} \\
t09 & \AblationHVCell{\textbf{0.886}}{0.000}{\textbf{0.886}}{0.659} & \AblationHVCell{\textbf{0.804}}{0.532}{\textbf{0.804}}{0.715} & \AblationHVCell{0.891}{\textbf{0.938}}{0.891}{0.706} & \AblationHVCell{\textbf{0.882}}{0.866}{\textbf{0.882}}{0.873} & \AblationHVCell{0.751}{\textbf{0.921}}{0.751}{0.725} & \AblationHVCell{0.834}{\textbf{0.912}}{0.834}{0.831} \\
\rowcolor{LiveOptBaselineRowB}
t10 & \AblationHVCell{\textbf{0.910}}{0.000}{\textbf{0.910}}{0.785} & \AblationHVCell{\textbf{0.918}}{0.238}{\textbf{0.918}}{0.797} & \AblationHVCell{\textbf{0.898}}{0.797}{\textbf{0.898}}{0.766} & \AblationHVCell{\textbf{0.891}}{0.866}{\textbf{0.891}}{0.870} & \AblationHVCell{0.823}{\textbf{0.921}}{0.823}{0.825} & \AblationHVCell{0.871}{\textbf{0.920}}{0.871}{0.867} \\
t11 & \AblationHVCell{\textbf{0.342}}{0.000}{0.230}{\textbf{0.342}} & \AblationHVCell{\textbf{0.239}}{0.000}{0.040}{\textbf{0.239}} & \AblationHVCell{0.397}{\textbf{0.656}}{0.395}{0.397} & \AblationHVCell{0.768}{\textbf{0.780}}{0.763}{0.768} & \AblationHVCell{0.762}{\textbf{0.770}}{0.767}{0.762} & \AblationHVCell{\textbf{0.858}}{0.000}{0.857}{\textbf{0.858}} \\
\rowcolor{LiveOptBaselineRowB}
t12 & \AblationHVCell{\textbf{0.427}}{0.000}{0.203}{\textbf{0.427}} & \AblationHVCell{\textbf{0.545}}{0.000}{0.458}{\textbf{0.545}} & \AblationHVCell{0.370}{\textbf{0.469}}{0.129}{0.370} & \AblationHVCell{0.473}{\textbf{0.486}}{0.475}{0.473} & \AblationHVCell{0.546}{\textbf{0.617}}{0.520}{0.546} & \AblationHVCell{\textbf{0.197}}{0.000}{0.178}{\textbf{0.197}} \\
}
\newcommand{\LiveOptTSSMechanismRows}{%
\rowcolor{LiveOptRowAccent}
\method{} & 100.0\% & 0 & \LiveOptPublicUpdateHV & \LiveOptPublicUpdateHV \\
\method{} w/o TSS & 80.6\% & 140 & \NoTSSPublicUpdateHV & \NoTSSPublicConditionalHV \\
}
\newcommand{\LiveOptTSSFailureRows}{%
P010 (routing) & 120/120 & one vehicle assigned to two routes & vehicle uniqueness omitted from the evaluation function \\
P015 (cloud) & 20/120 & GPU-incompatible placements & violations recorded without clearing the feasible flag \\
}
\newcommand{\LiveOptRestartRegretDMRows}{%
\rowcolor{LiveOptRowAccent}
\method{} & 120/720 & 0.779 & 162.8 & 0.004 \\
Matched Warm & 0/720 & 0.766 & 162.7 & 0.017 \\
Numerical selector & 245/720 & 0.765 & 169.4 & 0.018 \\
Matched Full & 720/720 & 0.731 & 199.0 & 0.053 \\
Best action after results & 300/720 & 0.783 & 167.8 & 0.000 \\
}
\newcommand{\LiveOptRestartRegretDSRows}{%
\rowcolor{LiveOptRowAccent}
\method{} & 4/36 & 0.927 & 66.7 & 0.016 \\
Matched Warm & 0/36 & 0.927 & 66.8 & 0.016 \\
Matched Full & 36/36 & 0.919 & 71.1 & 0.024 \\
Best action after results & 6/36 & 0.943 & 66.5 & 0.000 \\
}
\newcommand{\LiveOptChurnDiscriminatorRows}{%
P010 parameter rescaling & 0.716 & 0.510 & $+0.206\pm0.433$ & -0.294 & +0.478 \\
P010 ID relabel & 0.752 & 0.603 & $+0.149\pm0.196$ & -0.060 & +0.329 \\
\TightStack{P010: Added\\low-demand orders} & 0.769 & 0.061 & $+0.708\pm0.236$ & +0.442 & +0.892 \\
P015 parameter rescaling & 0.989 & 0.990 & $-0.002\pm0.003$ & -0.005 & +0.001 \\
P015 ID relabel & 0.980 & 0.980 & $0.000\pm0.012$ & -0.013 & +0.008 \\
\TightStack{P015: Added\\small jobs} & 0.968 & 0.946 & $+0.022\pm0.021$ & +0.005 & +0.045 \\
\midrule
\rowcolor{LiveOptRowAccent}
All six cases & 0.862 & 0.682 & +0.181 & -- & -- \\
}
\newcommand{\LiveOptChurnCompactRows}{%
Renamed entities & +0.149 & 0.000 \\
Added tasks & +0.708 & +0.022 \\
}
\newcommand{\MainDynamicHVSummaryRows}{%
\rowcolor{LiveOptRowAccent}
\method{} & 100\% & 0.951 & 100\% & 0.875 \\
Persistent ReAct\Adapted{} & 51.3\% & 0.501 & 11.5\% & 0.042 \\
ReAct\Adapted{} & 33.3\% & 0.333 & 2.6\% & 0.023 \\
\rowcolor{LiveOptBaselineRowB}
OptiMUS\Adapted{} & 28.2\% & 0.281 & 12.8\% & 0.084 \\
ORLM\Adapted{} & 12.0\% & 0.120 & 0\% & 0.000 \\
\rowcolor{LiveOptBaselineRowB}
OptimAI\Adapted{} & 7.7\% & 0.077 & 11.5\% & 0.051 \\
OR-LLM-Agent\Adapted{} & 11.1\% & 0.111 & 12.8\% & 0.046 \\
}
\newcommand{\MainDynamicProtocolRows}{%
\rowcolor{LiveOptRowAccent}
\method{} & 100\% & 0.951 & 0.951 & 100\% & 0.875 & 0.875 \\
Persistent ReAct\Adapted{} & 51.3\% & 0.501 & 0.763 & 11.5\% & 0.042 & 0.185 \\
ReAct\Adapted{} & 33.3\% & 0.333 & 0.376 & 2.6\% & 0.023 & 0.091 \\
\rowcolor{LiveOptBaselineRowB}
OptiMUS\Adapted{} & 28.2\% & 0.281 & 0.333 & 12.8\% & 0.084 & 0.118 \\
ORLM\Adapted{} & 12.0\% & 0.120 & 0.298 & 0\% & 0.000 & 0.021 \\
\rowcolor{LiveOptBaselineRowB}
OptimAI\Adapted{} & 7.7\% & 0.077 & 0.103 & 11.5\% & 0.051 & 0.059 \\
OR-LLM-Agent\Adapted{} & 11.1\% & 0.111 & 0.171 & 12.8\% & 0.046 & 0.046 \\
}
\newcommand{\MainStaticGroupedRows}{%
\rowcolor{LiveOptRowAccent}
\method{} & 93.2\% & 59.3\% & \textbf{100.0\%} & \textbf{80.0\%} & \textbf{100.0\%} & \textbf{1.000} & \textbf{100.0\%} & \textbf{\LiveOptPublicSMHV} \\
Persistent ReAct & 98.3\% & 62.7\% & \textbf{100.0\%} & 70.0\% & 88.9\%\Adapted & 0.889\Adapted & 50.0\%\Adapted & \PersistentPublicSMHV\Adapted \\
\rowcolor{LiveOptBaselineRowB}
ReAct & 78.0\% & 57.6\% & 95.0\% & 70.0\% & 88.9\%\Adapted & 0.889\Adapted & 33.3\%\Adapted & \ReactPublicSMHV\Adapted \\
OR-LLM-Agent & 96.6\% & 50.8\% & 95.0\% & 65.0\% & 77.8\%\Adapted & 0.778\Adapted & 66.7\%\Adapted & \ORAgentPublicSMHV\Adapted \\
\rowcolor{LiveOptBaselineRowB}
ORLM & 61.0\% & 23.7\% & 55.0\% & 45.0\% & 100.0\%\Adapted & 1.000\Adapted & 0.0\%\Adapted & \ORLMPublicSMHV\Adapted \\
OptiMUS & 93.2\% & \textbf{69.5\%} & \textbf{100.0\%} & 65.0\% & 88.9\%\Adapted & 0.889\Adapted & 66.7\%\Adapted & \OptimusPublicSMHV\Adapted \\
OptimAI & \textbf{98.3\%} & 61.0\% & 80.0\% & 55.0\% & 55.6\%\Adapted & 0.556\Adapted & 66.7\%\Adapted & \OptimAIPublicSMHV\Adapted \\
}
\newcommand{\NLDOExternalContinuityRows}{%
Persistent ReAct & 126/180 & 96.8\% & 58/180 & 3.87 \\
\rowcolor{LiveOptBaselineRowB}
ReAct & 55/180 & 83.6\% & 31/180 & 2.07 \\
OptiMUS & 53/180 & 83.0\% & 31/180 & 2.07 \\
\rowcolor{LiveOptBaselineRowB}
ORLM & 35/180 & 77.1\% & 5/180 & 0.33 \\
OptimAI & 18/180 & 61.1\% & 9/180 & 0.60 \\
\rowcolor{LiveOptBaselineRowB}
OR-LLM-Agent & 26/180 & 65.4\% & 12/180 & 0.80 \\
}
\newcommand{\LiveOptTransitionAuditRows}{%
LPD prediction matches accepted change & 180 & 170/180 (94.4\%) \\
Data/decision/evaluation agreement & 180 & 96.7\% / 98.9\% / 98.9\% \\
Both editable functions unchanged & 180 & 166/180 (92.2\%) \\
Reference entity/field/value & 30 & 30/30 (100.0\%) \\
Valid updated state & 180 & 180/180 (100.0\%) \\
}
\newcommand{\LiveOptMemoryChainRows}{%
1 & 3 & 3 & 3 & 3 \\
2 & 21 & 21 & 21 & 21 \\
3 & 6 & 6 & 6 & 6 \\
}
\newcommand{\LiveOptBenchmarkCharacterizationRows}{%
Selection/allocation & 3 & 36 & 6 & 0 & 22 \\
Precedence scheduling & 3 & 36 & 9 & 0 & 15 \\
Coverage rostering & 3 & 36 & 6 & 6 & 19 \\
Ordered service routing & 3 & 36 & 6 & 6 & 26 \\
Cloud placement & 3 & 36 & 3 & 6 & 26 \\
}
\newcommand{\PublicComponentRows}{%
\rowcolor{LiveOptRowAccent}
\method{} & 100.0\% & 0.879$\pm$0.014 & 0.879 & 0.079 \\
Always Warm & 100.0\% & 0.868$\pm$0.016 & 0.868 & 0.091 \\
Always Full & 100.0\% & 0.824$\pm$0.015 & 0.824 & 0.111 \\
w/o TSS & 80.6\% & 0.665$\pm$0.004 & 0.826 & 0.143 \\
}
\newcommand{\PublicComponentCompactRows}{%
\rowcolor{LiveOptRowAccent}
\method{} & 100.0\% & 0.879$\pm$0.014 & 0.079 \\
Always Warm & 100.0\% & 0.868$\pm$0.016 & 0.091 \\
Always Full & 100.0\% & 0.824$\pm$0.015 & 0.111 \\
w/o TSS & 80.6\% & 0.665$\pm$0.004 & 0.143 \\
}
\newcommand{\LivePublicOrdinaryHV}{0.915}
\newcommand{\LivePublicShockHV}{0.696}
\newcommand{\WarmPublicOrdinaryHV}{0.915}
\newcommand{\WarmPublicShockHV}{0.636}
\newcommand{\FullPublicOrdinaryHV}{0.850}
\newcommand{\FullPublicShockHV}{0.696}
\newcommand{\LiveOptCodeTokens}{51.6k}
\newcommand{\NoTSSCodeTokens}{72.8k}
\newcommand{\TSSCodeTokenReduction}{29.1\%}
\newcommand{\LiveOptControllerCostRows}{%
\method{} & 4.0 & 51.6k & 68.9k & 135.0k & 255.4k \\
\method{} w/o TSS & 4.8 & 72.8k & Reused & Reused & 72.8k \\
}

\newcommand{\LiveOptCodeConstructionRows}{%
\method{} & 4.0 & 51.6k & 68.9k & 135.0k \\
\method{} w/o TSS & 4.8 & 72.8k & Reused & Reused \\
}

We evaluate initial solving, continued plan revision, use of earlier information, and candidate reuse.

\paragraph{Compared agents.}
We compare MAPLE with ReAct, Persistent ReAct, and four optimization-agent workflows: ORLM, OptiMUS, OR-LLM-Agent, and OptimAI. The NLDO adapters provide cumulative public requests, execution feedback, and each method's available history, with at most three code-output repairs. Persistent ReAct carries its solver code and accepted plan between requests and can return a Pareto set for each request. Appendix~\ref{app:baseline-interfaces} specifies the workflow implementations and carried state.

\paragraph{Model runs and numerical search.}
The main comparison uses DeepSeek-V4-Pro (Preview), with API ID \texttt{deepseek-v4-pro}. For each MAPLE episode, one sequence of model outputs supplies the optimization programs and restart choices. We repeat the numerical search with ten random seeds using this same sequence. Each search uses 200 candidates and at most 200 generations. External methods contribute one model trajectory per episode. The independent-model controls in Appendix~\ref{app:component-details} vary Workbench generation separately.
We set the Workbench repair limit to $B=3$. Stopping uses the Workbench's search progress (Appendix~\ref{app:runtime-config}).

In full-sequence comparisons, each reuse strategy continues with the candidates produced by its own earlier searches. In single-update comparisons, the strategies start from MAPLE's same saved population. Only the initialization choice differs.

\paragraph{Evaluation.}
Search uses the Workbench evaluator; restart selection reads public history and the change summary. Held-out scores and references are used only offline. We evaluate saved Pareto plans using the revised public formulas and new reference archives (Appendix~\ref{app:execution-scoring-versions}). HV divides the submitted archive's hypervolume by the state-specific reference's and can exceed one. IGD uses the same scales for nonempty feasible archives. Alongside the strict prefix defined in Section~\ref{sec:task}, we report a mathematical prefix that accepts feasible saved plans regardless of auxiliary fields. Appendices~\ref{app:metric-definitions} and~\ref{app:update-semantics} provide normalization and formula details.

\subsection{Q1: Can the agent solve the initial request?}

We first evaluate the initial request before testing later revisions. Table~\ref{tab:main-static-results} reports execution and reference-objective agreement. MAPLE reaches 59.3\% on NLP4LP-hard and 80.0\% on BWOR20, with scalar quality of 1.000 and Pareto HV of 0.832 on the NLDO initial states. Appendix~\ref{app:static-solving-supplement} reports tolerances and workflow diagnostics.

\begin{table}[!htbp]
\centering
\begingroup
\MainResultTableSetup
\setlength{\tabcolsep}{2.8pt}
\begin{tabularx}{\linewidth}{@{}l*{8}{Y}@{}}
\toprule
\TableHead
& \multicolumn{2}{c}{\textbf{NLP4LP-hard}} & \multicolumn{2}{c}{\textbf{BWOR20}} & \multicolumn{2}{c}{\textbf{NLDO-SS}} & \multicolumn{2}{c}{\textbf{NLDO-SM}} \\
\cmidrule(lr){2-3}\cmidrule(lr){4-5}\cmidrule(lr){6-7}\cmidrule(lr){8-9}
\TableHead
\textbf{Method} & \textbf{Exec.} & \textbf{Obj.} & \textbf{Exec.} & \textbf{Obj.} & \textbf{Valid} & \textbf{Quality} & \textbf{Valid} & \textbf{HV} \\
\midrule
\MainStaticGroupedRows
\bottomrule
\end{tabularx}
\endgroup
\caption{\textbf{Static execution and solution quality.} Exec.: successful execution. Obj.: reference-objective agreement over all cases. Valid: accepted and feasible under the recorded output protocol. HV: public-formula ratio, zero for invalid outputs. Appendix~\ref{app:static-solving-supplement} reports supplementary diagnostic checks. \Adapted{} marks task-interface adaptation.}
\label{tab:main-static-results}
\end{table}

\subsection{Q2: Can the agent follow a sequence of revisions?}

\begin{figure}[!htbp]
\centering
\begin{minipage}[c]{0.65\linewidth}
\includegraphics[width=\linewidth]{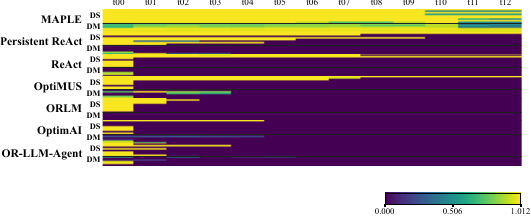}
\end{minipage}\hfill
\begin{minipage}[c]{0.32\linewidth}
\caption{\textbf{Strict-prefix quality through updates.} Upper: scalar; lower: Pareto. Columns cover t00--t12. Pareto quality uses public-formula evaluation. A protocol rejection ends the eligible prefix.}
\label{fig:main-dynamic-heatmap}
\end{minipage}
\end{figure}

We test whether each agent continues to return accepted, feasible plans as the user changes the request. Acceptance checks the required output fields and structure. Feasibility checks whether the saved plan satisfies the problem constraints.
Figure~\ref{fig:main-dynamic-heatmap} shows per-state quality, and Table~\ref{tab:main-dynamic-results} summarizes continuity and quality over the same sequences.

\begin{table}[!htbp]
\centering
\begingroup
\MainResultTableSetup
\setlength{\tabcolsep}{4pt}
\begin{tabularx}{\linewidth}{@{}l*{6}{Y}@{}}
\toprule
\TableHead
& \multicolumn{3}{c}{\textbf{Dynamic single-objective}} & \multicolumn{3}{c}{\textbf{Dynamic multi-objective}} \\
\cmidrule(lr){2-4}\cmidrule(lr){5-7}
\TableHead
\textbf{Method} & \TightStack{\textbf{Solve}\\\textbf{rate}} & \TightStack{\textbf{Strict}\\\textbf{Quality}} & \TightStack{\textbf{Math.}\\\textbf{Quality}} & \TightStack{\textbf{Solve}\\\textbf{rate}} & \TightStack{\textbf{Strict}\\\textbf{HV}} & \TightStack{\textbf{Math.}\\\textbf{HV}} \\
\midrule
\MainDynamicProtocolRows
\bottomrule
\end{tabularx}
\endgroup
\caption{\textbf{Continuity and quality}, t00--t12. Solve rate uses strict prefixes. Math.: mathematical prefix. Pareto scores use public formulas. Both protocols assign zero to missing suffixes. \Adapted{} marks dynamic adaptation.}
\label{tab:main-dynamic-results}
\end{table}

MAPLE completes every trajectory in the main run and reaches strict-prefix online quality of 0.951 on scalar tasks and 0.875 HV on Pareto tasks. Persistent ReAct obtains 0.501 and 0.042, respectively. Seven Persistent ReAct scalar trajectories first fail the output checks because a required reasoning field is missing. One of these outputs also contains an infeasible plan. Accepting mathematically feasible saved plans regardless of auxiliary fields raises Persistent ReAct's scores to 0.763 and 0.185. MAPLE's scores are unchanged.

\paragraph{Program-generation cost.}
\label{sec:cost}

We count input and output tokens for program generation, editing, and repair. Across four additional runs per method, these operations use 51.6k tokens per MAPLE trajectory and 72.8k without TSS. The version without TSS reuses saved decisions about which parts to edit and how to update the input tables. Appendix~\ref{app:component-details} reports these costs separately.

\FloatBarrier
\subsection{Q3: What helps the agent handle revisions correctly?}

\paragraph{TSS reduces invalid outputs.}
We compare MAPLE with a version that generates its own candidate representation, evaluation, search operations, and repair functions instead of using TSS. Both versions use the same fixed numerical search loop. Removing TSS lowers feasibility from 100.0\% to 80.6\% and mean HV from 0.879 to 0.665 over the Pareto updates (Table~\ref{tab:main-tss-mechanism}).

\begin{table}[H]
\LiveOptEqualHeightPair{0.675\linewidth}{0.295\linewidth}{%
\centering
\begingroup
\ResultTableSetup
\begin{tabularx}{\linewidth}{@{}Zccc@{}}
\toprule
\TableHead
\textbf{Method} & \textbf{Feasible} & \textbf{HV $\uparrow$} & \textbf{IGD} \\
\midrule
\PublicComponentCompactRows
\bottomrule
\end{tabularx}
\endgroup
\refstepcounter{table}\label{tab:main-tss-mechanism}
\par\vspace{3pt}
\raggedright\small
Table~\thetable: \textbf{Scaffolding and sequential reuse.} Six NLDO-DM episodes, t01--t12, ten seeds. Public-formula HV: mean $\pm$ sample SD across seed means, with invalid states scored zero. IGD uses valid outputs.
}{%
\small\raggedright
The version without TSS returns 140 invalid outputs among 720 update--seed results. Its routing program omits the check that a vehicle serves only one route. Its cloud-placement program fails to reject GPU-incompatible assignments.
\par\vfill
Its valid outputs have mean HV of 0.826.

}
\end{table}

\paragraph{Search operations tailored to the decisions.}
We change only how the search combines and modifies candidates. The control copies a complete decision block from one parent or generates a new valid block at random. The optimization program, repair procedure, starting population, and restart choice remain fixed. Mean HV falls from 0.774 to 0.618 across one routing task and one cloud-placement task (Table~\ref{tab:appendix-tss-operator-control}). The comparison covers twelve updates and three search seeds under the original evaluation rules and reference sets.

\paragraph{Variation across program generations and language models.}
We compare three independently generated program sequences per method for one routing task and one cloud-placement task. These include the main sequence, and restart choices remain fixed. Under the original evaluation rules, mean HV is 0.669 for MAPLE and 0.336 without TSS on routing. The corresponding cloud-placement values are 0.728 and 0.513 (Table~\ref{tab:controller-repeat}). A separate full-benchmark run uses Kimi k2.7 under the common public-formula evaluation. Its scalar solve rate and quality are 100.0\% and 0.937. Its Pareto solve rate and HV are 92.3\% and 0.766 (Table~\ref{tab:appendix-cross-model-main}).

\paragraph{Agreement between predicted and saved edits.}
We compare LPD's proposed changes to the input data, decision definitions, and evaluation logic with the changes saved after each update. They agree in 170 of 180 updates. Neither Workbench function changes in 166 updates. Appendix~\ref{app:component-details} reports the disagreements.

\paragraph{Using earlier updates and accepted plans.}
We first check 30 benchmark requests that identify an entity through earlier updates. LSM resolves all 30, including six that require following a chain of three references.

We then test 18 assignment requests across two domains, three information requirements, and three wordings. Six can be answered from earlier update records. Six explicitly name entities and ask to preserve their assignments from the accepted plan. The remaining six require both records. For example, keeping the vehicle assignment of the customer mentioned in a congestion note requires identifying the customer from that note and retrieving its vehicle from the saved plan.

With both records, LSM resolves all 18 requests. Each partial condition resolves only its six directly answerable requests; neither resolves those requiring both records (Table~\ref{tab:state-binding-factorial}a).

Separately, search replay uses saved queries and the original online plans in six cases with three search seeds each. Full LSM and a control given the required assignment directly each produce 18 valid results (Table~\ref{tab:state-binding-factorial}b): the correct assignment is retrieved and enforced, and every returned plan satisfies the constraints.

\FloatBarrier
\begingroup\interlinepenalty=10000
\subsection{Q4: When should search reuse previous solutions?}
\label{sec:reuse-evidence}

\paragraph{Restart choices across task revisions.}
We compare MAPLE with Always Warm and Always Full over six Pareto task sequences. The controls fix the restart action at every update; each strategy carries candidates from its own searches. Mean update HV is 0.879 for MAPLE, 0.868 for Always Warm, and 0.824 for Always Full.
\par\endgroup

\begin{figure}[!htbp]
\LiveOptCaptionedPair{0.665\linewidth}{0.305\linewidth}{%
\centering
\includegraphics[width=\linewidth]{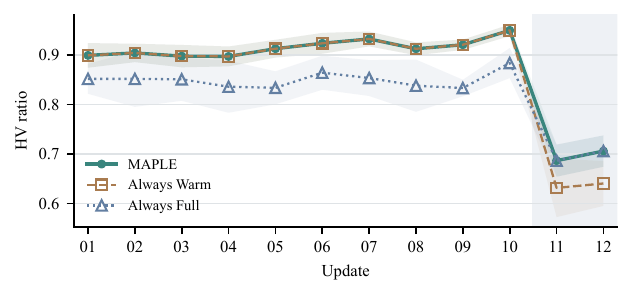}
}{%
\refstepcounter{figure}\label{fig:public-sequential-restart}
Figure~\thefigure: \textbf{Reuse before disruption, restart after it.} Public-formula HV over six episodes: ten-seed means and one-SD bands. Each policy carries its own population; gray marks disruptive updates.
}{%
\centering
\begingroup
\ResultTableSetup
\begin{tabularx}{\linewidth}{@{}Zcc@{}}
\toprule
\TableHead
\textbf{Change} & \textbf{Routing} & \textbf{Cloud} \\
\midrule
\LiveOptChurnCompactRows
\bottomrule
\end{tabularx}
\endgroup
\par\smallskip
\small\raggedright
\textbf{Chosen action:} MAPLE selects Warm; both rules select Full.
}{%
\refstepcounter{table}\label{tab:main-reuse-discriminator}
Table~\thetable: \textbf{Reuse after renaming or adding tasks.} Warm minus Full HV: three paired seeds per case, with separate pooled references under the original evaluation rules. Numerical decisions precede identifier mapping (Appendix~\ref{app:reuse-discriminator}).
}
\end{figure}

MAPLE and Always Warm both obtain 0.915 on the first ten updates. The last two updates replace substantial parts of the tasks and resources. MAPLE selects Full and obtains 0.696, compared with 0.636 for Always Warm. The gains are concentrated in routing (Figure~\ref{fig:public-sequential-restart}).

For each sequence, we compute the paired difference between MAPLE and Always Warm. The mean gain is 0.010, with an episode-bootstrap 95\% interval of [0.001, 0.021]. Five of the six differences are positive (two-sided exact sign test, $p=0.219$).

\paragraph{Earlier candidates remain useful after renaming or adding tasks.}
Four additional tests rename entities or add low-demand tasks while retaining existing resources. Both actions start from the same saved population with three search seeds. MAPLE selects Warm in all four cases. Adding tasks gives Warm an HV gain over Full of 0.708 in routing and 0.022 in cloud placement (Table~\ref{tab:main-reuse-discriminator}), using the original evaluation rules and separate pooled references per case.

A rule based on the fraction of changed table cells matches MAPLE on the main Pareto sequences but selects Full in all four added tests. A numerical rule re-evaluates old candidates: using old identifiers selects Full in the renaming tests, while mapping identifiers first makes all six decisions Warm (two tasks, three seeds). Appendix~\ref{app:component-details} gives the full comparisons.

\FloatBarrier
\section{Related Work}

\textbf{Natural-language optimization for domain practitioners.}
NL4Opt maps language to linear programs \citep{ramamonjison2023nl4opt}. Optimization agents generate formulations and solver code, using execution feedback for revision \citep{ahmadi2024optimus,zhang2025orllmagent,huang2025orlm,thind2025optimai}. MIRROR and ORPilot add problem elicitation and operational data handling \citep{shi2026mirror,xie2026orpilot}. MAPLE supports continued revision as operations change.

\textbf{Interactive and dynamic language agents.}
Agents retain dialogue and experience across interactions \citep{yao2023react}, with memory systems retrieving prior information \citep{packer2023memgpt,maharana2024locomo}. Flex-TravelPlanner introduces constraints over multiple turns \citep{oh2025flextravelplanner}; Gaia2 tests asynchronous events and actions that change the environment \citep{froger2026gaia2}. MAPLE retains executable optimization state, while NLDO evaluates continuity and solution quality across revisions.

\textbf{Multi-objective and dynamic numerical optimization.}
Dynamic optimization uses population repair, transfer, and restart under changing objectives and constraints \citep{branke2002dynamic,nguyen2012dynamic}. MOHOLLM uses LLM surrogate models and candidate samplers for hierarchical multi-objective search \citep{schwanke2026mohollm}. MAPLE revises executable problems through language, then runs established solvers such as NSGA-II \citep{deb2002nsga}.

\section{Conclusion}

MAPLE maintains executable optimization state across natural-language revisions. It completes all 15 NLDO trajectories, with online scalar quality of 0.951 and Pareto HV of 0.875. Controls show that TSS reduces invalid outputs and that some historical requests require both update records and accepted plans. Fresh initialization helps after disruptive routing changes; repaired candidates remain useful in renaming and task-addition controls.

\FloatBarrier
\label{sec:main-body-end}
			\linespread{1}\selectfont
			\paragraph{Ethics and reproducibility.}
\label{sec:ethics-reproducibility}
NLDO uses synthetic data.
The supplement provides tasks, prompts, settings, seed-level evidence, and offline scoring with references separate from agent inputs (Appendix~\ref{app:reproducibility}).

\bibliographystyle{iclr2026_conference}
	\bibliography{references}

	\appendix
	\section{Appendix}
\label{app:reader-guide}
\newcommand{\AppendixTableSetup}{%
  \fontsize{9pt}{11pt}\selectfont
  \setlength{\tabcolsep}{3.2pt}%
  \setlength{\extrarowheight}{1pt}%
  \renewcommand{\arraystretch}{1.18}%
  \arrayrulecolor{LiveOptTableRule}%
}
\let\ResultTableSetup\AppendixTableSetup
\let\DenseResultTableSetup\AppendixTableSetup
\let\MainResultTableSetup\AppendixTableSetup
\let\MainTextTableSetup\AppendixTableSetup
\renewcommand{\tabularxcolumn}[1]{m{#1}}
\newcolumntype{A}[1]{>{\raggedright\arraybackslash}m{#1}}
\newcolumntype{B}[1]{>{\centering\arraybackslash}m{#1}}
\raggedbottom
\setlength{\textfloatsep}{7pt plus 2pt minus 2pt}
\setlength{\floatsep}{7pt plus 2pt minus 2pt}
\setlength{\intextsep}{7pt plus 2pt minus 2pt}
\setlength{\abovecaptionskip}{3pt}
\setlength{\belowcaptionskip}{0pt}
\newcommand{\LiveOptSequentialRestartRows}{%
\methodours & 12/72 & 100.0\% & \LiveOptPublicUpdateHV & 162.8 & \textemdash{} \\
Always Warm & 0/72 & 100.0\% & \WarmPublicUpdateHV & 162.8 & $\PublicWarmDelta$ \\
Always Full & 72/72 & 100.0\% & \FullPublicUpdateHV & 199.0 & $\PublicFullDelta$ \\
}
\newcommand{\LiveOptReferenceSensitivityRows}{%
\method{} & +0.005 & +0.013 & +0.039 & 0.779 & 0.796 & 0.818 \\
\rowcolor{LiveOptBaselineRowB}
Always Warm & +0.005 & +0.013 & +0.039 & 0.769 & 0.786 & 0.808 \\
Always Full & +0.005 & +0.012 & +0.036 & 0.731 & 0.752 & 0.779 \\
\rowcolor{LiveOptBaselineRowB}
\method{} w/o TSS & +0.004 & +0.010 & +0.028 & 0.606 & 0.621 & 0.640 \\
}
\newcommand{\LiveOptUpdateStrataCompactRows}{%
All updates & 108 & 0.947 & All updates & 72 & 0.879 & 0.079 \\
Feasibility & 3 & 0.721 & Feasibility & 15 & 0.722 & 0.163 \\
Objective ordering & 63 & 0.942 & Objective ordering & 48 & 0.836 & 0.104 \\
Positive rescaling & 0 & -- & Positive rescaling & 12 & 0.940 & 0.044 \\
Diagnostic only & 6 & 1.000 & Diagnostic only & 12 & 0.988 & 0.010 \\
Unchanged & 39 & 0.948 & Unchanged & 0 & -- & -- \\
Historical reference & 21 & 0.996 & Historical reference & 9 & 0.894 & 0.070 \\
}
\newcommand{\LiveOptLandscapeRestartRows}{%
P010 & 23/100 & 20/20 & 0.729 & 0.752 & 0.724 \\
\rowcolor{LiveOptBaselineRowB}
P011 & 24/100 & 20/20 & 0.710 & 0.736 & 0.712 \\
P012 & 28/100 & 20/20 & 0.745 & 0.785 & 0.764 \\
\rowcolor{LiveOptBaselineRowB}
P013 & 10/100 & 20/20 & 0.855 & 0.854 & 0.853 \\
P014 & 20/100 & 20/20 & 0.768 & 0.763 & 0.761 \\
\rowcolor{LiveOptBaselineRowB}
P015 & 20/100 & 20/20 & 0.785 & 0.785 & 0.784 \\
All & 125/600 & 120/120 & 0.765 & 0.779 & 0.766 \\
}
\newcommand{\LiveOptArchiveCapSensitivityRows}{%
100 & 0.760 & 0.791 & 0.775 & 0.139 & 188.4 & 240/240 \\
\rowcolor{LiveOptBaselineRowB}
200 & 0.771 & 0.780 & 0.776 & 0.139 & 185.9 & 162/240 \\
500 & 0.763 & 0.785 & 0.774 & 0.138 & 183.1 & 0/240 \\
}
\newcommand{\PersistentReactFailureAttributionRows}{%
P007 & t03 & Missing output field & The code compiled and ran, but a required reasoning field was missing. Code-output repair did not check this field. \\
\rowcolor{LiveOptBaselineRowB}
P008 & t05 & Missing output field & The code compiled and ran, but a required reasoning field was missing. Code-output repair did not check this field. \\
P009 & t02 & Missing output field & The code compiled and ran, but a required reasoning field was missing. Code-output repair did not check this field. \\
}
\newcommand{\OutputProtocolSensitivityRows}{%
Persistent ReAct & 0.501 & 0.763 & 0.042 & 0.185 \\
ReAct & 0.333 & 0.376 & 0.023 & 0.091 \\
OptiMUS & 0.281 & 0.333 & 0.084 & 0.118 \\
ORLM & 0.120 & 0.298 & 0.000 & 0.021 \\
OptimAI & 0.077 & 0.103 & 0.051 & 0.059 \\
OR-LLM-Agent & 0.111 & 0.171 & 0.046 & 0.046 \\
}

\noindent\textbf{Appendix guide.}
The appendix first defines the mechanisms, metrics, and execution setup, then presents supplementary experiments. The Harness walkthrough follows the results. Prompt interfaces and the recorded benchmark requests appear last.

\begin{table}[!htbp]
\centering
\ResultTableSetup
\renewcommand{\arraystretch}{0.98}
\begin{tabularx}{\linewidth}{@{}A{0.30\linewidth}B{0.14\linewidth}Z@{}}
\toprule
\TableHead
\textbf{Topic} & \textbf{Go to} & \textbf{Evidence} \\
\midrule
Update execution & App.~\ref{app:method-details}--\ref{app:skill-catalog} & annotated algorithms, restart checks, and typed operators \\
\rowcolor{LiveOptBaselineRowB}
Metrics and experimental setup & App.~\ref{app:metric-definitions}--\ref{app:reproducibility} & metrics, task definitions, interfaces, and numerical settings \\
Supplementary results & App.~\ref{app:complete-stage-results}--\ref{app:reference-archive-sensitivity} & per-problem results, visual examples, controls, and sensitivity \\
\rowcolor{LiveOptBaselineRowB}
Application walkthrough & App.~\ref{app:harness-demo} & a complete request-to-update walkthrough \\
Recorded prompts and requests & App.~\ref{app:prompts-contracts}--\ref{app:complete-nldo-public-cases} & prompt interfaces, 15 initial requests, and 180 updates \\
\bottomrule
\end{tabularx}
\caption{Appendix contents and the corresponding evidence locations.}
\label{tab:appendix-reader-map}
\end{table}

\subsection{MAPLE Initialization and Update Procedure}
\label{app:architecture-details}
\label{app:method-details}

Algorithm~\ref{alg:liveopt-loop} shows how MAPLE builds the initial program and handles each later request.
LPD decides whether public data, the decision definition, or the evaluation function must change.
TSS checks the two editable Workbench functions.
The semantic reuse assessment reads the accepted public history and current change digest.
A fixed gate checks structural support for the proposed Warm--Full choice.
The fixed optimizer then runs an exact solver, scalar evolutionary search, or Pareto evolutionary search.
LSM stores events, the accepted state, and search archives with the public tables and Workbench.
For a request about an earlier assignment, LPD identifies the requested relation.
A fixed runtime procedure resolves the requested relation from stored events and accepted decisions, then enforces the binding during evaluation.

MAPLE w/o TSS builds its own representation, evaluation, search, and repair functions from the public \texttt{t00} task.
It carries that Workbench, accepted state, and search archives through the sequence.
The fixed optimization loop still performs selection, maintains the archive, records metrics, and checks stopping.
Table~\ref{tab:appendix-external-baseline-source-basis} specifies method inputs and carried state.

\paragraph{Algorithm notation.}
Let $x_0$ denote the initial request and $u_t$ the $t$-th update. $T$ is the number of updates.
The repair limit $B$ counts additional model attempts after a failed public check ($B=3$ in the reported runs).
The saved state is $S_t=(D_t,W_t,A_t,P_t)$. $E_t$ contains the accepted update history.
For Pareto tasks, $A_t$ contains an archive and its representative plan. Evolutionary search uses the candidate population $P_t$. The LP/MILP path does not use a population.
A tilde marks tentative tables or functions that have not yet replaced the accepted state.
The public-change summary $\Delta_t$ records differences between the accepted and tentative tables, decision declarations, and objectives.
A smoke test runs a small example from the available input data. It checks data access, function calls, and output structure before numerical search.

\begin{algorithm}[!htbp]
\caption{\method{} initialization and live update}
\label{alg:liveopt-loop}
\footnotesize
\setlength{\baselineskip}{10.5pt}
\begin{algorithmic}[1]
\Require initial request $x_0$, initial public tables $D_0$, updates $u_{1:T}$, repair limit $B$
\Ensure accepted states and event histories $(S_t,E_t)$, stopping at the first failed state
\State \textbf{LLM:} construct candidate Workbench $\widetilde W_0$ from $(x_0,D_0)$.
  \Comment{declare decisions and their evaluation}
\State Compile and test $\widetilde W_0$ on a small public example.
  \Comment{check signatures, types, and outputs}
\State On public errors, \textbf{LLM:} repair and repeat these checks, at most $B$ times.
\If{checks still fail} \State \Return failure. \Comment{no initial state is committed} \EndIf
\State Run the fixed solver; check execution and the public output contract.
\If{execution or output-contract checks fail} \State \Return failure. \Comment{no post-search LLM repair} \EndIf
\State Commit $S_0=(D_0,W_0,A_0,P_0)$ and start $E_0$.
  \Comment{save the accepted result and initial event}
\For{$t=1,\ldots,T$}
  \State \textbf{LLM (LPD):} identify edits and bindings required by $u_t$.
    \Comment{public state and history; no population}
  \State Resolve queries from stored events and accepted plans.
    \Comment{e.g., retrieve the customer's vehicle}
  \State Copy $(D_{t-1},W_{t-1})$ into $(\widetilde D_t,\widetilde W_t)$.
    \Comment{keep the committed state until acceptance}
  \State Apply supplied table edits; \textbf{LLM:} translate language-only edits if needed.
  \State Edit only the requested Workbench functions (Algorithm~\ref{alg:tss-surfaces}).
  \State Compile and smoke-test; on public errors, allow at most $B$ LLM repairs.
  \If{a check still fails} \State \Return failure. \Comment{retain $S_{t-1}$ and $E_{t-1}$} \EndIf
  \State Validate bindings in the revised decision domain.
    \Comment{check entities and assignments}
  \If{the solver uses an evolutionary population}
    \State Compare old and candidate states to form $\Delta_t$.
      \Comment{changed fields, decision types, objectives}
    \State \textbf{LLM:} assess reuse risk from $u_t$, public history, and $\Delta_t$.
    \State Check the response with Algorithm~\ref{alg:restart-gate}.
      \Comment{select Warm or Full before search}
    \State Build the initial population for the selected action.
      \Comment{Warm: at most half repaired; Full: all fresh}
  \EndIf
  \State Run the fixed solver and check the public output contract.
    \Comment{fixed LP/MILP, GA, or NSGA-II}
  \If{execution or output-contract checks fail} \State \Return failure. \Comment{retain $S_{t-1}$ and $E_{t-1}$} \EndIf
  \State Commit $S_t=(\widetilde D_t,\widetilde W_t,A_t,P_t)$; extend $E_{t-1}$ with the accepted event to obtain $E_t$.
\EndFor
\end{algorithmic}
\end{algorithm}

LLM calls are conditional on the requested data translation, function edit, or repair. Restart assessment uses the public update history and change digest before population construction. Held-out feasibility checks and reference scores are computed after execution.

\begin{algorithm}[!htbp]
\caption{Updating the two TSS Workbench functions}
\label{alg:tss-surfaces}
\footnotesize
\setlength{\baselineskip}{10.5pt}
\begin{algorithmic}[1]
\Require LPD edit labels, candidate public tables $\widetilde D_t$, accepted Workbench $W_{t-1}$
\Ensure checked candidate Workbench $\widetilde W_t$, or a public error for Algorithm~\ref{alg:liveopt-loop}
\State $\widetilde W_t\gets \operatorname{copy}(W_{t-1})$ \Comment{preserve functions outside the requested edit}
\If{the decision definition or public-data extraction changes}
  \State \textbf{LLM:} edit \texttt{build\_problem()}.
    \Comment{read tables; declare domains, solver, objectives}
\EndIf
\If{evaluation, constraints, decoding, or solution reporting changes}
  \State \textbf{LLM:} edit \texttt{evaluate()}.
    \Comment{decode one genome; compute violations and values}
\EndIf
\State Bind declared types to their fixed operators.
  \Comment{initialization, crossover, mutation, domain repair}
\State Compile; check solver choice, objective order, and required outputs.
\State Execute a small example using $\widetilde D_t$.
  \Comment{exercise data access, decoding, and evaluation}
\If{any public check fails}
  \State \Return the public error to Algorithm~\ref{alg:liveopt-loop}.
    \Comment{the outer loop owns the bounded repair budget}
\EndIf
\State \Return $\widetilde W_t$. \Comment{LP/MILP, GA, and NSGA-II routines remain fixed}
\end{algorithmic}
\end{algorithm}

\subsection{Checks for Restart Recommendations}
\label{app:semantic-restart-gate}

The model proposes whether to reuse earlier candidates. The fixed checks determine whether the response meets the conditions for a fresh start.

\paragraph{Parsing the model response.}
\label{app:restart-gate-definition}
The parser lowercases the risk string and trims its whitespace. It removes unrecognized mechanisms and duplicate list entries. It also strips whitespace and outer slashes from cited field paths. A cited path is retained only if it belongs to the public change summary's changed-field set. The five allowed identifiers are \texttt{decision\_support\_replacement}, \texttt{objective\_preference\_reversal}, \texttt{constraint\_regime\_change}, \texttt{resource\_role\_reversal}, and \texttt{distant\_basin\_risk}. Their checks are listed below.

\begin{table}[!htbp]
\centering
\MainResultTableSetup
\begin{tabularx}{\linewidth}{@{}A{0.28\linewidth}Z@{}}
\toprule
\textbf{Mechanism} & \textbf{Required structural change in the public summary} \\
\midrule
Decision-support replacement & Changed typed segments; or active-ID Jaccard overlap at most 0.5; or row-ID overlap at most 0.5 in a table with at least four old and four new rows. \\
Objective-preference reversal & Both a preference path and a policy path change; a changed evaluation function can supply the policy path. \\
Constraint-regime change & Changed typed segments, active-ID overlap at most 0.5, or at least two changed paths ending in a recognized constraint field. \\
Resource-role reversal & At least two changed resource-table paths ending in a recognized capacity, compatibility, emissions, or energy field. \\
Distant-basin risk & At least one of the preceding four structural tests passes. \\
\bottomrule
\end{tabularx}
\caption{Structural support tests used by the fixed restart gate. The model must also report high risk, cast a Full vote, and cite at least one changed path.}
\label{tab:restart-mechanism-checks}
\end{table}

Recognized constraint fields are \texttt{active}, \texttt{available}, \texttt{availability}, \texttt{capacity}, \texttt{cpu}, \texttt{mem}, \texttt{gpu}, \texttt{gpu\_required}, \texttt{max\_shifts}, \texttt{required}, \texttt{eligibility}, \texttt{eligible}, and \texttt{shift\_end}.
Resource tables are \texttt{machines}, \texttt{vehicles}, \texttt{nurses}, \texttt{staff}, and \texttt{resources}; their recognized fields are \texttt{capacity}, \texttt{cpu}, \texttt{mem}, \texttt{gpu}, \texttt{emission\_rate}, \texttt{energy\_idle}, \texttt{energy\_per\_cpu}, and \texttt{max\_shifts}.
Preference paths contain \texttt{/preferences/} or \texttt{/preference}. Policy paths contain \texttt{/policy/} or equal \texttt{workbench/fitness}.

The path and mechanism tests are separate. The \texttt{reason} string is logged and does not enter the gate. The prompt requests a JSON boolean for \texttt{full\_vote}. The implementation applies Python boolean conversion. Missing fields have defaults of unknown risk, empty lists, and a false vote. Ordinary parsing or provider errors return Warm. A detected quota-limit error is passed back to the caller.

\subsection{TSS Workbench and Available Choices}
\label{app:skill-catalog}

\begingroup\interlinepenalty=10000
The TSS Workbench is a reusable optimization program.
A small catalog describes the decision types and solvers available to the Workbench-building model, together with the fixed checks and restart actions available to the runtime.
The model edits only the decision-definition function and the evaluation function.
The first reads public tables, declares typed decisions, chooses the solver, and names the outputs.
The second checks one candidate against the public constraints and objectives.
Table~\ref{tab:appendix-skill-catalog} summarizes this boundary.
\par\endgroup

\begingroup\interlinepenalty=10000
\paragraph{How linear problems are sent to the solver.}
The Workbench specifies variables, bounds, the objective, and linear constraints for the fixed solver. Each variable has a name and a continuous or integer domain. The objective declaration includes coefficients and a direction. Each constraint supplies a coefficient row, a relation, and a right-hand side.
The fixed backend builds the coefficient arrays and uses HiGHS through SciPy's linear or mixed-integer solver.
A declared extraction rule maps returned values to selected entities, assignments, or schedules.
The runtime validates that output and retains the solver status and objective.
This numerical formulation is produced within the same Workbench construction call as the data access code.
\par\endgroup

\begin{figure}[!t]
\centering
\begin{minipage}[t]{0.31\linewidth}
\textbf{Public-data change: P015-t02}\\[-2pt]
\begin{Verbatim}[fontsize=\scriptsize]
tables/machines[id=M04]
- energy_per_cpu: 0.23
+ energy_per_cpu: 0.285
\end{Verbatim}
\vspace{-4pt}
The remaining public fields and both editable functions are unchanged.
\end{minipage}\hfill
\begin{minipage}[t]{0.66\linewidth}
\textbf{Evaluation-function change: P010-t02}\\[-2pt]
\begin{Verbatim}[fontsize=\scriptsize]
--- Before update
+++ After update
@@ congestion anchor @@
+    congestion_anchor_set = set()
+    for oid, o in data["order_by_id"].items():
+        if abs(o["x"] - 48) < 1e-9 and abs(o["y"] - 11) < 1e-9:
+            congestion_anchor_set.add(oid)
+            break   # at most one anchor
@@ route traversal @@
+        prev_oid = None   # no order before first stop
+            if prev_oid in congestion_anchor_set or oid in congestion_anchor_set:
+                leg_dist *= 1.25
+            prev_oid = oid
@@ return-to-depot leg @@
+        if prev_oid in congestion_anchor_set:
+            leg_dist *= 1.25
\end{Verbatim}
\vspace{-4pt}
Excerpt from the saved patch; unrelated lines are omitted.
\end{minipage}
\caption{\textbf{Two localized Workbench edits.} The left patch changes public data. The right patch changes route evaluation using a coordinate-based congestion anchor and a distance multiplier. Appendix~\ref{app:update-semantics} examines the historical binding and the travel-time calculation required by the public request.}
\label{fig:appendix-real-patch-diff}
\end{figure}

\paragraph{Adding a decision type or solver.}
Each catalog entry specifies its inputs, outputs, use conditions, and possible failures.
An entry can call an existing solver or restart policy.
Adding a decision type requires implementing its initialization, variation, and repair operators and registering them in the catalog.

\begin{table}[!t]
\centering
\MainResultTableSetup
\begin{tabularx}{\linewidth}{@{}l>{\hsize=1.15\hsize}Z>{\hsize=0.85\hsize}Z@{}}
\toprule
\TableHead
\textbf{Choice} & \textbf{Available options} & \textbf{When code must be added} \\
\midrule
Decision types & binary, categorical, integer, real, permutation, assignment, composition & a new decision type \\
\rowcolor{LiveOptBaselineRowB}
Solvers & LP/MILP, scalar EA, Pareto EA & a new solver \\
Population start & Warm (repaired history plus fresh candidates) or Full (fresh candidates) & a new restart action \\
\rowcolor{LiveOptBaselineRowB}
Checks & compilation, output fields, feasibility, and archive checks & a new safety rule \\
\bottomrule
\end{tabularx}
\caption{Choices exposed by the TSS runtime. The Workbench-building model selects decision types and a solver route. Fixed code supplies checks and operators, while the separate semantic assessment proposes Warm or Full after an update.}
\label{tab:appendix-skill-catalog}
\end{table}

TSS rejects a decision space that cannot be expressed by its declared decision types.
Solver selection preserves the public objective names and cannot turn a Pareto task into a scalar one.
Warm migrates historical candidates into the current decision domains and re-evaluates their feasibility.
It deduplicates the repaired seeds and fills the remaining population slots with fresh candidates.

\paragraph{How old candidates are repaired before search.}
Warm first adapts earlier candidates to the updated decisions and checks their feasibility. These candidates come from the previous population and archive.
For multiple objectives, the seed ordering includes objective extremes, knee candidates, and nondominated selection with crowding.
Each candidate is first migrated to the current domains, then evaluated.
The local search tests type-specific moves and accepts the first improvement in the pair
\((\mathbb{1}[\text{infeasible}], f_{\mathrm{scalar}})\).
It stops after a sweep with no improvement or 80 evaluated proposals.
The scalar fitness includes the generated penalty, whereas the subsequent NSGA-II loop retains the separate objective vector.
After deduplication, seed feasibility is checked again for reporting. The returned seeds enter feasibility-first search, and the remaining slots are sampled afresh.

With population 200 and at most 100 historical candidates, Warm uses up to 8,200 additional evaluation calls before the shared evolutionary loop. The loop itself uses $N(g+1)$ calls for population size $N$ and $g$ completed generations. Stage wall time also includes initialization, repair, and validation.

\subsection{Limitations and Ethics}
\label{app:limitations-ethics}

MAPLE relies on a fixed catalog of decision types, solver routes, and restart actions. Tasks requiring a new decision type or numerical operator need a runtime extension. Typed operators preserve representation membership, but feasibility still depends on the generated evaluation function.

LSM grounds references in retained events and accepted results. The agent needs clarification when a request could refer to several earlier entities or leaves a constraint unspecified. The semantic restart gate similarly depends on the model's interpretation of the update.

The evolutionary routes optimize within finite budgets. Repairing retained candidates adds work before search, so the choice between reuse and fresh initialization also depends on the available latency.

The current implementation runs generated Python functions in the application process. Operational use requires stronger isolation of files, imports, credentials, and network access. MAPLE is intended for batch or human-supervised re-optimization. High-impact decisions should support inspection, human approval, and rollback.

\subsection{Metric Definitions}
\label{app:metric-definitions}

NLDO evaluates every state with a fixed hidden evaluator.
For scalar profiles, let $f_t(x)$ be the evaluation objective and let $f_t^\star$ be the exact or best-known reference after sign conversion.
The normalized scalar quality is
\[
Q_t^{\mathrm{scalar}}
= \mathbf{1}[\mathrm{feasible}]
\max\!\left(0,
1 - \frac{\max(0, f_t(x)-f_t^\star)}
{\max(|f_t(x)|, |f_t^\star|, 1)}
\right).
\]
An infeasible output receives zero.
An output that matches the reference receives one.
Scalar tables use bounded 0--1 quality.
Pareto tables use the raw ratio below.
Dynamic tables separate \emph{solve rate} from \emph{online quality} using the quantities in Section~\ref{sec:task}.
The recorded indicator $a_t$ includes the original output-protocol and validity decisions, and $r_t$ marks their consecutive accepted prefix.
The new mathematical check $v_t$ requires at least one feasible plan in the saved output.
Solve rate averages $r_tv_t$, and online quality averages $r_tq_t$.
An output may therefore be mathematically feasible but ineligible because its original protocol failed or an earlier state was rejected.
Unattempted states have $v_t=q_t=0$ and remain in the denominator.
The archive scorer computes mathematical validity, while the trajectory aggregation applies recorded eligibility.
For MAPLE, we average quality across ten search seeds at each state, then across states. The reported numerical variation is computed from the seed means.
The DS denominator is \(9\) scalar episodes \(\times\,13\) sequential states and the DM denominator is \(6\) Pareto episodes \(\times\,13\) sequential states.
Analyses restricted to updates have denominators 108 and 72, respectively, and are explicitly labelled t01--t12.
Online quality is the mean quality over those same states, with infeasible and unsolved states contributing zero.

For multi-objective profiles, the evaluator compares a submitted archive $A_t$ with a withheld reference archive $R_t$ after removing infeasible candidates.
For every multi-objective episode and time step, \(R_t\) is built from 10 hidden reference runs.
Each run uses population 500 and 500 generations.
We pool all ten final populations, recheck feasibility, and keep the nondominated set without truncating the merged front.
Each time step therefore starts from 5,000 reference candidates before filtering and deduplication.
All 78 Pareto-state references are searched afresh under the public objective definitions, including the disruptive updates.
This construction is independent of any submitted method.
It uses the same public state as the submitted methods but a larger search budget.
The reference search has no early stopping and starts independently at each state.
We retain each seed's final population, random seed, completed generation count, and checksums.
All objectives are minimized.
Each time step uses fixed ideal and reference points derived from its held-out archive.
The submitted and reference archives use the same points.
For objective $i$, let $\ell_{t,i}$ and $h_{t,i}$ be its minimum and maximum over $R_t$.
The evaluation reference bound is the larger of its stored bound and
\[
 b_{t,i}=h_{t,i}+0.5\max(h_{t,i}-\ell_{t,i},|h_{t,i}|,1).
\]
The implementation adds $10^{-9}$ before rounding the bound to six decimal places.
Both archives use coordinates
\[
 \hat f_{t,i}=\operatorname{clip}_{[0,1.5]}
 \left(\frac{f_{t,i}-\ell_{t,i}}{\max(b_{t,i}-\ell_{t,i},10^{-9})}\right).
\]
The bounds are shared by every method at that stage and do not change as its search proceeds.
An objective value better than the ideal point is clipped to that point.
A value worse than the reference bound contributes no volume inside the unit HV box.
IGD uses these same clipped coordinates, including the upper cap of 1.5.
The benchmark has two or three objectives, for which the runtime computes HV exactly in this normalized box.
We report hypervolume ratio
\[
\mathrm{HVRatio}_t = \frac{\mathrm{HV}(A_t)}{\mathrm{HV}(R_t)}
\]
without clipping.
Because $R_t$ is a finite best-known set, a feasible archive may slightly exceed a ratio of one.
The evaluator removes infeasible candidates before computing HV and IGD. An archive with no feasible candidate receives zero HV ratio. IGD is reported only when an eligible archive contains at least one feasible candidate.
We also report inverted generational distance
\[
\mathrm{IGD}_t =
\frac{1}{|R_t|}
\sum_{y \in R_t}\min_{x \in A_t}\lVert \hat{x}-\hat{y}\rVert_2.
\]
We also report ideal gap, for which smaller is better.
It measures the distance from \(A_t\) to an ideal point placed 10\% beyond the best reference value on each objective.
Tables report raw HV ratio, IGD, and ideal gap.
The primary saved-plan evaluator rechecks all submitted candidates without imposing a new archive-size cap.
Dominated or duplicate points add no HV.
Any archive cap used while producing a method's saved output belongs to that method's search, not to the pooled reference construction.

\subsection{Benchmark and Evaluation Details}
\label{app:benchmark-details}

NLDO has five profiles with three episodes each, spanning established selection, scheduling, rostering, routing, and cloud-placement settings \citep{beasley1990orlib,brandimarte1993fjsp,ceschia2018inrc2,pillac2013dynamicvrp,erdogan2012greenvrp,bektas2011pollution,beloglazov2012energy}. Compact scalar tasks use exact references, larger scalar tasks use best-known solutions, and Pareto tasks use held-out archives.

\begingroup\interlinepenalty=10000
\paragraph{Effects of the requested updates.}
\label{app:update-semantics}
We label updates by changes to feasibility, candidate rankings, objective scale, or reported diagnostics. A changed table does not always imply changed optimal decisions. Multiplying one objective by a candidate-independent positive constant preserves Pareto dominance, while changing a coefficient for only one machine or vehicle can reorder candidates. A deadline used only in a reported diagnostic has a different role from a hard return deadline. Historical reference is a separate annotation.
\par\endgroup

\begin{table}[!htbp]
\centering
\ResultTableSetup
\renewcommand{\arraystretch}{0.98}
\begin{tabularx}{\linewidth}{@{}clllcZ@{}}
\toprule
\TableHead
\textbf{View} & \textbf{Setting} & \textbf{Objectives} & \textbf{Episodes} & \textbf{Time} & \textbf{Domains} \\
\midrule
SS & Static & Single-objective & P001--P009 & t00 & selection/allocation, precedence scheduling, coverage rostering \\
\rowcolor{LiveOptBaselineRowB}
SM & Static & Multi-objective & P010--P015 & t00 & ordered service routing, cloud-resource placement \\
DS & Dynamic & Single-objective & P001--P009 & t00--t12 & selection/allocation, precedence scheduling, coverage rostering \\
\rowcolor{LiveOptBaselineRowB}
DM & Dynamic & Multi-objective & P010--P015 & t00--t12 & ordered service routing, cloud-resource placement \\
\bottomrule
\end{tabularx}
\caption{The four NLDO views and their problem, time, and domain coverage. Dynamic views include initialization. Update-only analyses exclude t00 and are labelled separately.}
\label{app:nldo-part-map}
\end{table}

\begin{figure}[!htbp]
\centering
\includegraphics[width=\linewidth]{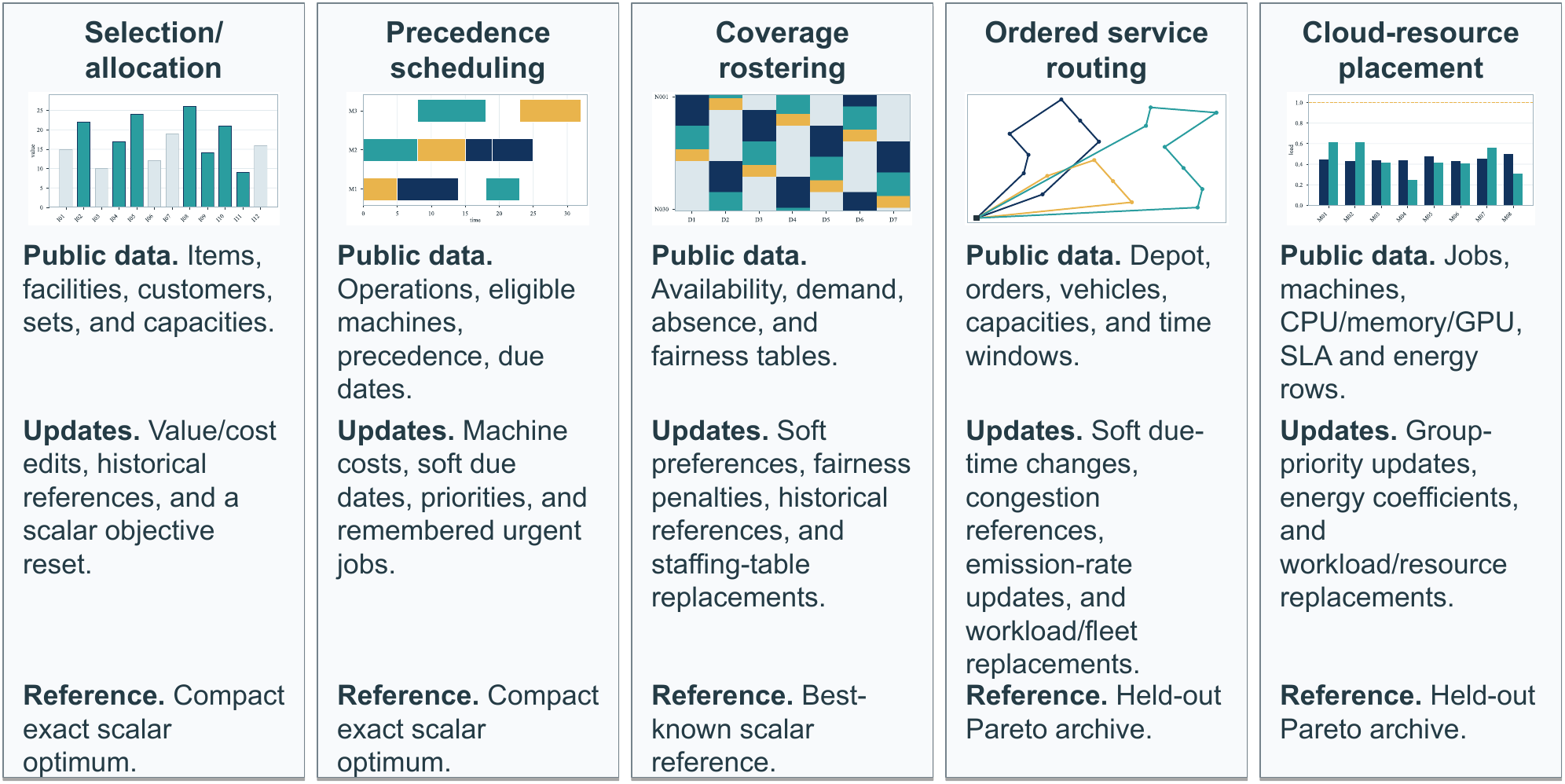}
\caption{\textbf{Five NLDO problem families.} Each card pairs a representative MAPLE solution with the public data, update types, and reference used for that family. The tasks span compact exact problems, larger scalar search, and Pareto routing and placement.}
\label{fig:benchmark-profile-cards}
\label{tab:benchmark-visual-cards}
\end{figure}
Each profile contributes 36 updates. Thirty of the 180 updates require information from earlier requests. Across all updates, 103 first paragraphs have distinct wording. Rostering, routing, and cloud placement contribute six disruptive t11--t12 states each. All table changes retain the initial columns and listed decision entities.

\paragraph{Agent inputs and evaluation annotations.}
The agent receives the problem description and input tables. Separate annotations identify the intended entities and reference chains for evaluation. Table~\ref{tab:appendix-external-baseline-source-basis} defines method inputs and retained information.

\paragraph{How saved routes and assignments are scored.}
\label{app:public-formula-evaluation}
We recompute the objectives of saved plans using the formulas stated below.

Routing travel time uses distance divided by speed. Reducing speed by 25\% multiplies the affected travel time by $4/3$, leaving physical distance unchanged.
Emissions sum route distance times vehicle emission rate and the carbon multiplier.
The public definition applies congestion only on arrival at the named order, with its identity retained across later edits to that order.
Every nonempty route leaves the current depot at time zero and returns by the vehicle's shift end. Waiting until an order's ready time and its service duration both count toward elapsed time. Lateness measures service-start time beyond the soft due time.
A requested reduction of a soft due target applies to that target directly. It can precede the ready time, in which case the earliest service already incurs lateness.
For example, reducing a soft due target of 87 by 50 gives 37. With a ready time of 40 and a revised soft due time of 37, the earliest service incurs three units of lateness.

\begingroup\interlinepenalty=10000
Cloud energy sums one idle charge per used machine and the assigned CPU demand times that machine's per-CPU energy rate, scaled by energy price and carbon intensity.
The~hidden evaluator measures imbalance as
\[
 B = 100\sum_{m\in\mathcal{M}_{\mathrm{avail}}}|u_m-\bar u|,
 \qquad
 u_m = \frac{\sum_{j:a(j)=m}c_j}{C_m},
 \qquad
 \bar u = \frac{\sum_{m\in\mathcal{M}_{\mathrm{avail}}}u_m}
 {|\mathcal{M}_{\mathrm{avail}}|}.
\]
Here $c_j$ is an active job's CPU demand and $C_m$ is machine capacity. Available but unused machines contribute zero utilization to the mean.
\par\endgroup

\paragraph{Execution and scoring versions.}
\label{app:execution-scoring-versions}
The original runs searched with their generated Workbench functions. The primary tables re-score the saved plans with the public formulas stated above.
The original hidden routing evaluator multiplies incoming distance, time, and emissions by 1.25 and adds an undeclared load-dependent emission factor.
The saved generated patch identifies the affected customer by coordinates and uses a different affected-leg scope (Figure~\ref{fig:appendix-real-patch-diff}).
The original hidden cloud evaluator accumulates an energy charge after each assignment, making energy depend on dictionary order.
For two jobs with CPU demands 2 and 4, idle energy 5, and per-CPU rate 2, this rule gives 26 or 30 depending on order, while the public formula gives 17 before global scaling.
Energy is invariant to the order of keys in the returned assignment map.
The original wording allowed variance or absolute deviation. The release now states this absolute-deviation formula explicitly in the initial request and every update.
We re-score saved routes and assignments using the public formulas above and construct new common references under those formulas.
The main agent comparison, the complete TSS ablation, and the full-sequence restart comparison use the public-formula evaluation. The single-update restart, search-operator, independent-model, and reference-sensitivity controls use the evaluation functions recorded in their original runs.
We also checked all saved external Pareto outputs against the formal reference.
Persistent ReAct's old scores used embedded constructive reference sets on 30 feasible recorded rows, nine of which also retained objective vectors inconsistent with re-evaluation of the saved assignment maps.
The primary tables recompute objectives from plans and use the common formal reference pool.
Across 4,368 Pareto state--seed records, every originally eligible output remains mathematically feasible after re-evaluation.
Sixteen additional saved outputs are feasible but outside their recorded strict prefix and receive zero strict-prefix quality.

\subsection{Static Solving Setup}
\label{app:static-solving-supplement}

The static supplement uses NLP4LP-hard and BWOR20 to measure ordinary one-shot optimization from language.
These tasks contain no updates, memory, restart choice, or predecessor population.
Each row provides the problem text and public parameters.
Each method produces executable Python solver code and follows its published process when available.
\paragraph{Evaluation of initial solutions.}
We report whether each program runs and whether its returned objective matches the reference. Table~\ref{tab:main-static-results} reports execution rate and Objective, the fraction of all cases whose returned objective agrees with the reference under the recorded numeric tolerance.
Objective agreement accepts alternative optimal decision vectors. The comparison accepts $|\hat f-f^\star|\leq\max(\epsilon_a,\epsilon_r|f^\star|)$, including equality at the threshold. NLP4LP-hard uses $\epsilon_a=\epsilon_r=10^{-6}$. The BWOR20 manifest overrides these defaults with $\epsilon_a=0.1$ and $\epsilon_r=0$. An output without a predicted objective fails the objective check. Status-only acceptance is recorded separately.
Both marginal rates are available for all seven methods, with denominators of 59 NLP4LP-hard cases and 20 BWOR20 cases.

\paragraph{Additional diagnostic checks.}
We separately report compilation, execution, objective agreement, and the original output checks. For case $i$, let $c_i$ and $e_i$ denote compilation and execution, $o_i$ reference-objective agreement, and $s_i$ correct reference status on a status-only case.
The joint Match diagnostic in Tables~\ref{tab:static-solving-nlp4lp}--\ref{tab:static-solving-bwor} is
\[
\frac{1}{N}\sum_{i=1}^{N} c_i e_i
\mathbf{1}\!\left[o_i\ \text{or}\ (\text{status-only case }i\ \text{and }s_i)\right].
\]
Its denominator includes every benchmark case, including execution failures.
The tables also retain compilation, execution, objective agreement, and the original workflow-specific output and named-variable checks.
These diagnostics can overlap. Dashes denote unreported supplementary checks.
Joint Match is unreported for ReAct and OptimAI.

\paragraph{Selection of the static benchmark cases.}
NLP4LP-hard and BWOR20 use fixed subsets of the released datasets. NLP4LP-hard comes from the NLP4LP dataset released with OptiMUS \citep{ahmadi2024optimus}. Our saved snapshot contains 361 instances, including 65 labelled hard. We retain the 59 hard instances with a nonempty reference solution. The other six have no value in the solution field. The reference objectives are read from each retained instance's supplied \texttt{solution} record.
BWOR20 is our fixed 20-case sample from the 82-row BWOR collection distributed with OR-LLM-Agent \citep{zhang2025orllmagent}. The manifest samples without replacement using Python's random generator with seed 20260621, then sorts the selected case identifiers. Its references use the supplied \texttt{answer} field.

\subsection{Baseline Interfaces and Execution}
\label{app:baseline-interfaces}

\begin{table}[!htbp]
\centering
\DenseResultTableSetup
\begin{tabularx}{\linewidth}{@{}A{0.15\linewidth}Z A{0.09\linewidth}A{0.11\linewidth}A{0.13\linewidth}A{0.13\linewidth}A{0.11\linewidth}@{}}
\toprule
\TableHead
\textbf{Method} & \textbf{Implementation} & \textbf{Public inputs} & \textbf{Code retained for the next request} & \textbf{Accepted plan} & \TightStack{\textbf{Cross-stage}\\\textbf{population/}\\\textbf{archive}} & \textbf{Repair feedback} \\
\midrule
MAPLE & Typed Workbench runtime & Full history & Two functions & Own representative & Own population and archive & Compi\-lation and small-example errors \\
w/o TSS & Untyped Workbench runtime & Full history & Full artifact & Own representative & Own population and archive & Compi\-lation and small-example errors \\
Persistent ReAct & Stateful executed-code control & Full history & Complete solver & Own compact answer and plan & None separate from answer & Compile, runtime, output \\
ReAct & Public-code adapter & Full history & In own dialogue, if present & Own, on continuation & Own, if returned on continuation & Compile, runtime, output \\
ORLM / OptiMUS / OR-LLM-Agent & Workflows based on the named methods & Full history & In own dialogue, if present & Own, on continuation & Own, if returned on continuation & Compile, runtime, output \\
OptimAI & Paper-based reproduction & Full history & In own dialogue, if present & Own, on continuation & Own, if returned on continuation & Compile, runtime, output \\
\bottomrule
\end{tabularx}
\caption{\textbf{Executed NLDO interfaces.} Cumulative inputs include the initial request, public tables, required outputs, and requests through the current stage. Each external workflow uses its own history, with at most three code-output repairs.}
\label{tab:appendix-external-baseline-source-basis}
\end{table}

\paragraph{Code and plans retained by Persistent ReAct.}
\label{app:persistent-react-execution}

Persistent ReAct retains its preceding solver code and accepted answer, then edits the program for the next request. It is our stateful control based on the ReAct pattern. At initialization, the model writes a complete Python solver program using public parameters and solver libraries. At an update it receives the initial public tables, cumulative dialogue, and its carried workspace, then emits the revised program, which applies the requested changes before solving.

The runtime parses, compiles, and executes the program, then reads its printed solution. Compilation, runtime, or solution-format failures return the actual error and available output to the model for at most three code-output repairs. The reasoning and tool-plan fields are auxiliary outputs. Missing auxiliary fields are checked separately and did not trigger code repair in the saved runs.

The workspace retains the complete preceding solver code and a compact final answer. A separate history field contains the preceding plan and cumulative updates. Public execution success with an extracted solution advances this workspace. Nothing is carried at t00. A program may produce multiple candidates and a Pareto set within a stage; the cross-stage workspace retains code and answers, with no separate population or Pareto archive.

\subsection{Numerical Settings and Released Materials}
\label{app:runtime-config}

Table~\ref{tab:appendix-runtime-config} lists the settings used in the experiments.
Main model calls share a provider and temperature, with output limits determined by the requested operation.
The table separates repair limits from evolutionary-search settings. Stopping rules and budgets were set on pilot material.

\begin{table}[!htbp]
\centering
\setlength{\intextsep}{4pt plus 1pt minus 1pt}%
\MainResultTableSetup
\begin{tabularx}{\linewidth}{@{}l>{\hsize=1.15\hsize}Z>{\hsize=0.85\hsize}Z@{}}
\toprule
\TableHead
\textbf{Setting} & \textbf{Value} & \textbf{Applies to} \\
\midrule
Language model & DeepSeek-V4-Pro (Preview), API ID \texttt{deepseek-v4-pro}, temperature 0.0 & all main model calls \\
LLM output limits & 1200--18000 tokens & depend on the requested task \\
Workbench edit repair & 3 attempts & \method{} Workbench checks \\
Workbench check & compilation and hard-coded-ID check & public table identifiers only \\
External-code repair & 3 attempts & external methods \\
Search budget & population 200, at most 200 generations, seeds 0--9 & main numerical runs \\
Search archives & at most 500 feasible, unique, nondominated points & main numerical runs \\
restart selector & semantic choice from stored events and the public change summary & no trial search or hidden score \\
Restart actions & Warm: up to 100 repaired, then fresh. Full: 200 fresh & same solver and search budget \\
Scalar stopping & start at g=40, gain $\epsilon=5{\times}10^{-4}$, patience 25 & feasible scalar objective \\
Pareto stopping & g=40, min. archive 8, $\epsilon=5{\times}10^{-4}$, box 0.01 & fixed internal objective box \\
Metric record & every generation, at most 200 saved points & recording only \\
Held-out references & fixed $500\times500$, 10-seed set per time step & final metrics only \\
Model repeats & 3 Workbench sequences $\times$ seeds 0--2 & P010 routing and P015 cloud, fixed restart choices \\
\bottomrule
\end{tabularx}
\caption{Settings for the reported experiments.}
\label{tab:appendix-runtime-config}
\end{table}
\begingroup\interlinepenalty=10000
\paragraph{Stopping rule.}
Main evolutionary runs stop after at most 200 generations.
From generation 40 onward, a run may stop after 25 generations without sufficient improvement.
A scalar run requires a relative improvement above $5\times10^{-4}$ in its best feasible objective.
A Pareto run becomes eligible only with at least eight feasible archive members.
The run's objective box is then fixed with a 5\% upper margin.
The 25-generation count restarts after an internal HV gain above $5\times10^{-4}$.
It also restarts after the front occupies a new normalized cell of width $0.01$ or improves the normalized ideal point by more than $5\times10^{-4}$.
Stopping uses these internal Workbench signals. Held-out HV and IGD are computed from saved candidates for final scoring and generation-wise plots.
\par\endgroup

\subsection{Reproducibility Materials}
\label{app:reproducibility}

The supplementary materials contain the public requests and tables, prompts, generated Workbench examples, numerical settings, and saved result records. The evaluation package provides versioned scoring code and state-specific references for scoring saved plans. Aggregation combines these scores with the recorded acceptance prefixes. Provider credentials are excluded from the packages.

The evaluation package contains 195 states, scalar references, and 78 Pareto references from ten $500\times500$ searches per state, with reference manifests and file checksums.
The primary aggregate uses 4,368 Pareto state--seed records, original acceptance indicators, and the retained static execution/objective summaries.
Supplementary control records include P007--P015 matched restarts, 180 update transitions, 30 historical references, state-binding cases, independent P010/P015 model runs, and reference/archive checks.
The time-step mapping and reference-objective audit checks agree with the saved records.

Episode-bootstrap intervals use 20,000 resamples (NumPy generator seed 20260905).

The original DeepSeek-V4-Pro (Preview) checkpoint is no longer available for reruns, so supplementary joint-Match measurements were not obtained for ReAct and OptimAI.

\FloatBarrier
\subsection{Aggregate and Per-Problem Results}
\label{app:complete-stage-results}
\label{app:per-problem-results}

\paragraph{Checking the information retained by Persistent ReAct.}
We inspect the saved inputs to verify that each update receives the preceding program and plan from the same run. The reported source is one original, unmerged branch selected by continuity of its carried state.
All 141 records include successful final compilation and execution. Six records use repair, with ten repair calls in total and a cap of three per state.
All 126 update inputs match the preceding publicly executable stage's solver code, compact final answer, and public plan.
The trace check also verifies cumulative dialogue, unchanged initial public tables, and the absence of another method's search state.
The evidence includes original responses, execution records, predecessor checksums, and per-state acceptance flags.
The strict evaluation combines the saved output-protocol checks with mathematical validity for every external method.
The mathematical-prefix diagnostic isolates the effect of auxiliary output fields.

\paragraph{Routing and cloud-placement results.}
Table~\ref{tab:public-formula-evaluation} separates results for the two Pareto domains.
\newcommand{\PublicEvaluationSummaryRows}{%
\rowcolor{LiveOptRowAccent}
\method{} & 0.768 & 0.982 & 0.875 & 0.082 \\
\rowcolor{LiveOptBaselineRowB}
Always Warm & 0.749 & 0.982 & 0.866 & 0.093 \\
Always Full & 0.665 & 0.984 & 0.825 & 0.112 \\
\rowcolor{LiveOptBaselineRowB}
w/o TSS & 0.394 & 0.936 & 0.665 & 0.143 \\
}

\begin{table}[htbp]
\centering
\ResultTableSetup
\renewcommand{\arraystretch}{1.10}
\begin{tabularx}{\linewidth}{@{}l*{4}{Y}@{}}
\toprule
\TableHead
\textbf{Method} & \textbf{Routing HV} & \textbf{Cloud HV} & \textbf{Overall HV} & \textbf{IGD $\downarrow$} \\
\midrule
\PublicEvaluationSummaryRows
\bottomrule
\end{tabularx}
\caption{\textbf{Public-formula evaluation of saved Pareto plans.} All rows cover t00--t12 and ten numerical seeds. Fixed policies follow their own histories. HV includes zero-valued invalid states. IGD is conditional on valid archives, with 780 states for each full-system policy and 630 for w/o TSS.}
\label{tab:public-formula-evaluation}
\end{table}

\newcommand{\LiveOptMOHVVarianceRows}{%
P010 & Ordered service & 0.754$\pm$0.076 & 0.154$\pm$0.051 \\
\rowcolor{LiveOptBaselineRowB}
P011 & Ordered service & 0.759$\pm$0.070 & 0.150$\pm$0.040 \\
P012 & Ordered service & 0.790$\pm$0.052 & 0.157$\pm$0.040 \\
\rowcolor{LiveOptBaselineRowB}
P013 & Cloud resources & 0.986$\pm$0.004 & 0.011$\pm$0.004 \\
P014 & Cloud resources & 0.972$\pm$0.003 & 0.013$\pm$0.001 \\
\rowcolor{LiveOptBaselineRowB}
P015 & Cloud resources & 0.989$\pm$0.003 & 0.010$\pm$0.001 \\
}
\begin{table}[!t]
\centering
\DenseResultTableSetup
\renewcommand{\arraystretch}{1.10}
\begin{tabularx}{\linewidth}{@{}ll*{2}{Y}@{}}
\toprule
\TableHead
\textbf{Episode} & \textbf{Profile} & \TightStack{\textbf{HV}\\\textbf{mean $\pm$ SD}} & \TightStack{\textbf{IGD}\\\textbf{mean $\pm$ SD}} \\
\midrule
\LiveOptMOHVVarianceRows
\bottomrule
\end{tabularx}
\caption{\textbf{Public-formula Pareto quality by episode.} Every row uses 10 seeds and 13 states per seed against the same freshly searched reference union. HV and IGD first average over time steps and then report variation across seeds.}
\label{tab:appendix-mo-hv-variance}
\end{table}

MAPLE's overall HV is \LiveOptPublicDMHV{}, compared with \WarmPublicDMHV{} for own-history Warm and \FullPublicDMHV{} for own-history Full.
The domain breakdown locates the gain: adaptive reuse helps routing, while Full is slightly stronger in cloud placement.
The untyped Workbench can find strong cloud plans but loses validity on routing.

The following summaries cover all 15 episodes with population 200, at most 200 generations per state, and 10 seeds.


\begin{table}[!htbp]
\centering
\DenseResultTableSetup
\fontsize{10pt}{12pt}\selectfont
\renewcommand{\arraystretch}{1.10}
\begin{tabularx}{\linewidth}{@{}l*{6}{Y}@{}}
\toprule
\TableHead
\multicolumn{3}{c}{\textbf{DS}} & \multicolumn{4}{c}{\textbf{DM}} \\
\cmidrule(lr){1-3}\cmidrule(lr){4-7}
\TableHead
\textbf{Update type} & \textbf{States} & \TightStack{\textbf{Online}\\\textbf{quality}} & \textbf{Update type} & \textbf{States} & \TightStack{\textbf{Online}\\\textbf{quality}} & \textbf{IGD $\downarrow$} \\
\midrule
\LiveOptUpdateStrataCompactRows
\bottomrule
\end{tabularx}
\caption{\textbf{MAPLE results by audited update semantics}, t01--t12. Ten-seed state means use the original scalar evaluation for DS and public formulas for DM. Labels may overlap; historical reference is a separate dimension. Diagnostic-only changes preserve objectives and hard constraints. Solve rate is 100\% in every nonempty group; empty groups have unavailable scores.}
\label{tab:appendix-update-strata}
\end{table}

\newcommand{\FormalBenchmarkResultRows}{%
Selection/allocation & \methodours & 0.957 & 0.953 & 19.231 & -- \\
\midrule
\rowcolor{LiveOptBaselineRowB}
Precedence scheduling & \methodours & 0.965 & 0.962 & 11.038 & -- \\
\midrule
Coverage rostering & \methodours & 0.933 & 0.927 & 298.920 & -- \\
\midrule
\rowcolor{LiveOptBaselineRowB}
Ordered service routing & \methodours & 0.752 & 0.757 & 0.757 & 0.165 \\
\midrule
Cloud-resource placement & \methodours & 0.805 & 0.801 & 0.801 & 0.129 \\
}
\newcommand{\LiveOptCurrentOverallQuality}{0.882}
\newcommand{\LiveOptCurrentDynamicQuality}{0.880}
\newcommand{\LiveOptCurrentHV}{0.779}
\newcommand{\LiveOptCurrentIGD}{0.147}
\newcommand{\LiveOptCurrentTokenMean}{0}
\begin{table}[!htbp]
\centering
\DenseResultTableSetup
\fontsize{10pt}{12pt}\selectfont
\renewcommand{\arraystretch}{1.10}
\begin{tabularx}{\linewidth}{@{}ll*{4}{Y}@{}}
\toprule
\TableHead
\textbf{NLDO profile} & \textbf{Method} & \TightStack{\textbf{Overall}\\\textbf{mean}} & \TightStack{\textbf{Dynamic}\\\textbf{mean}} & \textbf{Gap/HV} & \textbf{IGD $\downarrow$} \\
\midrule
\FormalBenchmarkResultRows
\bottomrule
\end{tabularx}
\caption{\textbf{Recorded-evaluator results by profile.} Overall includes t00. Dynamic covers t01--t12. Scalar rows report objective gap. Pareto rows report HV and IGD. Public-formula results are in Table~\ref{tab:appendix-mo-hv-variance}.}
\label{tab:formal-benchmark-results}
\label{tab:scalar-benchmark-results}
\label{tab:mo-benchmark-results}
\end{table}

\newcommand{\LiveOptCurrentEpisodeRows}{%
P001 & Selection/allocation & 1.000 & 0.000 & -- & -- \\
\rowcolor{LiveOptBaselineRowB}
P002 & Selection/allocation & 0.871 & 57.692 & -- & -- \\
P003 & Selection/allocation & 1.000 & 0.000 & -- & -- \\
\rowcolor{LiveOptBaselineRowB}
P004 & Precedence scheduling & 0.894 & 33.115 & -- & -- \\
P005 & Precedence scheduling & 1.000 & 0.000 & -- & -- \\
\rowcolor{LiveOptBaselineRowB}
P006 & Precedence scheduling & 1.000 & 0.000 & -- & -- \\
P007 & Coverage rostering & 0.930 & 452.043 & -- & -- \\
\rowcolor{LiveOptBaselineRowB}
P008 & Coverage rostering & 1.000 & -7.873 & -- & -- \\
P009 & Coverage rostering & 0.868 & 452.592 & -- & -- \\
\rowcolor{LiveOptBaselineRowB}
P010 & NLDO-DM, ordered service & 0.747 & 0.415 & 0.747 & 0.165 \\
P011 & NLDO-DM, ordered service & 0.733 & 0.417 & 0.733 & 0.187 \\
\rowcolor{LiveOptBaselineRowB}
P012 & NLDO-DM, ordered service & 0.777 & 0.407 & 0.777 & 0.157 \\
P013 & NLDO-DM, cloud resources & 0.855 & 0.277 & 0.855 & 0.109 \\
\rowcolor{LiveOptBaselineRowB}
P014 & NLDO-DM, cloud resources & 0.771 & 0.322 & 0.771 & 0.130 \\
P015 & NLDO-DM, cloud resources & 0.789 & 0.326 & 0.789 & 0.139 \\
}

\newcommand{\NLDOExternalContinuityQualityRows}{%
Persistent ReAct & 126/180 & 96.8\% & 58/180 & 3.87 & 0.469 & \PersistentPublicUpdateHV{} \\
\rowcolor{LiveOptBaselineRowB}
ReAct & 55/180 & 83.6\% & 31/180 & 2.07 & 0.287 & \ReactPublicUpdateHV{} \\
OptiMUS & 53/180 & 83.0\% & 31/180 & 2.07 & 0.231 & \OptimusPublicUpdateHV{} \\
\rowcolor{LiveOptBaselineRowB}
ORLM & 35/180 & 77.1\% & 5/180 & 0.33 & 0.046 & \ORLMPublicUpdateHV{} \\
OptimAI & 18/180 & 61.1\% & 9/180 & 0.60 & 0.037 & \OptimAIPublicUpdateHV{} \\
\rowcolor{LiveOptBaselineRowB}
OR-LLM-Agent & 26/180 & 65.4\% & 12/180 & 0.80 & 0.056 & \ORAgentPublicUpdateHV{} \\
}

\begin{table}[!ht]
\begingroup
\centering
\DenseResultTableSetup
\renewcommand{\arraystretch}{1.10}
\begin{tabularx}{\linewidth}{@{}ll*{4}{Y}@{}}
\toprule
\TableHead
\textbf{Episode} & \textbf{Profile} & \textbf{Mean} & \textbf{Gap} & \textbf{HV} & \textbf{IGD $\downarrow$} \\
\midrule
\LiveOptCurrentEpisodeRows
\bottomrule
\end{tabularx}
\caption{\textbf{Recorded-evaluator scores by episode.} Ten seeds, t00--t12, $200\times200$ maximum budget. Scalar rows give objective gap; Pareto rows give recorded HV, ideal gap, and IGD. Table~\ref{tab:appendix-mo-hv-variance} gives public-formula Pareto scores.}
\label{tab:appendix-liveopt-current-per-problem}
\endgroup
\par\vspace{7pt}
\begingroup
\centering
\ResultTableSetup
\renewcommand{\arraystretch}{1.10}
\begin{tabularx}{\linewidth}{@{}l*{6}{Y}@{}}
\toprule
\TableHead
\textbf{Method} & \TightStack{\textbf{Attempted}\\\textbf{updates}} & \TightStack{\textbf{Feas.$\mid$}\\\textbf{attempt}} & \TightStack{\textbf{Valid}\\\textbf{prefix}} & \TightStack{\textbf{Mean}\\\textbf{accepted}\\\textbf{update}\\\textbf{count}} & \TightStack{\textbf{DS}\\\textbf{quality}} & \TightStack{\textbf{DM}\\\textbf{HV}} \\
\midrule
\NLDOExternalContinuityQualityRows
\bottomrule
\end{tabularx}
\caption{\textbf{Update-only continuity and quality}, t01--t12. Feas.$\mid$attempt considers attempted updates. Mean accepted update count, $\mathrm{E}[\mathrm{horizon}]$, counts updates in the strict prefix. Prefix, mean count, and quality assign zero to failed suffixes. DM uses public formulas. MAPLE's DS quality and DM HV are 0.947 and \LiveOptPublicUpdateHV{}. Table~\ref{tab:main-dynamic-results} includes initialization.}
\label{tab:appendix-external-continuity}
\endgroup
\par\vspace{7pt}
\begingroup
\centering
\ResultTableSetup
\renewcommand{\arraystretch}{1.10}
\begin{tabularx}{\linewidth}{@{}lcA{0.19\linewidth}Z@{}}
\toprule
\TableHead
\textbf{Episode} & \textbf{First break} & \textbf{Direct cause} & \textbf{Observable evidence} \\
\midrule
\PersistentReactFailureAttributionRows
\bottomrule
\end{tabularx}
\caption{\textbf{First protocol failures in three Persistent ReAct rostering sequences.} Each saved plan is mathematically feasible at the first rejected state, but omits a required reasoning field.}
\label{tab:appendix-persistent-react-failure-attribution}
\endgroup
\end{table}



\paragraph{Effect of auxiliary output fields.}
We recompute the eligible prefix while ignoring auxiliary fields and keeping the saved plans and feasibility checks unchanged.
Table~\ref{tab:appendix-external-continuity} includes every saved attempt at t01--t12, including attempts after an earlier output-check failure. Persistent ReAct's initial P004 program runs, but its schedule is infeasible and its output lacks a required reasoning field. Its accepted prefix therefore has length zero. Table~\ref{tab:appendix-persistent-react-failure-attribution} lists three feasible rosters rejected because required auxiliary fields were missing. Those omissions did not trigger code repair.

\begingroup\interlinepenalty=10000
For every external method, the mathematical prefix $\tilde r_t=\prod_{j=0}^{t}v_j$ accepts feasible plans regardless of auxiliary fields, retaining the same plans, public formulas, references, and zero missing suffixes. Persistent ReAct continued running after some auxiliary fields were omitted. Under mathematical-prefix scoring, its scalar solve rate is 85.5\%, compared with 51.3\% under strict scoring. Its Pareto solve rate is 42.3\%, compared with 11.5\%.
\par\endgroup

\FloatBarrier
\clearpage
\subsection{Case Study and Visual Result Atlas}
\label{app:visual-atlas}
\label{app:case-study}

\begin{figure}[H]
\centering
\includegraphics[width=0.93\linewidth]{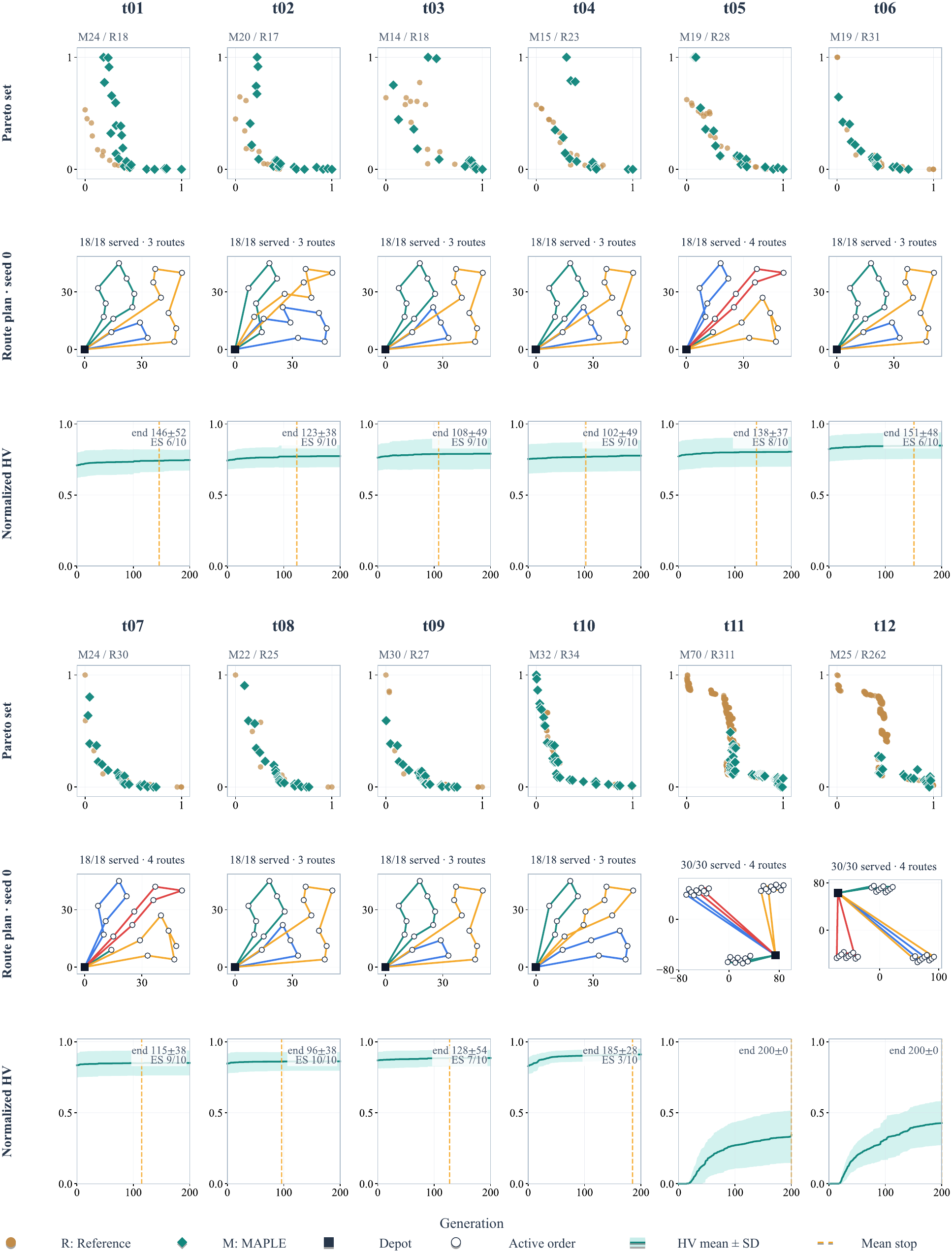}
\caption{\textbf{Ordered-service routing across t01--t12 (NLDO-P010).} The upper/lower blocks show t01--t06/t07--t12. Teal diamonds show MAPLE's ten-seed pooled Pareto set; amber circles show the recorded reference. Route panels show the seed-0 plan, zoomed to active orders; inactive orders are omitted. Front coordinates normalize distance and lateness within each stage. HV uses all three objectives, including emissions.}
\label{fig:p010-three-row-atlas}
\label{fig:appendix-green-stage-table-a}
\label{fig:appendix-green-stage-table-b}
\label{fig:appendix-green-stage-table-c}
\par\smallskip{\small\raggedright The third row follows normalized HV during search.
Generation 0 is the population just after restart.
Curves remain constant after termination. The dashed line marks the mean stopping generation.\par}
\end{figure}

\begin{figure}[H]
\centering
\includegraphics[width=0.97\linewidth]{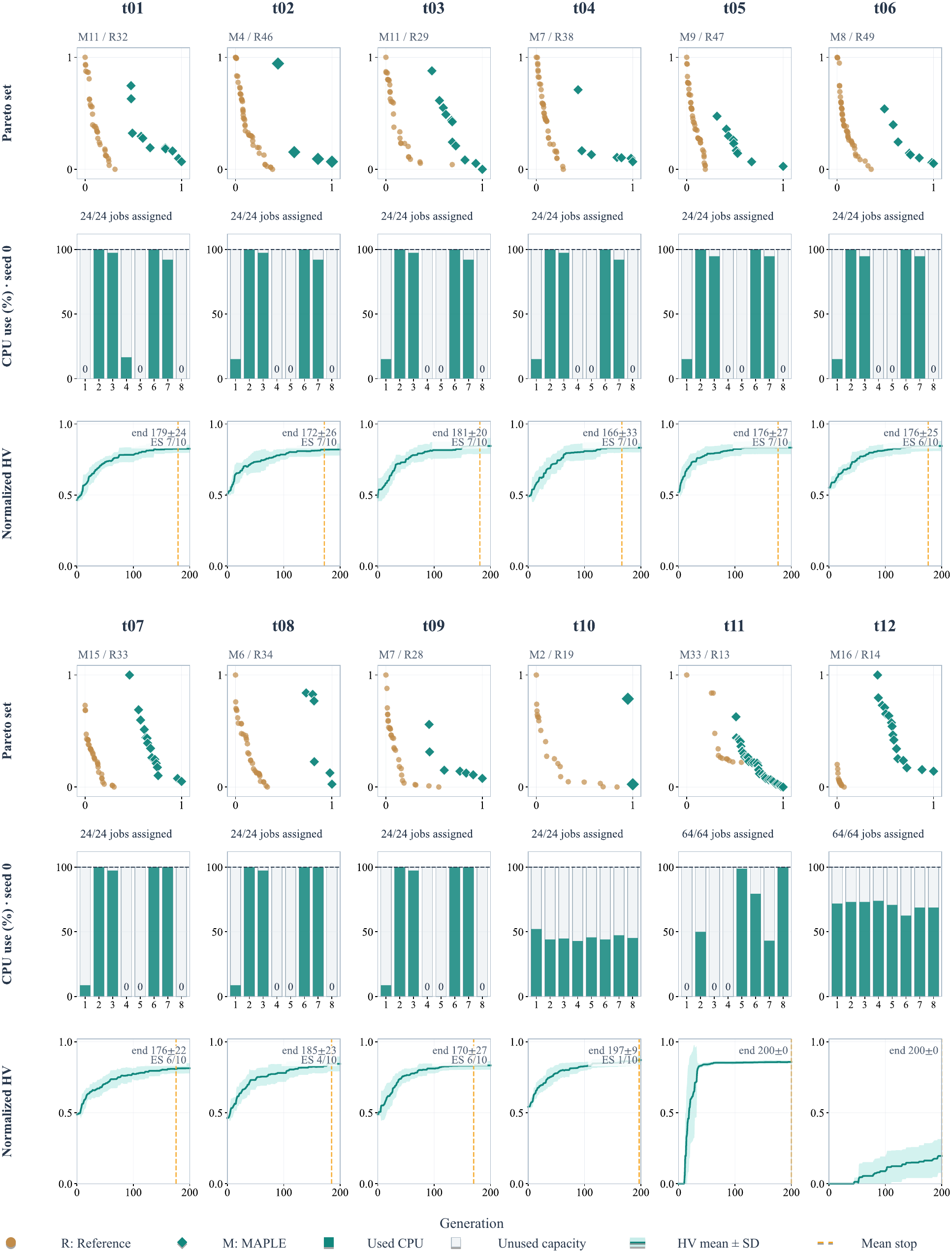}
\caption{\textbf{Cloud-resource placement across t01--t12 (NLDO-P015).} Blocks show t01--t06 and t07--t12. Diamonds show MAPLE's ten-seed pooled energy--load-imbalance fronts; circles show the reference, normalized within each stage. Seed-0 plans show CPU utilization on machines 1--8: teal marks usage, pale bars mark unused capacity, and 0 marks idle machines. Ten-seed HV curves show the initial quality loss and recovery at t11/t12.}
\label{fig:p015-case-study-snapshots}
\label{fig:p015-assignment-snapshots}
\label{fig:p015-real-population-snapshots}
\label{fig:p015-regime-reset-snapshots}
\par\smallskip{\small\raggedright \textbf{Accepted plans with lower solution quality.}
\label{app:low-quality-tail}
We inspect accepted plans whose normalized quality decreases after a revision. Accepted outputs pass the runtime's public checks, but their evaluated quality can still be low. At P002-t10, the scalar objective changes from monetary cost to service time. The retained assignment remains feasible and has lower normalized quality under the revised objective. At cloud update t12, the machine identifiers remain the same, but workloads and resource properties change. A feasible placement can have lower quality under the revised problem.\par}
\end{figure}

Figures~\ref{fig:p010-three-row-atlas} and~\ref{fig:p015-case-study-snapshots} show how accepted plans, Pareto fronts, and search traces change across successive routing and cloud-placement requests.

\FloatBarrier
\subsection{Component and Restart Evidence}
\label{app:component-details}

\paragraph{Agreement between predicted and saved edits.}
\label{app:transition-audit}
\begingroup\interlinepenalty=10000
We compare LPD's three edit decisions with the changes saved after each update. The saved decisions are those used at seed 0.
All ten numerical seeds share the same LLM-generated data and Workbench.
The analysis contains 180 language decisions.
LPD predicts changes to data, decision definition, and evaluation.
All three predictions match the accepted changes in 170 updates.
The separate counts are 174, 178, and 178.
The six data differences are explicit roster-table replacements at P007--P009 t11--t12.
In the other four cases, LPD requests a cautious code edit but the model returns the function unchanged.
Whenever a function changes, LPD has marked it for change.
\par\endgroup

Thirty benchmark updates refer to entities through earlier requests (Table~\ref{tab:appendix-referential-resolution}). LSM resolves all 30, including six requests that require following three reference links across the update history.

\begin{table}[!htbp]
\centering
\ResultTableSetup
\begin{tabularx}{\linewidth}{@{}*{5}{Y}@{}}
\toprule
\TableHead
\textbf{Reference links} & \textbf{Updates} & \textbf{Entity/field} & \textbf{Value} & \textbf{Valid result} \\
\midrule
\LiveOptMemoryChainRows
\bottomrule
\end{tabularx}
\caption{\textbf{Reference grounding with stored events.} None of the 30 current requests states the target entity ID. Entity/field and value compare the accepted public-data change with the benchmark's intended update. The intended answer is never shown to the language model.}
\label{tab:appendix-referential-resolution}
\end{table}

These requests test whether earlier update records identify the required entities. The following control also tests requests that need assignments from the accepted plan.

\paragraph{Information needed to interpret assignment requests.}
\label{app:state-binding-control}
We vary whether the model receives earlier update records, the accepted plan, both, or neither. We use saved routing states from P010 and cloud-placement states from P015 at t04, t08, and t10. The requests require three kinds of information. One kind identifies the requested assignment through earlier updates. Another names the relevant entities and asks to preserve assignments from the accepted plan. The third uses an earlier update to identify the entities and the accepted plan to retrieve their assignments. Each kind has three equivalent wordings in each domain, giving 18 language requests.

For example, the first routing request assigns the customer identified in a congestion note to the vehicle identified in an emissions update. The second names orders O001, O002, and O003 and preserves their saved vehicle assignments. The third preserves the saved vehicle assignment of the customer identified in the congestion note. Table~\ref{tab:appendix-state-binding-requests} contains the recorded request texts.

A fifth condition supplies the required assignment through a fixed procedure. The required answers are stored separately for evaluation.

\begin{table}[!htbp]
\centering
\MainResultTableSetup
\begin{tabularx}{\linewidth}{@{}lA{0.19\linewidth}Z@{}}
\toprule
\TableHead
\textbf{Profile} & \textbf{Information needed} & \textbf{Public request q0} \\
\midrule
P010 & Earlier update records & For the next dispatch window, the customer previously marked as the congestion anchor must be served by the vehicle whose telemetry update changed its emission rate. Treat this as a hard assignment constraint. All other feasibility rules and Pareto objectives stay unchanged. \\
\rowcolor{LiveOptBaselineRowB}
P010 & Accepted plan & For the next dispatch, keep the currently accepted vehicle assignments of orders \texttt{O001}, \texttt{O002}, and \texttt{O003}. Other orders may move, and the original feasibility rules and Pareto objectives remain unchanged. \\
P010 & Both records & Preserve the currently accepted vehicle assignment of the customer referred to as the congestion anchor. Other orders may be reassigned, and the original objectives remain unchanged. \\
\midrule
P015 & Earlier update records & For the next placement window, assign the first job named when workload group \texttt{burst-2} was introduced to machine \texttt{M07}. This is a hard assignment constraint. All other feasibility rules and Pareto objectives stay unchanged. \\
\rowcolor{LiveOptBaselineRowB}
P015 & Accepted plan & For the next placement, keep the currently accepted machine assignments of jobs \texttt{J012}, \texttt{J013}, and \texttt{J014}. Other jobs may move, and the original feasibility rules and Pareto objectives remain unchanged. \\
P015 & Both records & Keep the current accepted machine assignment of every job in the workload group called \texttt{burst-2}. Other jobs may move, and the original objectives remain unchanged. \\
\bottomrule
\end{tabularx}
\caption{\textbf{Requests that use different stored information.} The released cases include all three wordings and identifiers for their source states.}
\label{tab:appendix-state-binding-requests}
\end{table}

\paragraph{Saved plans used in the language tests and search replay.}
The original language tests and the search replay use representatives selected by different rules. The language tests read a representative reconstructed from each seed's saved archive. They select the plan with the lowest recorded scalar fitness and break ties by archive order. The original online run selects its representative by a different rule. It stores the first member of the final NSGA-II population, ordered by nondomination rank, decreasing crowding distance, and violation score, with stable ties.

For the search replay, we use the saved query from the first wording and read assignments from the originally exported online plan. We run six cases with three search seeds under each of the five information conditions. This gives 18 numerical results per condition and 90 results in total. Each search uses a population of 200 and at most 200 generations. The public tables, Workbench, Warm choice, starting population, and search budget are held fixed across conditions. The requested assignment is imposed after the restart population is built.

\FloatBarrier
\begingroup\interlinepenalty=10000
\paragraph{Identifying and enforcing the requested assignment.}
We evaluate assignment identification in the language tests and assignment enforcement in a separate search replay. Full LSM resolves all 18 requests. With only earlier update records, the model resolves the six requests answerable from those records. With only the accepted plan, it resolves the six requests that explicitly name the relevant entities. Neither condition resolves the six requests that need both types of information. With neither record, no request is resolved. The supplied-assignment condition returns the required relation directly.
\par\endgroup

The numerical test requires the correct assignment to be retrieved and enforced. Every plan in the returned archive must satisfy the problem constraints and preserve that assignment. Missing or incorrect assignments and infeasible archives receive zero quality. Full LSM and the supplied-assignment control each produce 18 valid results and mean HV of 0.825. The earlier-records-only condition produces six valid results and mean HV of 0.261. The accepted-plan-only condition produces six valid results and mean HV of 0.258. Table~\ref{tab:state-binding-factorial} separates these numerical results from the original language tests.

\begin{table}[!htbp]
\centering
\begingroup
\DenseResultTableSetup
\textbf{(a) Identifying the requested assignment: original language tests}\par\smallskip
\begin{tabularx}{\linewidth}{@{}l*{3}{>{\centering\arraybackslash}X}@{}}
\toprule
\TableHead
\textbf{Information provided} & \textbf{Uses earlier updates} & \textbf{Uses the accepted plan} & \textbf{Uses both records} \\
\midrule
Full LSM & 6/6 & 6/6 & 6/6 \\
\rowcolor{LiveOptBaselineRowB}
Earlier update records only & 6/6 & 0/6 & 0/6 \\
Accepted plan only & 0/6 & 6/6 & 0/6 \\
\rowcolor{LiveOptBaselineRowB}
Current request only & 0/6 & 0/6 & 0/6 \\
Assignment supplied directly & 6/6 & 6/6 & 6/6 \\
\bottomrule
\end{tabularx}
\par\smallskip
\textbf{(b) Preserving the assignment during search: replay results}\par\smallskip
\begin{tabularx}{\linewidth}{@{}l*{2}{>{\centering\arraybackslash}X}@{}}
\toprule
\TableHead
\textbf{Information provided} & \textbf{Valid search results} & \textbf{Mean HV} \\
\midrule
Full LSM & 18/18 & 0.825 \\
\rowcolor{LiveOptBaselineRowB}
Earlier update records only & 6/18 & 0.261 \\
Accepted plan only & 6/18 & 0.258 \\
\rowcolor{LiveOptBaselineRowB}
Current request only & 0/18 & 0.000 \\
Assignment supplied directly & 18/18 & 0.825 \\
\bottomrule
\end{tabularx}
\endgroup
\caption{\textbf{Using stored events and the accepted state.} (a) Original grounding results across three wordings, using minimum-recorded-fitness archive representatives. (b) Replay of the saved q0 queries against original online plans: six cases and three numerical seeds per condition, scored by the recorded evaluator. Missing or incorrect assignments and infeasible archives receive zero. In the supplied-assignment condition, a fixed procedure provides the expected relation directly.}
\label{tab:state-binding-factorial}
\end{table}

\paragraph{Comparison of reconstructed and online plans.}
We compare the requested assignments in the reconstructed representatives with those in the original online plans. The reconstructed representative and the original online plan differ in 12 of the 18 case--seed states. Among the twelve cases that read an accepted plan, the requested assignments differ in five. The differences occur in the requests that name orders and preserve their assignments in P010 (seeds 0 and 1), and in the requests that use both records in P015 (seeds 0--2). The requested assignments agree in the other seven cases. The six requests answered from earlier update records do not read the accepted plan.

Checksums confirm that all 18 starting populations match those in the original control and are shared across its five conditions. In the thirteen cases whose requested assignments are unchanged, the replay reproduces the scored objective sets for all five conditions.

\paragraph{Program construction without TSS.}

Without TSS, the model defines its own candidate representation and the functions that generate, evaluate, modify, and repair candidates. It retains this program through the sequence.
At \texttt{t00}, the model receives the public problem and the required Workbench functions.
It receives no TSS decision types, encoding guidance, or type-specific search operators.
It also defines data loading and conversion from a candidate to a solution.
Under Warm, it can repair only solutions from its own preceding state.
LPD still identifies which part of the problem changed.
LSM still supplies stored events, the method's own accepted state, and its search archives.
The fixed optimization loop still summarizes the change, applies the same restart selector, performs selection, maintains the archive, checks stopping, and formats the output.
Compilation, public validation, and the repair limit are unchanged.
This control removes the typed representation, type-specific operators, and restricted editing interface together.

One generated Workbench sequence is used per episode.
Seeds 0--9 evolve independent populations under that fixed model run.
Across P010--P015, the six model runs contain 13 distinct Workbench versions.
All 13 saved Workbench versions compile. The failing routing version omits vehicle uniqueness, producing 120 invalid update--seed outputs. The failing cloud version records GPU incompatibility without clearing the feasibility flag, producing 20 invalid outputs. The held-out evaluator assigns zero quality to these outputs.

\begin{table}[!htbp]
\centering
\ResultTableSetup
\begin{tabularx}{\linewidth}{@{}lcA{0.25\linewidth}Z@{}}
\toprule
\TableHead
\textbf{Episode} & \textbf{Failed results} & \textbf{Observed violation} & \textbf{Missing check} \\
\midrule
\LiveOptTSSFailureRows
\bottomrule
\end{tabularx}
\caption{\textbf{Hidden constraint failures of the persistent untyped Workbench.} Counts use the hidden evaluator on the final result of every failed update--seed result. The Workbench reports all 140 results as feasible. The hidden evaluator finds a hard-constraint violation, so each result contributes zero to online quality.}
\label{tab:appendix-no-tss-failure-attribution}
\end{table}
\FloatBarrier

\paragraph{Combining and modifying candidates.}
This test holds the optimization program, starting candidates, repair procedure, and restart choice fixed. Table~\ref{tab:appendix-tss-operator-control} reports the results.
At every P010/P015 update, both methods receive the same MAPLE population, Workbench, restart action, typed repair, and consistency checks.
The control replaces search operators designed for permutations, assignments, and values.
It instead selects a complete decision segment from one parent or samples a new valid segment at random.

\newcommand{\LiveOptTSSOperatorControlRows}{%
Typed operators & 100.0\% & 0.756 & 0.792 & 0.774 \\
\rowcolor{LiveOptBaselineRowB}
Generic random sampling & 91.7\% & 0.657 & 0.579 & 0.618 \\
}
\begin{table}[H]
\centering
\ResultTableSetup
\begin{tabularx}{\linewidth}{@{}l*{4}{Y}@{}}
\toprule
\TableHead
\textbf{Variation} & \TightStack{\textbf{Solve}\\\textbf{rate}} & \textbf{P010 HV} & \textbf{P015 HV} & \textbf{All HV} \\
\midrule
\LiveOptTSSOperatorControlRows
\bottomrule
\end{tabularx}
\caption{\textbf{Comparison of type-specific search operators.} Recorded-evaluator values cover t01--t12 with numerical seeds 0--2 under the same $200\times200$ limit and held-out reference. Only crossover and mutation change.}
\label{tab:appendix-tss-operator-control}
\end{table}

\begingroup\interlinepenalty=10000
\paragraph{Variation across independently generated programs.}
We generate three program sequences for MAPLE and its w/o-TSS variant on P010 routing and P015 cloud placement. The three sequences include the original main sequence. Each sequence uses search seeds 0--2 across all twelve updates. The Warm--Full choices, evaluation functions from the original runs, $200\times200$ budget, archive limit of 500, and stopping rule remain fixed. Final scores use the same held-out $500\times500\times10$ reference.
A model run that does not produce a valid Workbench receives zero solve rate and online quality for its requested results.
\par\endgroup

\begin{table}[!htbp]
\centering
\ResultTableSetup

\renewcommand{\arraystretch}{1.28}
\begin{tabularx}{\linewidth}{@{}l*{6}{Y}@{}}
\toprule
\TableHead
\textbf{Method} & \TightStack{\textbf{Model}\\\textbf{run}} & \TightStack{\textbf{Execution}\\\textbf{completed}} & \TightStack{\textbf{Solve}\\\textbf{rate (\%)}} & \TightStack{\textbf{Online}\\\textbf{quality}} & \TightStack{\textbf{Quality}\\\textbf{if valid}} & \TightStack{\textbf{Numerical}\\\textbf{SD}} \\
\midrule
\rowcolor{LiveOptTableLight}
\multicolumn{7}{l}{\textbf{P010 (routing)}} \\
MAPLE & 0 & yes & 100.0 & 0.756 & 0.756 & 0.120 \\
 & 1 & yes & 100.0 & 0.693 & 0.693 & 0.013 \\
 & 2 & yes & 100.0 & 0.560 & 0.560 & 0.038 \\
\rowcolor{LiveOptRowAccent}
 & \textit{All} & 3/3 & \TightStack{$100.0$\\$\pm 0.0$} & \TightStack{$0.669$\\$\pm 0.100$} & 0.669 & 0.057 \\
\midrule
MAPLE w/o TSS & 0 & yes & 0.0 & 0.000 & -- & 0.000 \\
 & 1 & yes & 100.0 & 0.435 & 0.435 & 0.016 \\
 & 2 & yes & 100.0 & 0.572 & 0.572 & 0.029 \\
\rowcolor{LiveOptBaselineRowB}
 & \textit{All} & 3/3 & \TightStack{$66.7$\\$\pm 57.7$} & \TightStack{$0.336$\\$\pm 0.299$} & 0.503 & 0.015 \\
\midrule
\rowcolor{LiveOptTableLight}
\multicolumn{7}{l}{\textbf{P015 (cloud)}} \\
MAPLE & 0 & yes & 100.0 & 0.792 & 0.792 & 0.022 \\
 & 1 & yes & 83.3 & 0.708 & 0.850 & 0.004 \\
 & 2 & yes & 83.3 & 0.685 & 0.822 & 0.010 \\
\rowcolor{LiveOptRowAccent}
 & \textit{All} & 3/3 & \TightStack{$88.9$\\$\pm 9.6$} & \TightStack{$0.728$\\$\pm 0.056$} & 0.819 & 0.012 \\
\midrule
MAPLE w/o TSS & 0 & yes & 83.3 & 0.751 & 0.901 & 0.011 \\
 & 1 & yes & 100.0 & 0.059 & 0.059 & 0.001 \\
 & 2 & yes & 83.3 & 0.729 & 0.874 & 0.019 \\
\rowcolor{LiveOptBaselineRowB}
 & \textit{All} & 3/3 & \TightStack{$88.9$\\$\pm 9.6$} & \TightStack{$0.513$\\$\pm 0.393$} & 0.577 & 0.011 \\
\bottomrule
\end{tabularx}
\caption{\textbf{Independent model runs on one routing and one cloud episode.} Recorded-evaluator control. Each run independently generates a Workbench sequence. Numerical seeds 0--2 then search that sequence over t01--t12. Warm--Full choices remain fixed, so only Workbench construction and maintenance vary. Solve rate uses the held-out validity check. Online quality sets invalid results to zero. The \textit{All} rows report mean $\pm$ sample standard deviation across model runs. Quality if valid includes accepted states only. Numerical SD measures variation across search seeds within one model run.}
\label{tab:controller-repeat}
\end{table}

Across three Workbench runs, MAPLE's mean online quality is 0.669 in routing and 0.728 in cloud, compared with 0.336 and 0.513 without TSS. All three typed routing runs remain valid. The two additional typed cloud runs fail the held-out check at t11--t12. Table~\ref{tab:controller-repeat} reports model-run variation separately from numerical-seed variation for these two episodes.

\paragraph{Evaluation with a second language model.}
We repeat the full benchmark with Kimi k2.7 using the same evaluation and search settings. Table~\ref{tab:appendix-cross-model-main} reports the results under the common public-formula evaluation protocol.

\newcommand{\LiveOptCrossModelMainRows}{%
\rowcolor{LiveOptRowAccent}
\method{} (DeepSeek-V4-Pro Preview) & 100.0\% & 0.951 & 100.0\% & 0.875 \\
\method{} (Kimi k2.7) & 100.0\% & 0.937 & 92.3\% & 0.766 \\
}
\begin{table}[!htbp]
\centering
\ResultTableSetup
\begin{tabularx}{\linewidth}{@{}l*{4}{Y}@{}}
\toprule
\TableHead
& \multicolumn{2}{c}{\textbf{DS}} & \multicolumn{2}{c}{\textbf{DM}} \\
\cmidrule(lr){2-3}\cmidrule(lr){4-5}
\TableHead
\textbf{Language model} & \TightStack{\textbf{Solve}\\\textbf{rate}} & \TightStack{\textbf{Online}\\\textbf{quality}} & \TightStack{\textbf{Solve}\\\textbf{rate}} & \TightStack{\textbf{Online}\\\textbf{quality}} \\
\midrule
\LiveOptCrossModelMainRows
\bottomrule
\end{tabularx}
\caption{\textbf{Full-benchmark MAPLE replication across language models.} Both rows use the complete \method{} system with one model run per episode and ten numerical seeds. Public-formula evaluation, search budget, and stopping rule match Table~\ref{tab:main-dynamic-results}.}
\label{tab:appendix-cross-model-main}
\end{table}
\FloatBarrier

\begingroup\interlinepenalty=10000
\paragraph{Model calls and tokens for program construction.}
We group recorded calls into program construction, identification of requested edits, and input-table updates. The grouping uses each recorded call's system prompt. Tables~\ref{tab:appendix-controller-cost}--\ref{tab:execution-cost-metadata} report means over four additional runs per method on P010/P015. Each run covers twelve updates with three search seeds and fixed restart choices. Code construction includes generation, modification, and repair. Token counts include both input and output. The w/o-TSS runs reuse saved localization decisions and data patches.
\par\endgroup

\begin{table}[!htbp]
\centering
\ResultTableSetup
\begin{tabularx}{\linewidth}{@{}l*{4}{Y}@{}}
\toprule
\TableHead
\textbf{Method} & \TightStack{\textbf{Code}\\\textbf{calls}} & \TightStack{\textbf{Code}\\\textbf{tokens}} & \TightStack{\textbf{Tokens for}\\\textbf{identifying}\\\textbf{requested edits}} & \TightStack{\textbf{Tokens for}\\\textbf{updating}\\\textbf{input tables}} \\
\midrule
\LiveOptCodeConstructionRows
\bottomrule
\end{tabularx}
\caption{\textbf{Code construction uses fewer tokens with TSS.} Reused denotes update processing supplied from existing records.}
\label{tab:appendix-controller-cost}
\end{table}

Initial code generation averages 19.1k tokens with TSS and 26.0k without it. Code modification averages 32.5k and 46.9k, respectively, including repair calls in both categories. The combined reduction is computed before rounding.

\begin{table}[!htbp]
\centering
\ResultTableSetup
\begin{tabularx}{\linewidth}{@{}lZcccc@{}}
\toprule
\TableHead
\textbf{Method} & \textbf{Recorded scope} & \textbf{Tokens} & \textbf{Calls} & \TightStack{\textbf{Provider}\\\textbf{min.}} & \TightStack{\textbf{Wall}\\\textbf{min.}} \\
\midrule
MAPLE & Code, localization, data patches & 255.4k & 27.8 & 14.7 & 19.9 \\
w/o TSS & Code; supplied update processing & 72.8k & 4.8 & 6.5 & 11.1 \\
\bottomrule
\end{tabularx}
\caption{\textbf{Execution metadata for the same runs as Table~\ref{tab:appendix-controller-cost}.} Wall time includes numerical search.}
\label{tab:execution-cost-metadata}
\end{table}
Mean input/output counts are 202.6k/52.9k for MAPLE and 47.7k/25.1k for w/o TSS.
Input cache hits of 41.4k and 6.7k are included in the input totals.

\paragraph{Reuse strategies over complete update sequences.}
Each strategy starts from the same initial state and continues with candidates produced by its own searches. The initial state is the typed Workbench and population at t00. Each strategy proceeds through t01--t12.
Always Full keeps the TSS Workbench but draws all 200 candidates anew at every update.
The policies use the same saved Workbench at each state: 60 initial execution instances and 720 update execution instances across six episodes and ten numerical seeds.

\begin{table}[!htbp]
\centering
\ResultTableSetup
\begin{tabularx}{\linewidth}{@{}l*{5}{Y}@{}}
\toprule
\TableHead
\textbf{Sequential policy} & \textbf{Full restarts} & \TightStack{\textbf{Solve}\\\textbf{rate}} & \TightStack{\textbf{Online}\\\textbf{quality}} & \textbf{Generations} & \TightStack{\textbf{$\Delta$ quality}\\\textbf{vs. MAPLE}} \\
\midrule
\LiveOptSequentialRestartRows
\bottomrule
\end{tabularx}
\caption{\textbf{Restart policies that follow their own populations on P010--P015.} Public-formula evaluation of t01--t12. Warm and Full carry their own results through all updates. Every policy uses the same TSS Workbenches, $200\times200$ maximum budget, 10 seeds, and stopping rule. Generations are taken from the original executions. Paired differences are computed before rounding. Displayed means are rounded separately.}
\label{tab:appendix-sequential-restart}
\end{table}

Across six episodes, the mean paired gain over Always Warm is 0.010. Five episode means are positive. The episode-cluster bootstrap interval is [0.001, 0.021], and the two-sided exact sign test gives p = 0.219. The gain is concentrated in routing. The mean gain over Always Full is 0.054. Table~\ref{tab:public-formula-evaluation} reports the results by domain.

\paragraph{Recorded reuse and restart decisions.}
\label{app:restart-ablation-details}
We report the initialization action selected before search for each saved update.

Across P007--P015, the selector chooses Warm at t01--t10 for all nine episodes.
It chooses Full at t11 for all nine.
At t12, it chooses Full for P008 and P010--P015 and Warm for P007/P009.
The total is 16 Full and 92 Warm decisions.
These decisions are made before numerical optimization and are identical across the 10 numerical seeds.
The disruptive updates appear at t11--t12 by design.
At P010--P012 t08, the selector chooses Warm and cites the retained decision representation, active entities, and objective declarations. On the main Pareto trajectories, the deterministic table-change rule makes the same Warm--Full choices as the selector.

\paragraph{Restart decisions based on changed input cells.}
We compare MAPLE with a rule that starts afresh when at least half of the counted input cells change.
For every table with an \texttt{active} field, it counts the cells in rows whose current \texttt{active} value is true, excluding the \texttt{id} and \texttt{active} fields themselves. This count is the table's denominator.
Rows are matched by their raw public \texttt{id} values. A row with an identifier absent from the preceding table, or previously inactive, contributes all its counted cells to the numerator. For other rows, only unequal cell values count as changed. The rule uses the maximum table fraction, with a zero fraction for an empty denominator.
The rule chooses Full when this fraction is at least 0.5 and Warm otherwise.
Its choices are fixed before the Warm and Full results are examined.
Across P010--P015, the ratio ranges from 0 to 0.060 on ordinary updates and from 0.705 to 1.000 at t11--t12.
The rule therefore matches all 12 Full decisions from the restart selector.
It obtains the same 0.779 HV under the recorded evaluator and 162.8 mean generations without a model call or a new optimizer run.

\paragraph{Reuse after rescaling, renaming, and task addition.}
\label{app:reuse-discriminator}
Starting from saved routing and cloud-placement states, we rescale values, rename entities, or add small tasks. Each additional test starts at t05 of P010 or P015.
Each affects more than half of the counted table cells.
The editable Workbench functions, decision types, and required output remain unchanged, while the number of active decisions grows when tasks are added.
A fixed procedure constructs the tables and identifier mappings without model calls.

\textbf{Rescaling routing parameters.}
For routing rescaling, depot and order coordinates, order demand, ready/due/service times, vehicle capacity, and shift end are multiplied by 10. Priorities and emission rates stay fixed.
The historical Workbench identifies congestion by the original coordinate pair $(48,11)$ (Figure~\ref{fig:appendix-real-patch-diff}).
After coordinate rescaling, that match disappears, so this branch changes the incident's effect as well as the parameter scale.
Its table-change ratio is 0.86.

\textbf{Rescaling cloud parameters.}
For cloud rescaling, job CPU and memory demands, machine CPU and memory capacities, and idle energy are multiplied by 10.
Job deadlines, latency sensitivities, and priorities are also multiplied by 10 but enter only diagnostics in this Workbench.
GPU requirements, availability, per-CPU energy rates, and global energy multipliers stay fixed.
For any fixed feasible assignment, utilization and the recorded imbalance $B$ are unchanged, while energy becomes $E'=10E$.
Thus feasibility and Pareto dominance are preserved, although scalar Warm repair uses the changed weighting $10E+B$.
The table-change ratio is 0.71.

\textbf{Renaming entities.}
Relabeling consistently renames orders and vehicles, or jobs and machines, without changing their other attributes (ratio 1.00).
The table-change rule does not apply the identifier mapping before matching rows, so every renamed row is treated as new even though its non-identifier values are retained.

\textbf{Adding tasks.}
Task additions introduce 20 low-demand orders near the depot or 26 small CPU/memory jobs with loose deadlines (ratios 0.53 and 0.52).
These additions enlarge the task while retaining the existing resources and earlier demands.

\textbf{Comparing the restart choices.}
Warm and Full begin with the same saved candidate population and use three search seeds. They use the original objective definitions and the combined nondominated reference for each additional test. The table-change rule and the numerical rule that evaluates candidates before updating their identifiers choose Full on all six cases.
The restart selector chooses Warm on all six and cites the rescaling, renaming, or easily placed additions in its recorded assessment.

Warm obtains mean HV of 0.862 across the six cases, compared with 0.682 for Full.
The largest difference is +0.708 for Warm (Table~\ref{tab:churn-discriminator}).
The six cases cover routing and cloud placement. The routing-rescaling case also changes how congestion is applied.
P015 recovers almost fully from fresh restarts, so most of the quality difference comes from routing.
Warm uses fewer generations in five of the six branches.
Table~\ref{tab:main-reuse-discriminator} presents the relabeling and task-addition results compactly alongside the disruptive main-sequence updates. The rescaling branches and the pure-scaling numerical control below are supplementary diagnostics.

\begin{table}[!htbp]
\centering
\begingroup
\ResultTableSetup
\begin{tabularx}{\linewidth}{@{}lYYcYY@{}}
\toprule
\TableHead
\textbf{Case} & \TightStack{\textbf{Warm}\\\textbf{HV}} & \TightStack{\textbf{Full}\\\textbf{HV}} & \TightStack{\textbf{Paired $\Delta$ HV}\\\textbf{Mean $\pm$ SD}} & \TightStack{\textbf{Min.}\\\textbf{$\Delta$ HV}} & \TightStack{\textbf{Max.}\\\textbf{$\Delta$ HV}} \\
\midrule
\LiveOptChurnDiscriminatorRows
\bottomrule
\end{tabularx}
\endgroup
\caption{\textbf{Reuse after rescaling, relabeling, and task additions.} Six branches start at t05, with three matched seeds per action and 36 executions. For each branch, $\Delta_s=\mathrm{HV}_{\mathrm{Warm},s}-\mathrm{HV}_{\mathrm{Full},s}$ uses the same incoming population and branch-specific pooled reference. SD is the sample deviation of the three paired differences. Min./max. are observed seed extrema, not confidence bounds. The last row averages the six branch means. Recorded-evaluator, action-matched control. Routing rescaling also removes the coordinate-based congestion match. Rounding follows Table~\ref{tab:appendix-sequential-restart}.}
\label{tab:churn-discriminator}
\end{table}

For each branch, $\bar\Delta=\frac{1}{3}\sum_{s=0}^{2}\Delta_s$ and
$s_\Delta=\sqrt{\frac{1}{2}\sum_{s=0}^{2}(\Delta_s-\bar\Delta)^2}$.
The three routing task-addition differences are all positive, ranging from 0.442 to 0.892.
Routing rescaling and relabeling have positive means but each includes a negative seed difference.
The cloud differences are small, consistent with the similar final quality and earlier Warm stopping discussed above.

\paragraph{Updating candidate identifiers before evaluation.}
\label{app:representation-aligned-controls}
We replace the identifiers in each old candidate before evaluating it against the renamed tables.
For relabeling, we compare $f_{\mathrm{old}}(x)$ with $f_{\mathrm{new}}(T(x))$, where $T$ applies the public ID mapping to assignment keys, assigned resources, and route permutations. The numerical selector retains its original sample indices and thresholds. All 1,200 incoming candidates across the two profiles and three seeds preserve their objective vectors and feasibility after mapping. The six sampled decisions change from Full to Warm: both objective displacement and feasibility loss become zero.

\paragraph{Reuse under uniform routing rescaling.}
We keep the congestion rule attached to the same order before and after scaling, then compare Warm and Full. We replace the coordinate lookup with the same public order ID in both pre-update and scaled Workbenches. The incident's historical effect and all other recorded scoring rules remain fixed. On all 600 incoming candidates, the unscaled rewrite reproduces the historical evaluator, and scaling multiplies each objective by 10 while preserving feasibility. The numerical selector has $r=0.9$ and chooses Full for all three seeds.

\newcommand{\RepresentationAlignedRows}{%
\TightStack{P010: Candidates updated\\to renamed IDs} & 0.786 & 0.395 & $+0.391 \pm 0.278$ \\
\TightStack{P015: Candidates updated\\to renamed IDs} & 0.984 & 0.982 & $+0.002 \pm 0.004$ \\
P010 pure scaling & 0.791 & 0.466 & $+0.325 \pm 0.299$ \\
}
\newcommand{\RepresentationAlignedSeedRows}{%
\TightStack{P010: Candidates updated\\to renamed IDs} & 0 & 0.848 & 0.310 & $+0.538$ & 75/200 \\
 & 1 & 0.966 & 0.402 & $+0.564$ & 70/200 \\
 & 2 & 0.545 & 0.474 & $+0.071$ & 158/200 \\
\midrule
\TightStack{P015: Candidates updated\\to renamed IDs} & 0 & 0.983 & 0.977 & $+0.005$ & 176/200 \\
 & 1 & 0.983 & 0.986 & $-0.002$ & 200/200 \\
 & 2 & 0.985 & 0.984 & $+0.002$ & 200/200 \\
\midrule
P010 pure scaling & 0 & 0.840 & 0.649 & $+0.191$ & 65/200 \\
 & 1 & 0.971 & 0.303 & $+0.668$ & 70/200 \\
 & 2 & 0.562 & 0.446 & $+0.116$ & 158/200 \\
}
\begin{table}[!htbp]
\centering
\ResultTableSetup
\begin{tabularx}{\linewidth}{@{}lYYc@{}}
\toprule
\TableHead
\textbf{Additional control} & \textbf{Warm HV} & \textbf{Full HV} & \textbf{Paired $\Delta$ HV} \\
\midrule
\RepresentationAlignedRows
\bottomrule
\end{tabularx}
\caption{\textbf{Reuse with consistent representations.} Eighteen new searches: three branches, three paired seeds, and two actions, with population 200 and a 200-generation cap. Each branch uses the pooled nondominated reference from its six searches. Differences are Warm minus Full, reported as mean $\pm$ sample SD.}
\label{tab:representation-aligned-controls}
\end{table}

Warm has higher HV for every routing seed in these controls. The cloud difference is small: the three paired gains are 0.0054, $-$0.0023, and 0.0015. In the pure-scaling case, Warm uses 97.7 generations on average and Full uses 200. The generation counts do not include the extra evaluations used to adapt and repair old candidates.

\begin{table}[!htbp]
\centering
\ResultTableSetup
\begin{tabularx}{\linewidth}{@{}Zccccc@{}}
\toprule
\TableHead
\textbf{Control} & \textbf{Seed} & \textbf{Warm HV} & \textbf{Full HV} & \textbf{$\Delta$ HV} & \textbf{Generations} \\
\midrule
\RepresentationAlignedSeedRows
\bottomrule
\end{tabularx}
\caption{\textbf{Individual seeds for the offline controls.} Generations are Warm/Full. Identical incoming-population hashes are checked within each action pair; mapped-ID Full archives also reproduce all six historical Full archives exactly.}
\label{tab:representation-aligned-seeds}
\end{table}

The 18 restart-and-search calls take 170.6 seconds in total, including migration and repair but excluding source-state reconstruction and scoring.

\paragraph{Restart actions from a shared candidate population.}
\label{app:recorded-matched-restart}
For each update, both restart actions receive MAPLE's same saved candidate population. These are the Matched Warm and Matched Full controls. Only MAPLE's selected result advances the continuing trajectory.
All action-matched comparisons use the recorded evaluator with population 200, a 200-generation cap, seeds 0--9, the same stopping rule, and the same held-out 500\(\times\)500\(\times\)10 reference for final scoring.
The Warm/Full comparison contains 1,080 paired episode--time-step--seed results per method over P007--P015.
The numerical selector covers the 720 Pareto results on P010--P015.

\begin{table}[!htbp]
\centering
\MainResultTableSetup
\begin{tabularx}{\linewidth}{@{}l*{4}{Y}@{}}
\toprule
\TableHead
\textbf{Policy} & \textbf{Full restarts} & \textbf{HV} & \textbf{Generations} & \textbf{Gap to best action} \\
\midrule
\LiveOptRestartRegretDMRows
\bottomrule
\end{tabularx}
\caption{\textbf{Action-matched restart under the recorded evaluator.} All policies solve every state.}
\label{tab:main-restart-regret}
\end{table}

\paragraph{Re-evaluating old candidates to select a restart action.}
We evaluate a sample of earlier candidates under the old and revised problems before selecting Warm or Full. This comparison follows a common response to change in dynamic evolutionary optimization \citep{deb2007dynamic,liu2014adaptive,sahmoud2019dynamism}.
For each seed and update, it samples 20 of the 200 incoming candidates without replacement.
It evaluates the same candidates with the old and updated Workbenches.
For objective $j$, it computes
\[
 r_j=\frac{1}{K}\sum_{i=1}^{K}
 \frac{|f_j(x_i,t)-f_j(x_i,t-1)|}
 {\max(|f_j(x_i,t)|,|f_j(x_i,t-1)|,10^{-12})},
 \qquad r=\max_j r_j .
\]
$K$ is the number of retained valid objective pairs.
It chooses Full when $r\geq0.5$.
It also chooses Full when at least half of the previously feasible sampled candidates become infeasible, the objective dimension changes, or no valid objective pair remains. A valid pair contains finite objective vectors of the same nonzero length. If the old evaluation fails, the rule uses the saved fitness. Other unsuccessful pairs are omitted.
It chooses Warm otherwise.
The maximum over objectives keeps one large change from being averaged away.
The relative denominator makes objective units comparable.
Sampling is fixed by the numerical seed and time step.
The numerical selector evaluates each unchanged incoming genome under both Workbenches before migration or ID mapping. All six relabeling seed cases retain 20 defined objective pairs, but lose feasibility for all sampled candidates and have $r=1$. Their Full decisions therefore include representation mismatch. The mapped-candidate control in Appendix~\ref{app:representation-aligned-controls} removes this mismatch and changes all six decisions to Warm.

\begin{table}[!htbp]
\centering
\ResultTableSetup

\renewcommand{\arraystretch}{1.40}
\begin{tabularx}{\linewidth}{@{}l*{5}{Y}@{}}
\toprule
\TableHead
\textbf{Episode} & \textbf{Full t01--t10} & \textbf{Full t11--t12} & \textbf{Numerical HV} & \textbf{MAPLE HV} & \textbf{Matched Warm HV} \\
\midrule
\LiveOptLandscapeRestartRows
\bottomrule
\end{tabularx}
\caption{\textbf{Numerical change as a restart signal.} Recorded-evaluator, action-matched control over t01--t12. Counts are episode--time-step--seed results. The numerical selector chooses Full on all 120 disruptive update--seed cases and on 125 of the 600 ordinary cases.}
\label{tab:appendix-landscape-restart}
\end{table}

Appendix~\ref{app:restart-threshold-sensitivity} reports the threshold comparison.

\begin{table}[!htbp]
\centering
\DenseResultTableSetup

\renewcommand{\arraystretch}{1.40}
\begin{tabularx}{\linewidth}{@{}l*{6}{Y}@{}}
\toprule
\TableHead
\textbf{Variant} & \textbf{DS Quality} & \textbf{DS Gen.} & \textbf{DM HV} & \textbf{t11 HV} & \textbf{t12 HV} & \textbf{DM Gen.} \\
\midrule
\rowcolor{LiveOptRowAccent}
MAPLE & 0.927 & 66.7 & 0.779 & 0.561 & 0.426 & 162.8 \\
Matched Warm & 0.927 & 66.8 & 0.766 & 0.509 & 0.327 & 162.7 \\
\rowcolor{LiveOptBaselineRowB}
Matched Full & 0.919 & 71.1 & 0.731 & 0.561 & 0.426 & 199.0 \\
\bottomrule
\end{tabularx}
\caption{Recorded-evaluator, action-matched control over t01--t12. \textbf{Restart actions from the same incoming population.} All methods solve all 360 DS and 720 DM time-step--seed results. Quality is normalized scalar score for DS and held-out-reference HV ratio for DM. Gen. is the mean generation count. The t11/t12 columns show MAPLE's 0.053/0.099 gains over Matched Warm.}
\label{tab:main-restart-ablation}
\end{table}

\begingroup\interlinepenalty=10000
Table~\ref{tab:main-restart-ablation} reports stage-end solution quality and the number of evolutionary generations for each restart policy.
On P010--P015, MAPLE improves over Matched Warm by 0.013 normalized HV
(episode-cluster 95\% bootstrap interval $[0.004,0.022]$), with the gain concentrated at
t11--t12. All six episode means favor MAPLE over Matched Warm (two-sided exact sign test, $p=0.031$).
MAPLE recovers 74\% of the improvement obtained by choosing the better action after seeing both outcomes.
The mean HV gain over Matched Full is 0.048 (hierarchical paired 95\% CI $[0.005,0.097]$). MAPLE uses 162.8 generations on average, compared with 199.0 for Matched Full.
At t11--t12, where the selector chooses Full for all six Pareto episodes, it matches
the corresponding Matched Full controls.
On P007--P009, MAPLE and Matched Warm have nearly the same result.
Always using Full is slightly worse and uses more generations.

The per-episode breakdown in Table~\ref{tab:restart-p010-p015-problems} localizes this effect. Across routing episodes P010--P012, the disruptive-update gain over Matched Warm ranges from 0.122 to 0.168 HV. The corresponding cloud gains are 0.002--0.011, consistent with the stronger benefit of fresh initialization in routing.
\par\endgroup


\begin{table}[!htbp]
\centering
\begingroup
\AppendixTableSetup
\renewcommand{\arraystretch}{1.30}
\setlength{\tabcolsep}{2pt}
\noindent\textbf{Routing}\quad Family mean: \textbf{M} \textbf{0.757}, \textbf{N} 0.391, \textbf{W} 0.733, \textbf{F} 0.656\par\vspace{3pt}
\begin{tabularx}{\linewidth}{@{}l*{12}{Y}@{}}
\toprule
\TableHead
\textbf{Time} & \multicolumn{4}{c}{\textbf{P010}} & \multicolumn{4}{c}{\textbf{P011}} & \multicolumn{4}{c}{\textbf{P012}} \\
\cmidrule(lr){2-5}\cmidrule(lr){6-9}\cmidrule(lr){10-13}
\TableHead
 & \textbf{M} & \textbf{N} & \textbf{W} & \textbf{F} & \textbf{M} & \textbf{N} & \textbf{W} & \textbf{F} & \textbf{M} & \textbf{N} & \textbf{W} & \textbf{F} \\
\midrule
\rowcolor{LiveOptRowAccent}
\textbf{All-t} & \textbf{0.752} & 0.000 & 0.724 & 0.663 & \textbf{0.736} & 0.327 & 0.712 & 0.662 & 0.785 & \textbf{0.845} & 0.764 & 0.643 \\
t01 & \textbf{0.746} & 0.000 & \textbf{0.746} & 0.710 & \textbf{0.731} & 0.379 & \textbf{0.731} & 0.663 & 0.847 & \textbf{0.995} & 0.847 & 0.684 \\
\rowcolor{LiveOptBaselineRowB}
t02 & \textbf{0.774} & 0.000 & \textbf{0.774} & 0.657 & \textbf{0.776} & 0.492 & \textbf{0.776} & 0.726 & \textbf{0.872} & 0.838 & \textbf{0.872} & 0.744 \\
t03 & \textbf{0.791} & 0.000 & \textbf{0.791} & 0.769 & \textbf{0.778} & 0.440 & \textbf{0.778} & 0.730 & 0.815 & \textbf{0.880} & 0.815 & 0.642 \\
\rowcolor{LiveOptBaselineRowB}
t04 & \textbf{0.778} & 0.000 & \textbf{0.778} & 0.696 & \textbf{0.768} & 0.450 & \textbf{0.768} & 0.704 & 0.832 & \textbf{0.931} & 0.832 & 0.659 \\
t05 & \textbf{0.805} & 0.000 & \textbf{0.805} & 0.668 & \textbf{0.797} & 0.397 & \textbf{0.797} & 0.692 & 0.838 & \textbf{0.869} & 0.838 & 0.640 \\
\rowcolor{LiveOptBaselineRowB}
t06 & \textbf{0.849} & 0.000 & \textbf{0.849} & 0.751 & \textbf{0.842} & 0.241 & \textbf{0.842} & 0.772 & 0.872 & \textbf{0.895} & 0.872 & 0.723 \\
t07 & \textbf{0.850} & 0.000 & \textbf{0.850} & 0.733 & \textbf{0.846} & 0.289 & \textbf{0.846} & 0.736 & 0.876 & \textbf{0.940} & 0.876 & 0.688 \\
\rowcolor{LiveOptBaselineRowB}
t08 & \textbf{0.862} & 0.000 & \textbf{0.862} & 0.764 & \textbf{0.784} & 0.463 & \textbf{0.784} & 0.629 & 0.905 & \textbf{0.931} & 0.905 & 0.699 \\
t09 & \textbf{0.886} & 0.000 & \textbf{0.886} & 0.659 & \textbf{0.804} & 0.532 & \textbf{0.804} & 0.715 & 0.891 & \textbf{0.938} & 0.891 & 0.706 \\
\rowcolor{LiveOptBaselineRowB}
t10 & \textbf{0.910} & 0.000 & \textbf{0.910} & 0.785 & \textbf{0.918} & 0.238 & \textbf{0.918} & 0.797 & \textbf{0.898} & 0.797 & \textbf{0.898} & 0.766 \\
t11 & \textbf{0.342} & 0.000 & 0.230 & \textbf{0.342} & \textbf{0.239} & 0.000 & 0.040 & \textbf{0.239} & 0.397 & \textbf{0.656} & 0.395 & 0.397 \\
\rowcolor{LiveOptBaselineRowB}
t12 & \textbf{0.427} & 0.000 & 0.203 & \textbf{0.427} & \textbf{0.545} & 0.000 & 0.458 & \textbf{0.545} & 0.370 & \textbf{0.469} & 0.129 & 0.370 \\
\bottomrule
\end{tabularx}
\par\vspace{8pt}
\noindent\textbf{Cloud}\quad Family mean: \textbf{M} 0.801, \textbf{N} \textbf{0.821}, \textbf{W} 0.799, \textbf{F} 0.805\par\vspace{3pt}
\begin{tabularx}{\linewidth}{@{}l*{12}{Y}@{}}
\toprule
\TableHead
\textbf{Time} & \multicolumn{4}{c}{\textbf{P013}} & \multicolumn{4}{c}{\textbf{P014}} & \multicolumn{4}{c}{\textbf{P015}} \\
\cmidrule(lr){2-5}\cmidrule(lr){6-9}\cmidrule(lr){10-13}
\TableHead
 & \textbf{M} & \textbf{N} & \textbf{W} & \textbf{F} & \textbf{M} & \textbf{N} & \textbf{W} & \textbf{F} & \textbf{M} & \textbf{N} & \textbf{W} & \textbf{F} \\
\midrule
\rowcolor{LiveOptRowAccent}
\textbf{All-t} & \textbf{0.854} & 0.837 & 0.853 & 0.851 & 0.763 & \textbf{0.870} & 0.761 & 0.773 & 0.785 & 0.755 & 0.784 & \textbf{0.791} \\
t01 & \textbf{0.915} & 0.813 & \textbf{0.915} & 0.873 & 0.815 & \textbf{0.905} & 0.815 & 0.833 & 0.826 & \textbf{0.883} & 0.826 & 0.816 \\
\rowcolor{LiveOptBaselineRowB}
t02 & 0.917 & 0.901 & 0.917 & \textbf{0.929} & 0.743 & \textbf{0.880} & 0.743 & 0.763 & 0.821 & \textbf{0.896} & 0.821 & 0.836 \\
t03 & 0.899 & 0.874 & 0.899 & \textbf{0.901} & 0.782 & \textbf{0.881} & 0.782 & 0.764 & 0.846 & \textbf{0.915} & 0.846 & 0.859 \\
\rowcolor{LiveOptBaselineRowB}
t04 & 0.879 & 0.878 & 0.879 & \textbf{0.898} & 0.770 & \textbf{0.892} & 0.770 & 0.795 & 0.833 & \textbf{0.900} & 0.833 & 0.835 \\
t05 & 0.911 & 0.902 & 0.911 & \textbf{0.913} & 0.823 & \textbf{0.927} & 0.823 & 0.840 & 0.834 & \textbf{0.905} & 0.834 & 0.854 \\
\rowcolor{LiveOptBaselineRowB}
t06 & 0.898 & 0.888 & 0.898 & \textbf{0.909} & 0.843 & \textbf{0.927} & 0.843 & 0.860 & 0.846 & \textbf{0.908} & 0.846 & 0.859 \\
t07 & 0.924 & 0.914 & 0.924 & \textbf{0.924} & 0.748 & \textbf{0.907} & 0.748 & 0.786 & 0.813 & \textbf{0.900} & 0.813 & 0.840 \\
\rowcolor{LiveOptBaselineRowB}
t08 & \textbf{0.887} & 0.880 & \textbf{0.887} & 0.885 & 0.749 & \textbf{0.897} & 0.749 & 0.781 & 0.845 & \textbf{0.915} & 0.845 & 0.837 \\
t09 & \textbf{0.882} & 0.866 & \textbf{0.882} & 0.873 & 0.751 & \textbf{0.921} & 0.751 & 0.725 & 0.834 & \textbf{0.912} & 0.834 & 0.831 \\
\rowcolor{LiveOptBaselineRowB}
t10 & \textbf{0.891} & 0.866 & \textbf{0.891} & 0.870 & 0.823 & \textbf{0.921} & 0.823 & 0.825 & 0.871 & \textbf{0.920} & 0.871 & 0.867 \\
t11 & 0.768 & \textbf{0.780} & 0.763 & 0.768 & 0.762 & \textbf{0.770} & 0.767 & 0.762 & \textbf{0.858} & 0.000 & 0.857 & \textbf{0.858} \\
\rowcolor{LiveOptBaselineRowB}
t12 & 0.473 & \textbf{0.486} & 0.475 & 0.473 & 0.546 & \textbf{0.617} & 0.520 & 0.546 & \textbf{0.197} & 0.000 & 0.178 & \textbf{0.197} \\
\bottomrule
\end{tabularx}
\endgroup
\addtolength{\abovecaptionskip}{0pt}
\caption{\textbf{Recorded-evaluator results at each Pareto update.} M denotes MAPLE and N its independently generated untyped Workbench control. W and F denote Matched Warm and Matched Full, each initialized from MAPLE's incoming population. Values are ten-seed HV means, with invalid outputs set to zero. All-t averages t01--t12. Family means pool routing or cloud results.}
\label{tab:appendix-ablation-stage-grid}
\end{table}

\begin{table}[!htbp]
\renewcommand{\arraystretch}{1.0}
\centering
\DenseResultTableSetup
\renewcommand{\arraystretch}{1.0}
\begin{tabularx}{\linewidth}{@{}l*{6}{Y}@{}}
\toprule
\TableHead
& \multicolumn{3}{c}{\textbf{All t01--t12}} & \multicolumn{3}{c}{\textbf{Disruptive updates t11--t12}} \\
\cmidrule(lr){2-4}\cmidrule(lr){5-7}
\TableHead
\textbf{Episode} & \textbf{MAPLE} & \TightStack{\textbf{Matched}\\\textbf{Warm}} & \TightStack{\textbf{Matched}\\\textbf{Full}} & \textbf{MAPLE} & \TightStack{\textbf{Matched}\\\textbf{Warm}} & \TightStack{\textbf{Matched}\\\textbf{Full}} \\
\midrule
P010 & 0.752 & 0.724 & 0.663 & 0.385 & 0.217 & 0.385 \\
\rowcolor{LiveOptBaselineRowB}
P011 & 0.736 & 0.712 & 0.662 & 0.392 & 0.249 & 0.392 \\
P012 & 0.785 & 0.764 & 0.643 & 0.384 & 0.262 & 0.384 \\
\rowcolor{LiveOptBaselineRowB}
P013 & 0.854 & 0.853 & 0.851 & 0.621 & 0.619 & 0.621 \\
P014 & 0.763 & 0.761 & 0.773 & 0.654 & 0.644 & 0.654 \\
\rowcolor{LiveOptBaselineRowB}
P015 & 0.785 & 0.784 & 0.791 & 0.528 & 0.517 & 0.528 \\
\bottomrule
\end{tabularx}
\caption{\textbf{Online quality for the six Pareto episodes.} Recorded-evaluator, action-matched control. All three policies have 100\% solve rate. Each value is the normalized-HV mean over 10 seeds after infeasible results are set to zero. The right block isolates the disruptive t11--t12 updates.}
\label{tab:restart-p010-p015-problems}
\par\vspace{8pt}

\centering
\ResultTableSetup
\begin{tabularx}{\linewidth}{@{}l*{4}{Y}@{}}
\toprule
\TableHead
\textbf{Scalar restart policy} & \textbf{Full restarts} & \textbf{Online quality} & \textbf{Generations} & \textbf{Gap to best action} \\
\midrule
\LiveOptRestartRegretDSRows
\bottomrule
\end{tabularx}
\caption{\textbf{Restart results on P007--P009.} Recorded-evaluator, action-matched t01--t12. All policies solve every state. MAPLE and Matched Warm both reach mean scalar quality 0.927. Gap uses the post-hoc best action.}
\label{tab:appendix-restart-regret-ds}
\end{table}

\begin{figure}[!htbp]
\centering
\begin{minipage}[c]{0.61\linewidth}
\centering
\includegraphics[width=\linewidth]{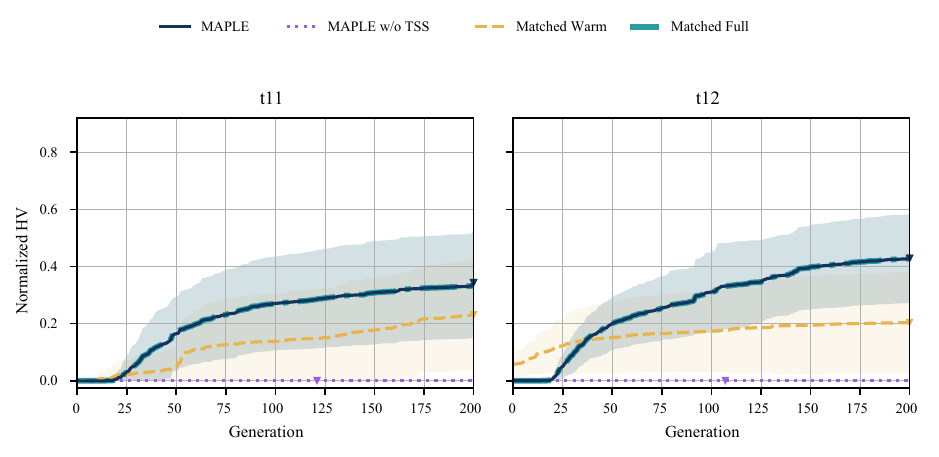}
\end{minipage}\hfill
\begin{minipage}[c]{0.355\linewidth}
\small
\textbf{Disruptive update.}
At t11--t12, the routing episode changes orders, depot, vehicles, and operating rules while retaining its representation and objectives.

\textbf{Current action.}
The restart selector selects Full. It matches the Matched Full control because both use the same action and incoming state, and exceeds Matched Warm under this recorded evaluator.

\textbf{Without TSS.}
The untyped Workbench assigns one vehicle to two routes, violating a hard constraint.
\end{minipage}
\caption{\textbf{Recorded within-search behavior at disruptive routing updates.} Ten-seed feasibility-gated HV means, standard-deviation bands, and mean stopping generations use the recorded reference. Figure~\ref{fig:public-sequential-restart} reports final quality under the fresh common public-formula references.}
\label{fig:main-large-shift-case}
\par\smallskip{\small\raggedright Curves remain constant after the marked stopping generation. Initial quality and recovery time differ: at t11--t12, Full can start below Warm and finish above it.\par}
\par\vspace{12pt}
\centering
\includegraphics[width=0.98\linewidth]{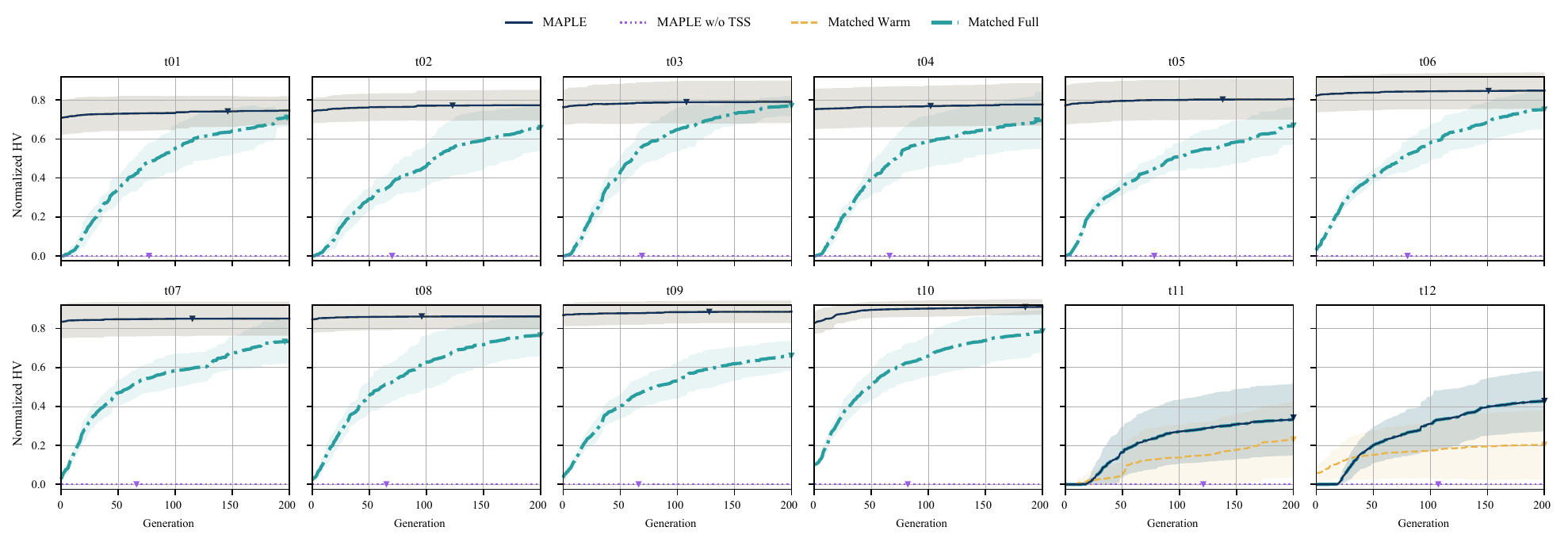}
\caption{\textbf{Green-routing component results (NLDO-P010, t01--t12).} Recorded-evaluator mean HV over ten seeds, standard-deviation bands, and mean stopping generations (triangles). MAPLE and Matched Warm/Full remain feasible. Without TSS, duplicate vehicle assignment makes every P010 update infeasible with zero quality.}
\label{fig:appendix-restart-hv-trajectories}
\end{figure}

\FloatBarrier
\subsection{Static Solving Results}
\label{app:static-solving-results}

Tables~\ref{tab:static-solving-nlp4lp} and~\ref{tab:static-solving-bwor} report the supplementary diagnostics defined in Appendix~\ref{app:static-solving-supplement}.

\providecommand{\StaticSolvingNLPFourLPRows}{%
\method{} static & 33.9\% & 93.2\% & 93.2\% & 59.3\% & 37.3\% & 11,984.2 & 167.8s \\
\rowcolor{LiveOptBaselineRowB}
Persistent ReAct & 37.3\% & 100.0\% & 98.3\% & 62.7\% & 44.1\% & 20,053.1 & 235.7s \\
ReAct & 27.1\% & 91.5\% & 78.0\% & 57.6\% & 32.2\% & 5,813.1 & 72.1s \\
\rowcolor{LiveOptBaselineRowB}
OptiMUS workflow & 69.5\% & 96.6\% & 93.2\% & 69.5\% & -- & 5,034.0 & 84.1s \\
ORLM workflow & 23.7\% & 84.7\% & 61.0\% & 23.7\% & -- & 3,880.1 & 77.0s \\
\rowcolor{LiveOptBaselineRowB}
OptimAI & 47.5\% & 100.0\% & 98.3\% & 61.0\% & 49.2\% & 6,384.3 & 73.3s \\
OR-LLM-Agent workflow & 50.8\% & 96.6\% & 96.6\% & 50.8\% & -- & 6,987.5 & 99.0s \\
}
\providecommand{\StaticSolvingNLPFourLPCompactRows}{%
\rowcolor{LiveOptBaselineRowB}
\method{} static & 33.9\% & 93.2\% & 93.2\% & 59.3\% & 37.3\% \\
Persistent ReAct & 37.3\% & 100.0\% & 98.3\% & 62.7\% & 44.1\% \\
\rowcolor{LiveOptBaselineRowB}
ReAct & 27.1\% & 91.5\% & 78.0\% & 57.6\% & 32.2\% \\
OptiMUS workflow & \textbf{69.5\%} & 96.6\% & 93.2\% & \textbf{69.5\%} & -- \\
\rowcolor{LiveOptBaselineRowB}
ORLM workflow & 23.7\% & 84.7\% & 61.0\% & 23.7\% & -- \\
OptimAI & 47.5\% & \textbf{100.0\%} & \textbf{98.3\%} & 61.0\% & 49.2\% \\
\rowcolor{LiveOptBaselineRowB}
OR-LLM-Agent workflow & 50.8\% & 96.6\% & 96.6\% & 50.8\% & -- \\
}
\providecommand{\StaticSolvingBWORCompactRows}{%
\method{} static & \textbf{80.0\%} & \textbf{100.0\%} & \textbf{100.0\%} & \textbf{80.0\%} & n.a. \\
\rowcolor{LiveOptBaselineRowB}
Persistent ReAct & 70.0\% & \textbf{100.0\%} & \textbf{100.0\%} & 70.0\% & n.a. \\
ReAct & 75.0\% & \textbf{100.0\%} & 95.0\% & 70.0\% & n.a. \\
\rowcolor{LiveOptBaselineRowB}
OptiMUS workflow & 65.0\% & \textbf{100.0\%} & \textbf{100.0\%} & 65.0\% & n.a. \\
ORLM workflow & 45.0\% & 90.0\% & 55.0\% & 45.0\% & n.a. \\
\rowcolor{LiveOptBaselineRowB}
OptimAI & 60.0\% & 80.0\% & 80.0\% & 55.0\% & n.a. \\
OR-LLM-Agent workflow & 65.0\% & 95.0\% & 95.0\% & 65.0\% & n.a. \\
}
\newcommand{\StaticNLPDiagnostics}{%
\rowcolor{LiveOptBaselineRowB}
\method{} static & 33.9\% & 93.2\% & 93.2\% & 59.3\% & 37.3\% & 59.3\% \\
Persistent ReAct & 37.3\% & 100.0\% & 98.3\% & 62.7\% & 44.1\% & 62.7\% \\
\rowcolor{LiveOptBaselineRowB}
ReAct & 27.1\% & 91.5\% & 78.0\% & 57.6\% & 32.2\% & -- \\
OptiMUS workflow & \textbf{69.5\%} & 96.6\% & 93.2\% & \textbf{69.5\%} & -- & 69.5\% \\
\rowcolor{LiveOptBaselineRowB}
ORLM workflow & 23.7\% & 84.7\% & 61.0\% & 23.7\% & -- & 23.7\% \\
OptimAI & 47.5\% & \textbf{100.0\%} & \textbf{98.3\%} & 61.0\% & 49.2\% & -- \\
\rowcolor{LiveOptBaselineRowB}
OR-LLM-Agent workflow & 50.8\% & 96.6\% & 96.6\% & 50.8\% & -- & 50.8\% \\
}
\newcommand{\StaticBWORDiagnostics}{%
\method{} static & \textbf{80.0\%} & \textbf{100.0\%} & \textbf{100.0\%} & \textbf{80.0\%} & n.a. & 80.0\% \\
\rowcolor{LiveOptBaselineRowB}
Persistent ReAct & 70.0\% & \textbf{100.0\%} & \textbf{100.0\%} & 70.0\% & n.a. & 70.0\% \\
ReAct & 75.0\% & \textbf{100.0\%} & 95.0\% & 70.0\% & n.a. & -- \\
\rowcolor{LiveOptBaselineRowB}
OptiMUS workflow & 65.0\% & \textbf{100.0\%} & \textbf{100.0\%} & 65.0\% & n.a. & 65.0\% \\
ORLM workflow & 45.0\% & 90.0\% & 55.0\% & 45.0\% & n.a. & 45.0\% \\
\rowcolor{LiveOptBaselineRowB}
OptimAI & 60.0\% & 80.0\% & 80.0\% & 55.0\% & n.a. & -- \\
OR-LLM-Agent workflow & 65.0\% & 95.0\% & 95.0\% & 65.0\% & n.a. & 65.0\% \\
}
\begin{table}[!htbp]
\centering
\ResultTableSetup
\begin{tabularx}{\linewidth}{@{}l*{6}{Y}@{}}
\toprule
\TableHead
\textbf{Method} & \TightStack{\textbf{Output}\\\textbf{pass}} & \textbf{Compile} & \textbf{Run} & \textbf{Objective} & \TightStack{\textbf{Named}\\\textbf{match}} & \TightStack{\textbf{Joint}\\\textbf{Match}} \\
\midrule
\StaticNLPDiagnostics
\bottomrule
\end{tabularx}
\caption{\textbf{NLP4LP-hard diagnostic checks.} Output pass records each workflow's original acceptance flag. Named-variable agreement is reported separately. Joint Match requires execution and reference-objective/status agreement.}
\label{tab:static-solving-nlp4lp}
\end{table}

\begin{table}[!htbp]
\centering
\ResultTableSetup
\begin{tabularx}{\linewidth}{@{}l*{6}{Y}@{}}
\toprule
\TableHead
\textbf{Method} & \TightStack{\textbf{Output}\\\textbf{pass}} & \textbf{Compile} & \textbf{Run} & \textbf{Objective} & \TightStack{\textbf{Named}\\\textbf{match}} & \TightStack{\textbf{Joint}\\\textbf{Match}} \\
\midrule
\StaticBWORDiagnostics
\bottomrule
\end{tabularx}
\caption{\textbf{BWOR20 diagnostic checks.} Output pass, numeric objective agreement, and joint Match are distinct recorded criteria. Named-variable matching does not apply.}
\label{tab:static-solving-bwor}
\end{table}

The two static benchmarks expose different weaknesses.
On BWOR20, all 20 \method{} rows compile and run, and 16 match the reference objective.
On NLP4LP-hard, 55 of 59 rows compile and run, 35 match the objective, and 20 pass the additional historical output check.

\subsection{Reference and Archive Sensitivity}
\label{app:reference-archive-sensitivity}

\paragraph{Sensitivity to the reference set and evaluation box.}
We vary the reference set and evaluation box and recompute the reported HV ratios.

Table~\ref{tab:reference-sensitivity} examines two choices that remain when the Pareto reference is finite.
For reference subsampling, each method uses the same fixed subset in each of 20 repeats.
The evaluation box from the full reference remains fixed.
For box sensitivity, both numerator and denominator are recomputed after expanding the ideal-to-reference span.

Under the recorded evaluator, five w/o-TSS outputs at P012-t01 exceed HV ratio 1. Expanding the HV box reduces the number of ratios above 1 from five to four and then one. The 500-member archive cap is inactive for the MAPLE results in this analysis.

\begin{table}[!htbp]
\centering
\ResultTableSetup
\renewcommand{\arraystretch}{1.55}
\begin{tabularx}{\linewidth}{@{}l*{6}{Y}@{}}
\toprule
\TableHead
& \multicolumn{3}{c}{\textbf{Reference subsampling: $\Delta$HV ratio}} & \multicolumn{3}{c}{\textbf{HV-box span: mean ratio}} \\
\cmidrule(lr){2-4}\cmidrule(lr){5-7}
\TableHead
\textbf{Method} & \textbf{75\%} & \textbf{50\%} & \textbf{25\%} & \textbf{$1.00\times$} & \textbf{$1.10\times$} & \textbf{$1.25\times$} \\
\midrule
\LiveOptReferenceSensitivityRows
\bottomrule
\end{tabularx}
\caption{\textbf{Sensitivity to the finite Pareto reference on P010--P015, t01--t12.} Recorded-evaluator results. Each method contributes 720 time-step--seed results. The Warm and Full rows follow their own populations through the sequence. Reference subsampling reports the change from the full-reference mean. The box columns recompute both volumes. Method ordering is unchanged in every setting. For \mbox{MAPLE w/o TSS}, the number of results above one changes from 5 to 4 to 1 as the box expands. The maximum changes from 1.020 to 1.014 to 1.008.}
\label{tab:reference-sensitivity}
\end{table}

\paragraph{Effect of the Pareto archive limit.}
We repeat the search with archive limits of 100, 200, and 500 while keeping the other settings fixed.

\begin{table}[!htbp]
\centering
\ResultTableSetup
\renewcommand{\arraystretch}{1.55}
\begin{tabularx}{\linewidth}{@{}*{7}{Y}@{}}
\toprule
\TableHead
\textbf{Cap} & \textbf{P014 HV} & \textbf{P015 HV} & \textbf{Mean HV} & \textbf{IGD} & \textbf{Gen.} & \textbf{Active/results} \\
\midrule
\LiveOptArchiveCapSensitivityRows
\bottomrule
\end{tabularx}
\caption{\textbf{Sensitivity to the archive limit on P014/P015 under the recorded evaluator.} Each row contains the same 240 update--seed results. The restart choices, $200\times200$ search budget, stopping rule, and held-out reference remain fixed. Only the maximum archive size changes.}
\label{tab:archive-cap-sensitivity}
\end{table}

Caps 100 and 200 truncate the archive in 240 and 162 results, respectively, while cap 500 is inactive.
Their mean HV ratios are 0.775, 0.776, and 0.774.
The largest change in the mean is 0.0016.
Mean IGD is 0.139, 0.139, and 0.138.
The smaller limits do change individual fronts.
Their mean absolute HV differences from limit 500 are 0.053 and 0.040.
They also use 5.3 and 2.8 more generations on average.
The mean is stable, but archive size can still matter at an individual time step.

\paragraph{Sensitivity to restart thresholds.}
\label{app:restart-threshold-sensitivity}
We vary the numerical-change threshold and identify the interval that leaves the table-change decisions unchanged.
The numerical selector uses a change threshold of 0.5 in the reported comparison.
For thresholds of 0.25, 0.50, and 0.75, the numerical selector obtains HV ratios of 0.755, 0.765, and 0.767, with Full rates of 48.8\%, 34.0\%, and 25.0\%. All three HV ratios are below MAPLE's 0.779 under the recorded evaluator.
At threshold 0.50 it uses 169.4 generations, compared with 162.7 for Matched Warm and 162.8 for MAPLE.
The table-change threshold lies in a ten-fold gap.
Ordinary updates score 0--0.060, while disruptive updates score 0.705--1.000.
Any value in $[0.3, 0.7]$ therefore makes identical decisions.
Tables~\ref{tab:archive-cap-sensitivity} and~\ref{tab:reference-sensitivity} separately test the archive limit and Pareto reference.

\paragraph{Sensitivity to congestion and return-deadline checks.}
We separately recompute scores after changing the congestion scope and enforcing the return deadline.
Under the recorded reference, extending congestion to adjacent departure/return legs changes routing HV from 0.7471 to 0.7466 without changing MAPLE's ordering relative to Warm/Full. Four saved candidate occurrences violate the return deadline. Enforcing this deadline leaves the primary aggregate HV unchanged.

\FloatBarrier
\subsection{Using MAPLE through an Interactive Interface}
\label{app:harness-demo}

\begin{figure}[!htbp]
\centering
\includegraphics[width=\linewidth]{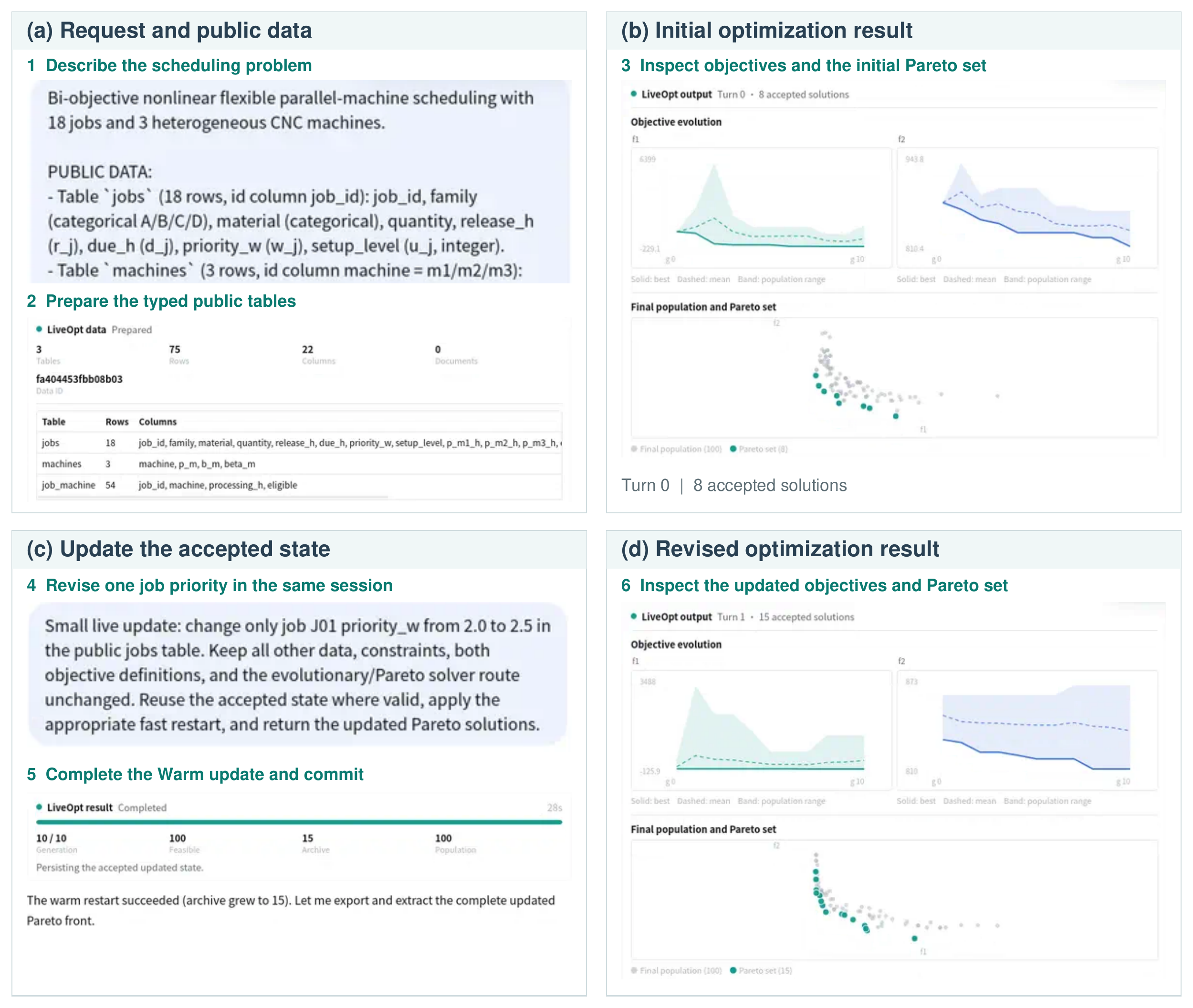}
\caption{\textbf{A complete Harness interaction from request to accepted update.} Recorded steps: (a) scheduling request and public tables; (b) initial traces and eight Pareto solutions; (c) J01's priority changes from 2.0 to 2.5, followed by a Warm update; (d) revised traces and fifteen Pareto solutions.}
\label{fig:harness-case-explorer}
\end{figure}

The interface lets users submit a planning request, inspect the solutions, and revise the same problem. The application is called MAPLE Harness. Each session retains the Workbench, candidate plans, objective traces, and restart actions across updates.
Figure~\ref{fig:harness-case-explorer} shows a CNC scheduling session from the published walkthrough, separate from NLDO.

The interface sends planning requests to a separate service that runs MAPLE through the Model Context Protocol (MCP). Each request returns a job identifier, which the interface uses to display search progress. After the result is accepted, the service saves the updated session. Users can inspect the solutions and export them as CSV or JSON.

\FloatBarrier
\subsection{Prompt Interfaces and Execution Rules}
\label{app:prompts-contracts}

This section summarizes the recorded prompt interfaces, required output fields, and state-transfer rules. The TSS prompts specify two editable functions and the required table format. The other interfaces specify the functions used by search. Field names and response delimiters below retain their recorded spelling.
Problem text, public tables, update history, previous output, and repair messages fill the variable fields.
Table~\ref{tab:appendix-external-baseline-source-basis} defines the carried-state interfaces.

\begingroup
\let\RecordedFootnoteSize\footnotesize

\renewcommand{\footnotesize}{\RecordedFootnoteSize\setlength{\baselineskip}{12.0pt}}

\setlength{\OuterFrameSep}{\topsep}
\addtolength{\OuterFrameSep}{-0.1pt}

\newcommand{\PromptField}[2]{\par\addvspace{0.10em}\noindent\textbf{\textcolor{LiveOptNavy}{#1.}}\quad#2}
\newenvironment{PromptBody}{%
  \def\FrameCommand{%
    {\color{LiveOptTeal}\vrule width 2.2pt}%
    \hspace{5pt}%
    \colorbox{LiveOptPale}%
  }%
  \MakeFramed{\advance\hsize-\width\FrameRestore}%
  \footnotesize\raggedright\sloppy
  \clubpenalty=10000 \widowpenalty=10000
}{\endMakeFramed}
\newcommand{\PromptBox}[3]{%
  \par\addvspace{0.25em}%
  \begin{PromptBody}%
    \begingroup\setlength{\fboxsep}{2.6pt}%
    \noindent\colorbox{LiveOptNavy}{%
      \parbox{\dimexpr\hsize-2\fboxsep\relax}{\color{white}\footnotesize\textbf{#1}\hfill\textsf{#2}}%
    }\endgroup%
    \par\nobreak
    #3%
  \end{PromptBody}%
  \par\addvspace{0.15em}%
}

\PromptBox{\method{} TSS Workbench builder}{initialization prompt}{
\PromptField{User task}{The prompt shows a complete runnable Workbench, then asks the model to make the smallest necessary edits for the public problem. The two editable surfaces are \texttt{build\_problem(public\_context)} and \texttt{evaluate(genome, data)}.}
\PromptField{Framework-owned code}{Main loop, population initialization, GA/NSGA-II selection, LP/MILP solving, segment-specific crossover/mutation, nondominated sorting, crowding distance, archive tracking, and penalty-result construction.}
\PromptField{Required response}{Exactly two slot blocks. The decision-definition function receives fixed-code-normalized public tables, builds reusable data views, declares typed decision segments, chooses the solver mode, and lists objective names. The evaluation function evaluates one genome by decoding the candidate, computing public violations, scalar rank, diagnostics, and optional objective vector.}
\PromptField{Rules}{Do not generate a separate model schema, GA/NSGA code, a large data contract, or unused helper code. Do not hard-code entity ids, resource names, source ids, hidden references, or benchmark-specific constants. Loop over public table rows. Submitted solution identifiers must be scalar public IDs, not full row objects.}
\PromptField{Public task definition}{If \texttt{objective\_mode=multi\_objective}, the prompt requires \texttt{solver\_mode="moea"} and objective names/order exactly matching the public task definition. Scalarization and literal constant Pareto dimensions are forbidden.}
\PromptField{Repair turn}{If compilation or smoke execution fails, the next prompt includes the previous public error plus the previous slot contents and asks for corrected Workbench slots.}
}

\PromptBox{\method{} update localizer}{dynamic classification prompt}{
\PromptField{System}{\texttt{Classify the dynamic update for a fixed LiveOpt Workbench. Return exactly one JSON object.}}
\PromptField{Input}{Natural-language update, current segment signature, current public-context summary, representative accepted decision, public event/reference ledger, and allowed restart-skill names. The search population and archive are not shown.}
\PromptField{Required JSON fields}{\texttt{data\_update}, \texttt{patch\_setup}, \texttt{patch\_fitness}, \texttt{restart\_skill}, \texttt{reason}, and \texttt{state\_binding\_queries}.}
\PromptField{Decision rules}{Use \texttt{patch\_setup} only when search segments or data extraction must change. Use \texttt{patch\_fitness} when objectives, constraints, penalties, decode, or solution reporting must change. If public tables changed but the existing Workbench slots loop over \texttt{public\_context}, prefer \texttt{data\_update=true} and \texttt{patch\_setup=false}.}
\PromptField{State binding}{For an explicit cross-stage assignment hold, return a typed \texttt{lock\_assignment}, \texttt{preserve\_demand\_assignments}, or \texttt{preserve\_resource\_assignments} query and leave the code-patch flags false unless another part of the request changes the Workbench. Fixed code resolves accepted assignment values and checks them against the current typed segment.}
\PromptField{Restart hint}{One allowed restart identifier for recording the choice. The active run replaces it with the separately checked semantic Warm--Full selector.}
}

\PromptBox{\method{} semantic restart selector}{reuse-risk assessment prompt}{
\PromptField{System}{\texttt{Assess whether repaired solutions near the previous population are likely to remain competitive.}}
\PromptField{Input}{An opaque update id, the public update text, and a fixed summary containing changed public field paths, representative row changes, old/new decision types, and old/new public objective declarations. The selector also keeps the initial request and earlier accepted public updates in its conversation context.}
\PromptField{Required JSON}{\texttt{reuse\_risk}, \texttt{change\_mechanisms}, \texttt{full\_vote}, \texttt{evidence\_paths}, and \texttt{reason}. Evidence paths must exactly match the digest.}
\PromptField{Decision rule}{Warm already mixes at most 50\% repaired history with fresh candidates. Full is reserved for a supported replacement of key decisions, reversal of objective preferences or resource roles, new constraint system, or risk that good solutions move far from the old population. Ordinary changes and uncertainty return Warm.}
\PromptField{Boundary}{The prompt explicitly forbids trial optimization, benchmark/time-step identity, difficulty labels, search-action advice in the event text, hidden references, HV, and IGD. The gate accepts Full when the assessment reports high reuse risk, casts a Full vote, includes a structurally supported mechanism, and cites at least one changed field path.}
}

\PromptBox{\method{} public-table updater}{natural language to public table patch}{
\PromptField{System}{\texttt{Patch public optimization data from a natural-language update. Return exactly one JSON object.}}
\PromptField{Writable root}{Only \texttt{public\_context["tables"]}. Every table is a \texttt{list[dict]}. Singleton tables such as policy, settings, depot, or global parameters are still one-row lists.}
\PromptField{Operations}{\texttt{update\_row}, \texttt{append\_row}, \texttt{delete\_row}, \texttt{set\_value}, \texttt{replace\_value}, and \texttt{replace\_table}.}
\PromptField{Required shape}{A compact JSON object with \texttt{operations}, where each operation records \texttt{op}, \texttt{path}, optional \texttt{match}, \texttt{row}/\texttt{value}/\texttt{values}, and a public \texttt{reason}.}
\PromptField{Repair turn}{If an operation misses a row, has the wrong path, or violates the list-of-dicts ABI, the next prompt includes the error and previous patch JSON and asks for a corrected patch.}
}

\PromptBox{\method{} TSS Workbench editor}{slot-level code patch}{
\PromptField{System}{\texttt{Patch only the requested Workbench slots. Return code blocks only.}}
\PromptField{Input}{Current Workbench functions, updated public context, public task definition, predicted changes, restart choice selected outside code, and previous validation error if any.}
\PromptField{Slot rule}{Return only requested slots. If public data extraction/search segments are unaffected, do not edit the decision-definition function. If candidate evaluation is unaffected, do not edit the evaluation function. Restart logic is never implemented inside the slots.}
\PromptField{Output rule}{Preserve solver mode, objective names, and objective-vector order required by the public task definition. If new public policy values appear, pass them through the data reader and read them in the evaluator. Do not hard-code update constants.}
\PromptField{Validation}{Returned functions are compiled, run on a small public example, and checked against the public task definition before they are accepted.}
}

\PromptBox{\method{} w/o TSS}{untyped Workbench construction}{
\PromptField{System}{\texttt{Build a complete persistent Python optimization artifact from scratch.}}
\PromptField{Input}{The public natural-language problem and normalized public context. No TSS slot, \texttt{SegmentSpec}, candidate schema, encoding hint, or operator strategy is supplied.}
\PromptField{Required callbacks}{\texttt{build\_state}, \texttt{initialize}, \texttt{evaluate\_decode}, \texttt{crossover}, \texttt{mutate}, and \texttt{coerce\_repair}. The model chooses the representation and implements every callback.}
\PromptField{Fixed optimization loop}{Selection, nondominated sorting, archive maintenance, early stopping, restart selection, and the evolutionary loop remain fixed and cannot be reimplemented by the generated program.}
\PromptField{Update and repair}{Later prompts provide the complete accepted artifact, the new public update/context, and any compile or smoke-test error, then request one complete replacement Python block.}
}

\PromptBox{External baseline time-step prompts}{public-stage instructions}{
\PromptField{User task}{\texttt{Solve the current NLDO stage from public information only.} The prompt then gives the initial request, public tables, every public update through the selected time step, the public task definition, and the required solution fields.}
\PromptField{State boundary}{The saved source contains 127 initial or early rows based on each original method and 135 continuation rows. Continuation receives only that method's own dialogue, preceding accepted output, and candidate population. No row receives a LiveOpt population, TSS state, restart decision, hidden change, reference archive, or hidden objective/checker implementation.}
\PromptField{Executable output}{Every method returns strict JSON containing executable Python solver code and a final solution. A method may also return a candidate archive on Pareto tasks. The runner compiles the code, executes it on the public tables, and parses its JSON output.}
\PromptField{Repair}{Compile, runtime, or output-schema failures receive at most three code-output repairs. The repair prompt includes public execution feedback but never hidden feasibility or reference quality.}
}

\PromptBox{ReAct+Public Tools}{instruction based on the original method}{
\PromptField{System}{\texttt{You are a ReAct+Tools optimization agent. Use a Thought -> public tool-plan -> validation pattern, then return executable solver code and the final answer. Return JSON only.}}
\PromptField{Required fields}{\texttt{thought}, \texttt{react\_trace}, \texttt{tool\_plan}, \texttt{solver\_code}, \texttt{validation\_plan}, and \texttt{final\_answer}.}
}

\PromptBox{Persistent ReAct}{stateful public baseline}{
\PromptField{System}{\texttt{You are a Persistent ReAct+Tools optimization agent.} Same Thought $\rightarrow$ tool-plan $\rightarrow$ validation pattern and required fields as ReAct+Public Tools, plus: \texttt{You keep a persistent code workspace across stages\ldots{} Apply the smallest necessary edits for the new public update while keeping parts that are still valid\ldots{} Always return the COMPLETE updated solver code, never a diff. You have no TSS workbench slots, no candidate population or archive, no restart interface, and no LiveOpt artifacts.}}
}

\PromptBox{OptiMUS and ORLM}{modeling and code instructions based on the original methods}{
\PromptField{OptiMUS system}{\texttt{You are an OptiMUS-style optimization code generator. Produce Gurobi solver code and debug it. Return JSON only.}}
\PromptField{ORLM system}{\texttt{You are an ORLM-style optimization modeler. Produce a mathematical model and COPT solver code. Return JSON only.}}
\PromptField{Required fields}{\texttt{formulation}, \texttt{solver\_choice}, \texttt{solver\_plan}, \texttt{solver\_code}, \texttt{validation\_plan}, and \texttt{final\_answer}. Public execution uses the available local solver route rather than exposing evaluation-only code.}
}

\PromptBox{OptimAI}{multi-role instruction based on the original method}{
\PromptField{System}{\texttt{You are an OptimAI-style optimization agent. Formulate the task, propose multiple solver-code plans, select a plan, and include a debug/validation plan. Return JSON only.}}
\PromptField{Required fields}{\texttt{formulation}, \texttt{candidate\_solver\_plans}, \texttt{selected\_plan}, \texttt{solver\_code}, \texttt{debug\_validation\_plan}, and \texttt{final\_answer}.}
}

\PromptBox{OR-LLM-Agent}{executed OR-LLM-Agent instruction}{
\PromptField{System}{\texttt{You are an OR-LLM-Agent-style agent. Produce reasoning/modeling, formulation, code-generation, self-verification, and self-repair artifacts for the public optimization task. Return JSON only.}}
\PromptField{Required fields}{\texttt{reasoning\_modeling\_trace}, \texttt{formulation}, \texttt{code\_generation\_plan}, \texttt{solver\_code}, \texttt{self\_verification}, \texttt{self\_repair}, and \texttt{final\_answer}.}
}
\endgroup

Static adapters use the shared DeepSeek model with COPT for ORLM and Gurobi for OptiMUS and OR-LLM-Agent.
For the five non-persistent external NLDO workflows, 127 initial/early rows use method-based prompts and 135 continuation rows expose their own preceding outputs and candidates when available. Sixteen earlier repair rows also contain their own preceding accepted output.


\begingroup
\let\RecordedFootnoteSize\footnotesize
\renewcommand{\footnotesize}{\RecordedFootnoteSize\setlength{\baselineskip}{10.90pt}}
\subsection{Recorded Public NLDO Requests and Updates}
\label{app:complete-nldo-public-cases}

This section reproduces the business-language requests from the saved executions, before their shared evaluation reminder.
The annotations describe how each update changes the evaluated problem. F marks a change to feasibility. O marks objective changes that can alter the ordering of candidate solutions. R marks positive objective rescaling that preserves dominance. D marks changes to reported diagnostics only. U marks an unchanged mathematical problem.
M independently marks a historical-reference requirement and can overlap any semantic label.
The labels do not enter agent prompts. Table~\ref{tab:appendix-update-strata} aggregates these audited labels.
Exact reproduction uses the original machine-readable requests, public tables, output contracts, and checksums.

\begingroup
\footnotesize
\setlength{\parskip}{0pt}
\sloppy
\NLDOEpisodeHead{NLDO-P001 (NLDO; cobench)}
\noindent\textbf{Initial public request.} The instance data is provided in attached tables; treat the listed rows and columns as the source of truth. Select a feasible subset or portfolio under fixed capacity. Later updates revise public item values but do not change capacity, item weights, or eligibility fields. The constraints table gives the current objective\_mode. Expose a minimization scalar suitable for the generic solver. Select a feasible portfolio under fixed capacity 34. Items: I01 value 15 weight 7, I02 value 22 weight 3, I03 value 10 weight 8, I04 value 17 weight 4, I05 value 24 weight 9, I06 value 12 weight 5, I07 value 19 weight 10, I08 value 26 weight 6, I09 value 14 weight 2, I10 value 21 weight 7, I11 value 9 weight 3, I12 value 16 weight 8. Maximize value; later natural-language updates may revise public item values but do not change capacity or eligibility.

\noindent\textbf{Public updates.}
\begin{NLDOUpdateList}
\item[\NLDOUpdateLabel{t01}{O}] A sponsor review increases item I03's public value by 6. The capacity and item eligibility rules stay unchanged.
\item[\NLDOUpdateLabel{t02}{O}] A risk review lowers item I07's public value by 5; it remains selectable under the same capacity rule.
\item[\NLDOUpdateLabel{t03}{O}] Quality audit increases item I10's public value by 4 while the feasible portfolio definition remains fixed.
\item[\NLDOUpdateLabel{t04}{O,M}] The item praised in the sponsor review receives another value increase of 6; infer which item that was from memory.
\item[\NLDOUpdateLabel{t05}{O,M}] The item from the risk review recovers 3 value points, while the sponsor-reviewed item gains 1 more point; resolve both items from memory.
\item[\NLDOUpdateLabel{t06}{U}] Operations asks for a checkpoint only: keep the current public objective and hard constraints, revalidate the plan, and change it only if it is no longer feasible.
\item[\NLDOUpdateLabel{t07}{U}] No new business event is reported for this stage. Refresh the solution against the current public tables.
\item[\NLDOUpdateLabel{t08}{U}] Treat this as a routine review. The model, objective terms, and public rows stay as they are; preserve feasibility under the current state.
\item[\NLDOUpdateLabel{t09}{U}] The planning team wants the current public contract revalidated without adding new constraints.
\item[\NLDOUpdateLabel{t10}{O}] A late value-model reset changes several public item scores without changing capacity or eligibility. Item I10 receives a value bonus of 10, item I11 receives a value penalty of 8, and item I05 receives a value bonus of 10. Rebuild the portfolio under the same feasible subset definition rather than carrying forward the old value ranking.
\item[\NLDOUpdateLabel{t11}{U}] Operations asks for a checkpoint only: keep the current public objective and hard constraints, revalidate the plan, and change it only if it is no longer feasible.
\item[\NLDOUpdateLabel{t12}{U}] No new business event is reported for this stage. Refresh the solution against the current public tables.
\end{NLDOUpdateList}

\NLDOEpisodeHead{NLDO-P002 (NLDO; cobench)}
\noindent\textbf{Initial public request.} The instance data is provided in attached tables; treat the listed rows and columns as the source of truth. Choose open facilities and assign each customer to an open facility using only public facilities, customers, and constraints tables. Respect fixed facility capacity; later updates revise public opening/service costs but do not change capacity, demand, or facility eligibility. The constraints table gives the current objective\_mode. Expose a minimization scalar suitable for the generic solver. Choose facilities and assign customers to open facilities. Facilities: F1 open cost 19 capacity 22, F2 open cost 23 capacity 26, F3 open cost 27 capacity 30, F4 open cost 18 capacity 18, F5 open cost 22 capacity 22. Customers: C01 demand 3, C02 demand 4, C03 demand 5, C04 demand 6, C05 demand 2, C06 demand 3, C07 demand 4, C08 demand 5, C09 demand 6, C10 demand 2. Minimize opening plus service cost; later public updates may revise costs while capacities, customer demand, and eligibility stay fixed.

\noindent\textbf{Public updates.}
\begin{NLDOUpdateList}
\item[\NLDOUpdateLabel{t01}{O}] Facility F2's opening cost increases by 4 after an energy-price review; capacity and facility eligibility stay unchanged.
\item[\NLDOUpdateLabel{t02}{O}] Serving customer C03 from facility F1 becomes 3 cost units cheaper per demand unit after a local contract update.
\item[\NLDOUpdateLabel{t03}{O}] Facility F5's opening cost decreases by 5 because of a regional service credit.
\item[\NLDOUpdateLabel{t04}{O,M}] The facility from the regional service credit receives one more opening-cost discount of 2; infer that facility from memory.
\item[\NLDOUpdateLabel{t05}{O,M}] The customer-facility pair from the local contract update receives another 1 unit service-cost discount, while the facility from the energy-price review gets 1 more opening-cost penalty; resolve both references from memory.
\item[\NLDOUpdateLabel{t06}{U}] Operations asks for a checkpoint only: keep the current public objective and hard constraints, revalidate the plan, and change it only if it is no longer feasible.
\item[\NLDOUpdateLabel{t07}{U}] No new business event is reported for this stage. Refresh the solution against the current public tables.
\item[\NLDOUpdateLabel{t08}{U}] Treat this as a routine review. The model, objective terms, and public rows stay as they are; preserve feasibility under the current state.
\item[\NLDOUpdateLabel{t09}{U}] The planning team wants the current public contract revalidated without adding new constraints.
\item[\NLDOUpdateLabel{t10}{O}] A same-day service-level reset flips the scalar objective. The public facility rows, customer rows, demand values, and capacity limits remain unchanged, but stop minimizing monetary opening plus service cost. The new objective is to minimize total service time using the public serve\_time columns, even if that requires a more expensive assignment.
\item[\NLDOUpdateLabel{t11}{U}] Operations asks for a checkpoint only: keep the current public objective and hard constraints, revalidate the plan, and change it only if it is no longer feasible.
\item[\NLDOUpdateLabel{t12}{U}] No new business event is reported for this stage. Refresh the solution against the current public tables.
\end{NLDOUpdateList}

\NLDOEpisodeHead{NLDO-P003 (NLDO; cobench)}
\noindent\textbf{Initial public request.} The instance data is provided in attached tables; treat the listed rows and columns as the source of truth. Select a minimum-cost set collection using only public sets, elements, and constraints tables. Cover the fixed active element universe; later updates revise public set costs but do not change coverage requirements or set eligibility. Expose a minimization scalar suitable for the generic solver. Select a minimum-cost collection of public sets that covers the fixed active element universe. Elements: E01, E02, E03, E04, E05, E06, E07, E08, E09, E10. Sets: S01 cost 9 covers E02/E05/E07/E08, S02 cost 14 covers E01/E04/E07/E10, S03 cost 6 covers E03/E06/E07/E09, S04 cost 11 covers E02/E05/E07/E08, S05 cost 16 covers E01/E04/E07/E10, S06 cost 8 covers E01/E02/E03/E04/E05/E06/E07/E08/E09/E10, S07 cost 13 covers E02/E05/E07/E08, S08 cost 5 covers E01/E04/E07/E10. Later updates may revise public set costs but do not add/remove coverage requirements or set eligibility.

\noindent\textbf{Public updates.}
\begin{NLDOUpdateList}
\item[\NLDOUpdateLabel{t01}{O}] Set S03 receives a public service-quality rebate of 3 cost units; all element coverage requirements stay unchanged.
\item[\NLDOUpdateLabel{t02}{O}] Set S06 becomes 2 cost units cheaper because it carries a compliance certificate.
\item[\NLDOUpdateLabel{t03}{O}] Provider audit increases set S02's public cost by 4, but the set remains available and the coverage universe is unchanged.
\item[\NLDOUpdateLabel{t04}{O,M}] The set carrying the compliance certificate becomes 3 cost units cheaper; infer which set that was from memory.
\item[\NLDOUpdateLabel{t05}{O,M}] The audited set receives a 1 cost-unit correction, and the high-priority rebate set receives another 2 unit discount; resolve both sets from memory.
\item[\NLDOUpdateLabel{t06}{U}] Operations asks for a checkpoint only: keep the current public objective and hard constraints, revalidate the plan, and change it only if it is no longer feasible.
\item[\NLDOUpdateLabel{t07}{U}] No new business event is reported for this stage. Refresh the solution against the current public tables.
\item[\NLDOUpdateLabel{t08}{U}] Treat this as a routine review. The model, objective terms, and public rows stay as they are; preserve feasibility under the current state.
\item[\NLDOUpdateLabel{t09}{U}] The planning team wants the current public contract revalidated without adding new constraints.
\item[\NLDOUpdateLabel{t10}{U}] Run a maintenance update with the same public contract. Revalidate feasibility and quality under the current public state.
\item[\NLDOUpdateLabel{t11}{U}] Operations asks for a checkpoint only: keep the current public objective and hard constraints, revalidate the plan, and change it only if it is no longer feasible.
\item[\NLDOUpdateLabel{t12}{U}] No new business event is reported for this stage. Refresh the solution against the current public tables.
\end{NLDOUpdateList}

\NLDOEpisodeHead{NLDO-P004 (NLDO; fjsp)}
\noindent\textbf{Initial public request.} The instance data is provided in attached tables; treat the listed rows and columns as the source of truth. Create a flexible job-shop schedule. Assign every operation to one eligible resource, respect fixed job precedence and resource capacity, and use settings.objective\_mode for the current scalar objective. Later updates may revise soft due/priority values or machine cost\_rate but do not change machine availability, operation eligibility, or processing requirements. Plan-change disruption is reported only as a diagnostic and never defines hard feasibility. Schedule flexible job-shop operations with precedence and machine capacity constraints. Minimize weighted tardiness; report disruption only as an auxiliary diagnostic. J1 due 28: J1O1 can run on M3/M1 for 9 minutes, J1O2 can run on M1/M2 for 4 minutes, J1O3 can run on M2/M3 for 6 minutes J2 due 32: J2O1 can run on M1/M2 for 5 minutes, J2O2 can run on M2/M3 for 7 minutes, J2O3 can run on M3/M1 for 9 minutes J3 due 36: J3O1 can run on M2/M3 for 8 minutes, J3O2 can run on M3/M1 for 10 minutes, J3O3 can run on M1/M2 for 5 minutes

\noindent\textbf{Public updates.}
\begin{NLDOUpdateList}
\item[\NLDOUpdateLabel{t01}{D}] Machine M1's public cost\_rate increases by 1. All machines remain available and operation eligibility is unchanged.
\item[\NLDOUpdateLabel{t02}{O}] J1 becomes urgent: set priority to 3 and tighten its soft due time by 5 minutes. This changes tardiness cost but not schedule feasibility.
\item[\NLDOUpdateLabel{t03}{O,M}] For the job that became urgent in the previous note, increase its priority by 1 again; infer the job from memory.
\item[\NLDOUpdateLabel{t04}{O,M}] The same urgent job receives a 3-minute soft due-time relaxation after customer confirmation; infer the job from memory.
\item[\NLDOUpdateLabel{t05}{D,M}] The machine from the earliest cost-rate note receives another 0.5 cost-rate increase; infer it from memory and re-optimize the same feasible schedule space.
\item[\NLDOUpdateLabel{t06}{U}] Operations asks for a checkpoint only: keep the current public objective and hard constraints, revalidate the plan, and change it only if it is no longer feasible.
\item[\NLDOUpdateLabel{t07}{U}] No new business event is reported for this stage. Refresh the solution against the current public tables.
\item[\NLDOUpdateLabel{t08}{U}] Treat this as a routine review. The model, objective terms, and public rows stay as they are; preserve feasibility under the current state.
\item[\NLDOUpdateLabel{t09}{U}] The planning team wants the current public contract revalidated without adding new constraints.
\item[\NLDOUpdateLabel{t10}{O}] A late accounting reset flips the scalar objective. All machines remain available, and due dates remain soft diagnostics. Stop minimizing tardiness; the new objective is to minimize total machine operating cost using each machine's public cost\_rate. This may prefer cheaper machines even when the schedule finishes later.
\item[\NLDOUpdateLabel{t11}{U}] Operations asks for a checkpoint only: keep the current public objective and hard constraints, revalidate the plan, and change it only if it is no longer feasible.
\item[\NLDOUpdateLabel{t12}{U}] No new business event is reported for this stage. Refresh the solution against the current public tables.
\end{NLDOUpdateList}

\NLDOEpisodeHead{NLDO-P005 (NLDO; fjsp)}
\noindent\textbf{Initial public request.} The instance data is provided in attached tables; treat the listed rows and columns as the source of truth. Create a flexible job-shop schedule. Assign every operation to one eligible resource, respect fixed job precedence and resource capacity, and use settings.objective\_mode for the current scalar objective. Later updates may revise soft due/priority values or machine cost\_rate but do not change machine availability, operation eligibility, or processing requirements. Plan-change disruption is reported only as a diagnostic and never defines hard feasibility. Schedule flexible job-shop operations with precedence and machine capacity constraints. Minimize weighted tardiness; report disruption only as an auxiliary diagnostic. J1 due 29: J1O1 can run on M1/M2 for 10 minutes, J1O2 can run on M2/M3 for 5 minutes, J1O3 can run on M3/M1 for 7 minutes J2 due 33: J2O1 can run on M2/M3 for 6 minutes, J2O2 can run on M3/M1 for 8 minutes, J2O3 can run on M1/M2 for 10 minutes J3 due 37: J3O1 can run on M3/M1 for 9 minutes, J3O2 can run on M1/M2 for 4 minutes, J3O3 can run on M2/M3 for 6 minutes

\noindent\textbf{Public updates.}
\begin{NLDOUpdateList}
\item[\NLDOUpdateLabel{t01}{D}] Machine M2's public cost\_rate increases by 2. All machines remain available and operation eligibility is unchanged.
\item[\NLDOUpdateLabel{t02}{O}] J2 becomes urgent: set priority to 3 and tighten its soft due time by 5 minutes. This changes tardiness cost but not schedule feasibility.
\item[\NLDOUpdateLabel{t03}{O,M}] For the job that became urgent in the previous note, increase its priority by 1 again; infer the job from memory.
\item[\NLDOUpdateLabel{t04}{O,M}] The same urgent job receives a 3-minute soft due-time relaxation after customer confirmation; infer the job from memory.
\item[\NLDOUpdateLabel{t05}{D,M}] The machine from the earliest cost-rate note receives another 0.5 cost-rate increase; infer it from memory and re-optimize the same feasible schedule space.
\item[\NLDOUpdateLabel{t06}{U}] Operations asks for a checkpoint only: keep the current public objective and hard constraints, revalidate the plan, and change it only if it is no longer feasible.
\item[\NLDOUpdateLabel{t07}{U}] No new business event is reported for this stage. Refresh the solution against the current public tables.
\item[\NLDOUpdateLabel{t08}{U}] Treat this as a routine review. The model, objective terms, and public rows stay as they are; preserve feasibility under the current state.
\item[\NLDOUpdateLabel{t09}{U}] The planning team wants the current public contract revalidated without adding new constraints.
\item[\NLDOUpdateLabel{t10}{U}] Run a maintenance update with the same public contract. Revalidate feasibility and quality under the current public state.
\item[\NLDOUpdateLabel{t11}{U}] Operations asks for a checkpoint only: keep the current public objective and hard constraints, revalidate the plan, and change it only if it is no longer feasible.
\item[\NLDOUpdateLabel{t12}{U}] No new business event is reported for this stage. Refresh the solution against the current public tables.
\end{NLDOUpdateList}

\NLDOEpisodeHead{NLDO-P006 (NLDO; fjsp)}
\noindent\textbf{Initial public request.} The instance data is provided in attached tables; treat the listed rows and columns as the source of truth. Create a flexible job-shop schedule. Assign every operation to one eligible resource, respect fixed job precedence and resource capacity, and use settings.objective\_mode for the current scalar objective. Later updates may revise soft due/priority values or machine cost\_rate but do not change machine availability, operation eligibility, or processing requirements. Plan-change disruption is reported only as a diagnostic and never defines hard feasibility. Schedule flexible job-shop operations with precedence and machine capacity constraints. Minimize weighted tardiness; report disruption only as an auxiliary diagnostic. J1 due 30: J1O1 can run on M2/M3 for 4 minutes, J1O2 can run on M3/M1 for 6 minutes, J1O3 can run on M1/M2 for 8 minutes J2 due 34: J2O1 can run on M3/M1 for 7 minutes, J2O2 can run on M1/M2 for 9 minutes, J2O3 can run on M2/M3 for 4 minutes J3 due 38: J3O1 can run on M1/M2 for 10 minutes, J3O2 can run on M2/M3 for 5 minutes, J3O3 can run on M3/M1 for 7 minutes

\noindent\textbf{Public updates.}
\begin{NLDOUpdateList}
\item[\NLDOUpdateLabel{t01}{D}] Machine M3's public cost\_rate increases by 3. All machines remain available and operation eligibility is unchanged.
\item[\NLDOUpdateLabel{t02}{O}] J3 becomes urgent: set priority to 3 and tighten its soft due time by 5 minutes. This changes tardiness cost but not schedule feasibility.
\item[\NLDOUpdateLabel{t03}{O,M}] For the job that became urgent in the previous note, increase its priority by 1 again; infer the job from memory.
\item[\NLDOUpdateLabel{t04}{O,M}] The same urgent job receives a 3-minute soft due-time relaxation after customer confirmation; infer the job from memory.
\item[\NLDOUpdateLabel{t05}{D,M}] The machine from the earliest cost-rate note receives another 0.5 cost-rate increase; infer it from memory and re-optimize the same feasible schedule space.
\item[\NLDOUpdateLabel{t06}{U}] Operations asks for a checkpoint only: keep the current public objective and hard constraints, revalidate the plan, and change it only if it is no longer feasible.
\item[\NLDOUpdateLabel{t07}{U}] No new business event is reported for this stage. Refresh the solution against the current public tables.
\item[\NLDOUpdateLabel{t08}{U}] Treat this as a routine review. The model, objective terms, and public rows stay as they are; preserve feasibility under the current state.
\item[\NLDOUpdateLabel{t09}{U}] The planning team wants the current public contract revalidated without adding new constraints.
\item[\NLDOUpdateLabel{t10}{U}] Run a maintenance update with the same public contract. Revalidate feasibility and quality under the current public state.
\item[\NLDOUpdateLabel{t11}{U}] Operations asks for a checkpoint only: keep the current public objective and hard constraints, revalidate the plan, and change it only if it is no longer feasible.
\item[\NLDOUpdateLabel{t12}{U}] No new business event is reported for this stage. Refresh the solution against the current public tables.
\end{NLDOUpdateList}

\NLDOEpisodeHead{NLDO-P007 (NLDO; inrc2)}
\noindent\textbf{Initial public request.} The instance data is provided in attached tables; treat the listed rows and columns as the source of truth. Create a staff roster over days and shifts. Respect current public coverage, absences, nurse max-shift capacities, and one assignment per staff member per day. Later updates may replace these public rows as well as soft preferences and penalty weights. Report roster disruption only as a diagnostic. Create a one-week nurse roster for 30 nurses over 7 days and day/evening/night shifts. Satisfy fixed coverage and one shift per nurse per day, then optimize workload fairness and soft preferences; report assignment-change counts only as a diagnostic.

\noindent\textbf{Public updates.}
\begin{NLDOUpdateList}
\item[\NLDOUpdateLabel{t01}{O}] Nurse N004 requests D2 off if possible. This is a soft preference only; coverage requirements and availability stay unchanged.
\item[\NLDOUpdateLabel{t02}{O}] Preference satisfaction becomes more important: increase the public preference penalty weight by 10.
\item[\NLDOUpdateLabel{t03}{O}] Nurse N009 requests D4 off if possible under the same hard coverage contract.
\item[\NLDOUpdateLabel{t04}{O,M}] The nurse from the first off-duty preference also requests D5 off if possible; infer the nurse from memory.
\item[\NLDOUpdateLabel{t05}{O}] Activate a temporary soft fairness rule: avoid extra evening shifts for nurses who already worked a night shift.
\item[\NLDOUpdateLabel{t06}{O}] Fairness becomes more important after a staff survey: increase the public fairness penalty weight by 8.
\item[\NLDOUpdateLabel{t07}{O}] Nurse N013 requests D6 off if possible; hard availability and coverage remain unchanged.
\item[\NLDOUpdateLabel{t08}{O}] Weekend preference review: increase sequence-penalty weight by 12 and preference-penalty weight by 5 under the same coverage requirements.
\item[\NLDOUpdateLabel{t09}{O,M}] The nurse from the latest off-duty preference should also avoid D7 night duty if possible; record it as a soft D7 preference from memory.
\item[\NLDOUpdateLabel{t10}{O}] A late policy reset changes only soft rostering priorities: increase fairness weight by 15, reduce preference weight by 5, and keep the evening-after-night sequence rule active. Rebuild the roster under the same fixed coverage and availability constraints.
\item[\NLDOUpdateLabel{t11}{F,O}] A hospital-wide staffing reset starts for the next planning week. Replace the current coverage, active-nurse, absence, preference, nurse max-shift, and policy tables by the public regime-A rows supplied with this update. The active nurse IDs and hard coverage remain valid, but off-duty preferences and their public penalty weight are replaced by the supplied current rows.
\item[\NLDOUpdateLabel{t12}{O}] The following planning week switches independently to public regime B. Replace the current coverage, active-nurse, absence, preference, nurse max-shift, and policy tables by the supplied regime-B rows. The same nurse IDs and hard coverage remain valid, while the off-duty preference rows and their public penalty weight are replaced by the supplied current rows.
\end{NLDOUpdateList}

\NLDOEpisodeHead{NLDO-P008 (NLDO; inrc2)}
\noindent\textbf{Initial public request.} The instance data is provided in attached tables; treat the listed rows and columns as the source of truth. Create a staff roster over days and shifts. Respect current public coverage, absences, nurse max-shift capacities, and one assignment per staff member per day. Later updates may replace these public rows as well as soft preferences and penalty weights. Report roster disruption only as a diagnostic. Create a one-week nurse roster for 34 nurses over 7 days and day/evening/night shifts. Satisfy fixed coverage and one shift per nurse per day, then optimize workload fairness and soft preferences; report assignment-change counts only as a diagnostic.

\noindent\textbf{Public updates.}
\begin{NLDOUpdateList}
\item[\NLDOUpdateLabel{t01}{O}] Nurse N007 requests D2 off if possible. This is a soft preference only; coverage requirements and availability stay unchanged.
\item[\NLDOUpdateLabel{t02}{O}] Preference satisfaction becomes more important: increase the public preference penalty weight by 10.
\item[\NLDOUpdateLabel{t03}{O}] Nurse N012 requests D4 off if possible under the same hard coverage contract.
\item[\NLDOUpdateLabel{t04}{O,M}] The nurse from the first off-duty preference also requests D5 off if possible; infer the nurse from memory.
\item[\NLDOUpdateLabel{t05}{O}] Activate a temporary soft fairness rule: avoid extra evening shifts for nurses who already worked a night shift.
\item[\NLDOUpdateLabel{t06}{O}] Fairness becomes more important after a staff survey: increase the public fairness penalty weight by 8.
\item[\NLDOUpdateLabel{t07}{O}] Nurse N018 requests D6 off if possible; hard availability and coverage remain unchanged.
\item[\NLDOUpdateLabel{t08}{O}] Weekend preference review: increase sequence-penalty weight by 12 and preference-penalty weight by 5 under the same coverage requirements.
\item[\NLDOUpdateLabel{t09}{O,M}] The nurse from the latest off-duty preference should also avoid D7 night duty if possible; record it as a soft D7 preference from memory.
\item[\NLDOUpdateLabel{t10}{O}] Preference pressure softens: reduce the public preference penalty weight by 6 while keeping all hard coverage requirements unchanged.
\item[\NLDOUpdateLabel{t11}{F,O}] A hospital-wide staffing reset starts for the next planning week. Replace the current coverage, active-nurse, absence, preference, nurse max-shift, and policy tables by the public regime-A rows supplied with this update. The active nurse IDs and hard coverage remain valid, but off-duty preferences and their public penalty weight are replaced by the supplied current rows.
\item[\NLDOUpdateLabel{t12}{O}] The following planning week switches independently to public regime B. Replace the current coverage, active-nurse, absence, preference, nurse max-shift, and policy tables by the supplied regime-B rows. The same nurse IDs and hard coverage remain valid, while the off-duty preference rows and their public penalty weight are replaced by the supplied current rows.
\end{NLDOUpdateList}

\NLDOEpisodeHead{NLDO-P009 (NLDO; inrc2)}
\noindent\textbf{Initial public request.} The instance data is provided in attached tables; treat the listed rows and columns as the source of truth. Create a staff roster over days and shifts. Respect current public coverage, absences, nurse max-shift capacities, and one assignment per staff member per day. Later updates may replace these public rows as well as soft preferences and penalty weights. Report roster disruption only as a diagnostic. Create a one-week nurse roster for 38 nurses over 7 days and day/evening/night shifts. Satisfy fixed coverage and one shift per nurse per day, then optimize workload fairness and soft preferences; report assignment-change counts only as a diagnostic.

\noindent\textbf{Public updates.}
\begin{NLDOUpdateList}
\item[\NLDOUpdateLabel{t01}{O}] Nurse N010 requests D2 off if possible. This is a soft preference only; coverage requirements and availability stay unchanged.
\item[\NLDOUpdateLabel{t02}{O}] Preference satisfaction becomes more important: increase the public preference penalty weight by 10.
\item[\NLDOUpdateLabel{t03}{O}] Nurse N015 requests D4 off if possible under the same hard coverage contract.
\item[\NLDOUpdateLabel{t04}{O,M}] The nurse from the first off-duty preference also requests D5 off if possible; infer the nurse from memory.
\item[\NLDOUpdateLabel{t05}{O}] Activate a temporary soft fairness rule: avoid extra evening shifts for nurses who already worked a night shift.
\item[\NLDOUpdateLabel{t06}{O}] Fairness becomes more important after a staff survey: increase the public fairness penalty weight by 8.
\item[\NLDOUpdateLabel{t07}{O}] Nurse N023 requests D6 off if possible; hard availability and coverage remain unchanged.
\item[\NLDOUpdateLabel{t08}{O}] Weekend preference review: increase sequence-penalty weight by 12 and preference-penalty weight by 5 under the same coverage requirements.
\item[\NLDOUpdateLabel{t09}{O,M}] The nurse from the latest off-duty preference should also avoid D7 night duty if possible; record it as a soft D7 preference from memory.
\item[\NLDOUpdateLabel{t10}{O}] Preference pressure softens: reduce the public preference penalty weight by 6 while keeping all hard coverage requirements unchanged.
\item[\NLDOUpdateLabel{t11}{F,O}] A hospital-wide staffing reset starts for the next planning week. Replace the current coverage, active-nurse, absence, preference, nurse max-shift, and policy tables by the public regime-A rows supplied with this update. The active nurse IDs and hard coverage remain valid, but off-duty preferences and their public penalty weight are replaced by the supplied current rows.
\item[\NLDOUpdateLabel{t12}{O}] The following planning week switches independently to public regime B. Replace the current coverage, active-nurse, absence, preference, nurse max-shift, and policy tables by the supplied regime-B rows. The same nurse IDs and hard coverage remain valid, while the off-duty preference rows and their public penalty weight are replaced by the supplied current rows.
\end{NLDOUpdateList}

\NLDOEpisodeHead{NLDO-P010 (NLDO; green\_vrp\_multiobjective)}
\noindent\textbf{Initial public request.} Plan same-day delivery routes from a single depot for a small urban fleet. Initial depot, vehicle, and order data are provided in public CSV tables. Use the attached depot, vehicle, order, and policy tables as the source of truth. All active orders should be served once under vehicle capacity. Rows with active=false are public candidate orders that are not served until a later natural-language update explicitly activates them. Later updates may replace public order geometry, demand, time windows, priorities, vehicle emissions, depot coordinates, or carbon policy; all current rows remain public. Optimize three main criteria: travel distance, delivery lateness, and emissions using public policy parameters when present. Also report route disruption and priority-customer penalties as auxiliary diagnostics.

\noindent\textbf{Public updates.}
\begin{NLDOUpdateList}
\item[\NLDOUpdateLabel{t01}{O}] Existing order O006 becomes a same-day priority delivery: increase its public priority by 2 and tighten its soft due target by 18 minutes. The order set, vehicle set, capacities, and active flags stay unchanged.
\item[\NLDOUpdateLabel{t02}{F,O}] Traffic around order O004 slows vehicles by 25 percent; remember this customer as the congestion anchor.
\item[\NLDOUpdateLabel{t03}{O}] Vehicle V1's emission\_rate increases by 0.18 after a telemetry update; it remains available with the same capacity.
\item[\NLDOUpdateLabel{t04}{R}] Carbon policy is tightened: set the public policy carbon\_multiplier to 1.35 so emissions count 35 percent more heavily in dispatch evaluation.
\item[\NLDOUpdateLabel{t05}{O,M}] The customer mentioned in the congestion note now needs a tighter delivery window ending 20 minutes earlier; infer the order from memory.
\item[\NLDOUpdateLabel{t06}{O}] An express-service wave temporarily reverses dispatch priorities. Set policy.carbon\_multiplier to 0.75 and tighten the soft due-time targets for O006, O009, and O012 by 45, 50, and 35 minutes respectively. Increase each of those existing orders' public priority by 1. These due-time targets affect lateness and priority-service quality rather than hard feasibility, so route quality should be reconsidered instead of blindly preserving the previous carbon-oriented plan.
\item[\NLDOUpdateLabel{t07}{O}] Existing pharmacy-like order O010 gets a tighter soft due target by 22 minutes and priority increases by 2. The order table remains fixed.
\item[\NLDOUpdateLabel{t08}{O}] A carbon-rationing notice takes effect after the express wave. Set policy.carbon\_multiplier to 4.80, increase every vehicle emission\_rate by 0.35, and reduce priorities for O006, O009, and O012 by 1. This is a regime change toward carbon-efficient routing over the same active orders and vehicles, so routes specialized for the previous express wave can be misleading.
\item[\NLDOUpdateLabel{t09}{O,M}] The vehicle from the telemetry update receives a 0.10 emission\_rate correction downward; use the vehicle identity from memory.
\item[\NLDOUpdateLabel{t10}{O}] A late emergency service wave changes priorities over the same order set. Set policy.carbon\_multiplier to 0.45, subtract 0.20 from every vehicle emission\_rate, and subtract another 0.08 from cleaner vehicle V2. Tighten the soft due-time targets for O001, O002, and O003 by 40, 45, and 50 minutes, and increase their priorities by 2. These due-time targets affect lateness and priority-service quality, not hard feasibility. Use the current public tables as the source of truth.
\item[\NLDOUpdateLabel{t11}{F,O}] A network-wide dispatch reset activates public regime A. Replace the active workload by the prelisted cohort A, replace the depot coordinates and every active order's x, y, demand, ready, due, service, and priority fields by the supplied current rows; also replace every vehicle capacity, shift\_end, availability, emission\_rate, and policy.carbon\_multiplier by the supplied values. Serve every public order row in the active cohort exactly once. The route encoding type and the three objective definitions remain unchanged.
\item[\NLDOUpdateLabel{t12}{F,O}] A network-wide dispatch reset activates public regime B. Replace the active workload by the prelisted cohort B, replace the depot coordinates and every active order's x, y, demand, ready, due, service, and priority fields by the supplied current rows; also replace every vehicle capacity, shift\_end, availability, emission\_rate, and policy.carbon\_multiplier by the supplied values. Serve every public order row in the active cohort exactly once. The route encoding type and the three objective definitions remain unchanged.
\end{NLDOUpdateList}

\NLDOEpisodeHead{NLDO-P011 (NLDO; green\_vrp\_multiobjective)}
\noindent\textbf{Initial public request.} Plan same-day delivery routes from a single depot for a small urban fleet. Initial depot, vehicle, and order data are provided in public CSV tables. Use the attached depot, vehicle, order, and policy tables as the source of truth. All active orders should be served once under vehicle capacity. Rows with active=false are public candidate orders that are not served until a later natural-language update explicitly activates them. Later updates may replace public order geometry, demand, time windows, priorities, vehicle emissions, depot coordinates, or carbon policy; all current rows remain public. Optimize three main criteria: travel distance, delivery lateness, and emissions using public policy parameters when present. Also report route disruption and priority-customer penalties as auxiliary diagnostics.

\noindent\textbf{Public updates.}
\begin{NLDOUpdateList}
\item[\NLDOUpdateLabel{t01}{O}] Existing order O007 becomes a same-day priority delivery: increase its public priority by 2 and tighten its soft due target by 18 minutes. The order set, vehicle set, capacities, and active flags stay unchanged.
\item[\NLDOUpdateLabel{t02}{F,O}] Traffic around order O005 slows vehicles by 25 percent; remember this customer as the congestion anchor.
\item[\NLDOUpdateLabel{t03}{O}] Vehicle V2's emission\_rate increases by 0.18 after a telemetry update; it remains available with the same capacity.
\item[\NLDOUpdateLabel{t04}{R}] Carbon policy is tightened: set the public policy carbon\_multiplier to 1.35 so emissions count 35 percent more heavily in dispatch evaluation.
\item[\NLDOUpdateLabel{t05}{O,M}] The customer mentioned in the congestion note now needs a tighter delivery window ending 20 minutes earlier; infer the order from memory.
\item[\NLDOUpdateLabel{t06}{O}] An express-service wave temporarily reverses dispatch priorities. Set policy.carbon\_multiplier to 0.75 and tighten the soft due-time targets for O007, O010, and O013 by 45, 50, and 35 minutes respectively. Increase each of those existing orders' public priority by 1. These due-time targets affect lateness and priority-service quality rather than hard feasibility, so route quality should be reconsidered instead of blindly preserving the previous carbon-oriented plan.
\item[\NLDOUpdateLabel{t07}{O}] Existing pharmacy-like order O011 gets a tighter soft due target by 22 minutes and priority increases by 2. The order table remains fixed.
\item[\NLDOUpdateLabel{t08}{O}] A carbon-rationing notice takes effect after the express wave. Set policy.carbon\_multiplier to 4.80, increase every vehicle emission\_rate by 0.35, and reduce priorities for O007, O010, and O013 by 1. This is a regime change toward carbon-efficient routing over the same active orders and vehicles, so routes specialized for the previous express wave can be misleading.
\item[\NLDOUpdateLabel{t09}{O,M}] The vehicle from the telemetry update receives a 0.10 emission\_rate correction downward; use the vehicle identity from memory.
\item[\NLDOUpdateLabel{t10}{O}] A late emergency service wave changes priorities over the same order set. Set policy.carbon\_multiplier to 0.45, subtract 0.20 from every vehicle emission\_rate, and subtract another 0.08 from cleaner vehicle V3. Tighten the soft due-time targets for O002, O003, and O004 by 40, 45, and 50 minutes, and increase their priorities by 2. These due-time targets affect lateness and priority-service quality, not hard feasibility. Use the current public tables as the source of truth.
\item[\NLDOUpdateLabel{t11}{F,O}] A network-wide dispatch reset activates public regime A. Replace the active workload by the prelisted cohort A, replace the depot coordinates and every active order's x, y, demand, ready, due, service, and priority fields by the supplied current rows; also replace every vehicle capacity, shift\_end, availability, emission\_rate, and policy.carbon\_multiplier by the supplied values. Serve every public order row in the active cohort exactly once. The route encoding type and the three objective definitions remain unchanged.
\item[\NLDOUpdateLabel{t12}{F,O}] A network-wide dispatch reset activates public regime B. Replace the active workload by the prelisted cohort B, replace the depot coordinates and every active order's x, y, demand, ready, due, service, and priority fields by the supplied current rows; also replace every vehicle capacity, shift\_end, availability, emission\_rate, and policy.carbon\_multiplier by the supplied values. Serve every public order row in the active cohort exactly once. The route encoding type and the three objective definitions remain unchanged.
\end{NLDOUpdateList}

\NLDOEpisodeHead{NLDO-P012 (NLDO; green\_vrp\_multiobjective)}
\noindent\textbf{Initial public request.} Plan same-day delivery routes from a single depot for a small urban fleet. Initial depot, vehicle, and order data are provided in public CSV tables. Use the attached depot, vehicle, order, and policy tables as the source of truth. All active orders should be served once under vehicle capacity. Rows with active=false are public candidate orders that are not served until a later natural-language update explicitly activates them. Later updates may replace public order geometry, demand, time windows, priorities, vehicle emissions, depot coordinates, or carbon policy; all current rows remain public. Optimize three main criteria: travel distance, delivery lateness, and emissions using public policy parameters when present. Also report route disruption and priority-customer penalties as auxiliary diagnostics.

\noindent\textbf{Public updates.}
\begin{NLDOUpdateList}
\item[\NLDOUpdateLabel{t01}{O}] Existing order O008 becomes a same-day priority delivery: increase its public priority by 2 and tighten its soft due target by 18 minutes. The order set, vehicle set, capacities, and active flags stay unchanged.
\item[\NLDOUpdateLabel{t02}{F,O}] Traffic around order O006 slows vehicles by 25 percent; remember this customer as the congestion anchor.
\item[\NLDOUpdateLabel{t03}{O}] Vehicle V3's emission\_rate increases by 0.18 after a telemetry update; it remains available with the same capacity.
\item[\NLDOUpdateLabel{t04}{R}] Carbon policy is tightened: set the public policy carbon\_multiplier to 1.35 so emissions count 35 percent more heavily in dispatch evaluation.
\item[\NLDOUpdateLabel{t05}{O,M}] The customer mentioned in the congestion note now needs a tighter delivery window ending 20 minutes earlier; infer the order from memory.
\item[\NLDOUpdateLabel{t06}{O}] An express-service wave temporarily reverses dispatch priorities. Set policy.carbon\_multiplier to 0.75 and tighten the soft due-time targets for O008, O011, and O014 by 45, 50, and 35 minutes respectively. Increase each of those existing orders' public priority by 1. These due-time targets affect lateness and priority-service quality rather than hard feasibility, so route quality should be reconsidered instead of blindly preserving the previous carbon-oriented plan.
\item[\NLDOUpdateLabel{t07}{O}] Existing pharmacy-like order O012 gets a tighter soft due target by 22 minutes and priority increases by 2. The order table remains fixed.
\item[\NLDOUpdateLabel{t08}{O}] A carbon-rationing notice takes effect after the express wave. Set policy.carbon\_multiplier to 4.80, increase every vehicle emission\_rate by 0.35, and reduce priorities for O008, O011, and O014 by 1. This is a regime change toward carbon-efficient routing over the same active orders and vehicles, so routes specialized for the previous express wave can be misleading.
\item[\NLDOUpdateLabel{t09}{O,M}] The vehicle from the telemetry update receives a 0.10 emission\_rate correction downward; use the vehicle identity from memory.
\item[\NLDOUpdateLabel{t10}{O}] A late emergency service wave changes priorities over the same order set. Set policy.carbon\_multiplier to 0.45, subtract 0.20 from every vehicle emission\_rate, and subtract another 0.08 from cleaner vehicle V2. Tighten the soft due-time targets for O003, O004, and O005 by 40, 45, and 50 minutes, and increase their priorities by 2. These due-time targets affect lateness and priority-service quality, not hard feasibility. Use the current public tables as the source of truth.
\item[\NLDOUpdateLabel{t11}{F,O}] A network-wide dispatch reset activates public regime A. Replace the active workload by the prelisted cohort A, replace the depot coordinates and every active order's x, y, demand, ready, due, service, and priority fields by the supplied current rows; also replace every vehicle capacity, shift\_end, availability, emission\_rate, and policy.carbon\_multiplier by the supplied values. Serve every public order row in the active cohort exactly once. The route encoding type and the three objective definitions remain unchanged.
\item[\NLDOUpdateLabel{t12}{F,O}] A network-wide dispatch reset activates public regime B. Replace the active workload by the prelisted cohort B, replace the depot coordinates and every active order's x, y, demand, ready, due, service, and priority fields by the supplied current rows; also replace every vehicle capacity, shift\_end, availability, emission\_rate, and policy.carbon\_multiplier by the supplied values. Serve every public order row in the active cohort exactly once. The route encoding type and the three objective definitions remain unchanged.
\end{NLDOUpdateList}

\NLDOEpisodeHead{NLDO-P013 (NLDO; cloud\_scheduling\_multiobjective)}
\noindent\textbf{Initial public request.} The instance data is provided in attached tables; treat the listed rows and columns as the source of truth. Assign active tasks to available compute resources. Rows with active=false are public candidate jobs that are not scheduled until a later update explicitly activates them. Respect current public CPU, memory, accelerator needs, and machine availability. Optimize energy and imbalance as main objectives and report SLA/latency/migration diagnostics without treating those diagnostics as hard feasibility constraints. Later updates may replace public workload requirements, machine capacity/accelerator profiles, availability, soft SLA, priority, energy, or carbon parameters. Place active cloud jobs onto available machines for the next scheduling window. Machines: M01 cpu 38 mem 80 gpu false, M02 cpu 42 mem 88 gpu true, M03 cpu 46 mem 96 gpu false, M04 cpu 34 mem 72 gpu false, M05 cpu 38 mem 80 gpu true, M06 cpu 42 mem 88 gpu false, M07 cpu 46 mem 96 gpu false, M08 cpu 34 mem 72 gpu true. Jobs: J001 cpu 7 mem 11 deadline 81 priority 2, J002 cpu 3 mem 18 deadline 92 priority 3, J003 cpu 8 mem 7 deadline 103 priority 4, J004 cpu 4 mem 14 deadline 114 priority 1, J005 cpu 9 mem 21 deadline 125 priority 2, J006 cpu 5 mem 10 deadline 136 priority 3, J007 cpu 10 mem 17 deadline 147 priority 4, J008 cpu 6 mem 6 deadline 158 priority 1, J009 cpu 2 mem 13 deadline 169 priority 2, J010 cpu 7 mem 20 deadline 180 priority 3, J011 cpu 3 mem 9 deadline 71 priority 4, J012 cpu 8 mem 16 deadline 82 priority 1, J013 cpu 4 mem 5 deadline 93 priority 2, J014 cpu 9 mem 12 deadline 104 priority 3, J015 cpu 5 mem 19 deadline 115 priority 4, J016 cpu 10 mem 8 deadline 126 priority 1, J017 cpu 6 mem 15 deadline 137 priority 2, J018 cpu 2 mem 4 deadline 148 priority 3, J019 cpu 7 mem 11 deadline 159 priority 4, J020 cpu 3 mem 18 deadline 170 priority 1, J021 cpu 8 mem 7 deadline 181 priority 2, J022 cpu 4 mem 14 deadline 72 priority 3, J023 cpu 9 mem 21 deadline 83 priority 4, J024 cpu 5 mem 10 deadline 94 priority 1, B20\_01 cpu 3 mem 8 deadline 69 priority 4, B20\_02 cpu 4 mem 11 deadline 73 priority 4, B20\_03 cpu 5 mem 14 deadline 77 priority 4, B20\_04 cpu 2 mem 7 deadline 81 priority 4, B20\_05 cpu 3 mem 10 deadline 85 priority 4, B20\_06 cpu 4 mem 13 deadline 89 priority 4, L001 cpu 8 mem 16 deadline 91 priority 2, L002 cpu 13 mem 25 deadline 102 priority 3, L003 cpu 6 mem 34 deadline 113 priority 4, L004 cpu 11 mem 15 deadline 124 priority 1, L005 cpu 4 mem 24 deadline 135 priority 2, L006 cpu 9 mem 33 deadline 146 priority 3, L007 cpu 14 mem 14 deadline 157 priority 4, L008 cpu 7 mem 23 deadline 168 priority 1, L009 cpu 12 mem 32 deadline 179 priority 2, L010 cpu 5 mem 13 deadline 190 priority 3, L011 cpu 10 mem 22 deadline 201 priority 4, L012 cpu 3 mem 31 deadline 212 priority 1, L013 cpu 8 mem 12 deadline 223 priority 2, L014 cpu 13 mem 21 deadline 234 priority 3, L015 cpu 6 mem 30 deadline 85 priority 4, L016 cpu 11 mem 11 deadline 96 priority 1, L017 cpu 4 mem 20 deadline 107 priority 2, L018 cpu 9 mem 29 deadline 118 priority 3, L019 cpu 14 mem 10 deadline 129 priority 4, L020 cpu 7 mem 19 deadline 140 priority 1, L021 cpu 12 mem 28 deadline 151 priority 2, L022 cpu 5 mem 9 deadline 162 priority 3, L023 cpu 10 mem 18 deadline 173 priority 4, L024 cpu 3 mem 27 deadline 184 priority 1, L025 cpu 8 mem 8 deadline 195 priority 2, L026 cpu 13 mem 17 deadline 206 priority 3, L027 cpu 6 mem 26 deadline 217 priority 4, L028 cpu 11 mem 7 deadline 228 priority 1, L029 cpu 4 mem 16 deadline 239 priority 2, L030 cpu 9 mem 25 deadline 90 priority 3, L031 cpu 14 mem 34 deadline 101 priority 4, L032 cpu 7 mem 15 deadline 112 priority 1, L033 cpu 12 mem 24 deadline 123 priority 2, L034 cpu 5 mem 33 deadline 134 priority 3, L035 cpu 10 mem 14 deadline 145 priority 4, L036 cpu 3 mem 23 deadline 156 priority 1, L037 cpu 8 mem 32 deadline 167 priority 2, L038 cpu 13 mem 13 deadline 178 priority 3, L039 cpu 6 mem 22 deadline 189 priority 4, L040 cpu 11 mem 31 deadline 200 priority 1. Jobs with active=false are public candidate jobs that are not scheduled until a later natural-language update explicitly activates them. Respect CPU, memory, GPU needs, and machine availability. Optimize two main criteria: energy use and load imbalance. Also report SLA violations, latency, and migrations as diagnostics; these diagnostics do not define feasibility.

\noindent\textbf{Public updates.}
\begin{NLDOUpdateList}
\item[\NLDOUpdateLabel{t01}{D}] A latency-sensitive workload group burst-0 is identified among existing jobs J001, J002, J003, J004, J005. Increase their public priority by 1 and remember this group name for later updates. The job table remains fixed.
\item[\NLDOUpdateLabel{t02}{O}] Machine M02's energy\_per\_cpu increases by 0.055 because of thermal throttling, but its CPU and memory capacity stay unchanged.
\item[\NLDOUpdateLabel{t03}{D}] Soft SLA target for job J005 is tightened by 35 minutes after an escalation. This affects latency penalty, not hard feasibility.
\item[\NLDOUpdateLabel{t04}{R}] Energy price doubles for the next control window, so energy cost becomes much more important.
\item[\NLDOUpdateLabel{t05}{D,M}] The workload burst mentioned at the beginning now gets premium priority 5; infer the affected jobs from memory.
\item[\NLDOUpdateLabel{t06}{O}] A carbon-aware operating window starts: machine M08's energy\_idle increases by 1.6, carbon\_intensity rises to 1.80, and energy\_price is set to 1.25. This changes the energy-versus-balance trade-off without changing the machine set or capacities.
\item[\NLDOUpdateLabel{t07}{D}] Existing analytics-like job J018 becomes less urgent: reduce its priority by 1 and extend its soft SLA target by 25 minutes. The active job set remains fixed.
\item[\NLDOUpdateLabel{t08}{R}] A balancing directive follows the scarcity window: energy\_price drops to 0.70 while carbon\_intensity remains elevated at 1.35. The plan should now expose better load balance instead of simply minimizing energy.
\item[\NLDOUpdateLabel{t09}{R}] Operations relaxes the analytics job J018 by extending its soft SLA target by 30 minutes, and carbon\_intensity is set to 1.10.
\item[\NLDOUpdateLabel{t10}{O}] A late reset changes energy and service priorities over the same jobs. Machine M02's energy\_per\_cpu is reduced by 0.04; machine M05's energy\_idle increases by 2.2; energy\_price is set to 0.60 and carbon\_intensity to 0.85. Existing jobs J010, J011, J012, J013, J014 receive priority +1 and soft SLA targets tightened by 20 minutes. The objective trade-off has moved, but current feasibility is still judged over the same fixed public resource and job set.
\item[\NLDOUpdateLabel{t11}{F,O}] A cluster-wide hardware and workload reset activates public regime A. Activate the prelisted late-window jobs, replace every machine's CPU, memory, GPU, availability, idle-energy, and per-CPU energy fields and replace every active job's CPU, memory, GPU requirement, deadline, latency sensitivity, and priority by the supplied rows. Job and machine IDs remain unchanged, and the placement encoding and two objective definitions remain unchanged. Also replace energy\_price and carbon\_intensity by the supplied current values.
\item[\NLDOUpdateLabel{t12}{F,O}] A cluster-wide hardware and workload reset activates public regime B. Activate the prelisted late-window jobs, replace every machine's CPU, memory, GPU, availability, idle-energy, and per-CPU energy fields and replace every active job's CPU, memory, GPU requirement, deadline, latency sensitivity, and priority by the supplied rows. Job and machine IDs remain unchanged, and the placement encoding and two objective definitions remain unchanged. Also replace energy\_price and carbon\_intensity by the supplied current values.
\end{NLDOUpdateList}

\NLDOEpisodeHead{NLDO-P014 (NLDO; cloud\_scheduling\_multiobjective)}
\noindent\textbf{Initial public request.} The instance data is provided in attached tables; treat the listed rows and columns as the source of truth. Assign active tasks to available compute resources. Rows with active=false are public candidate jobs that are not scheduled until a later update explicitly activates them. Respect current public CPU, memory, accelerator needs, and machine availability. Optimize energy and imbalance as main objectives and report SLA/latency/migration diagnostics without treating those diagnostics as hard feasibility constraints. Later updates may replace public workload requirements, machine capacity/accelerator profiles, availability, soft SLA, priority, energy, or carbon parameters. Place active cloud jobs onto available machines for the next scheduling window. Machines: M01 cpu 42 mem 96 gpu false, M02 cpu 46 mem 72 gpu true, M03 cpu 34 mem 80 gpu false, M04 cpu 38 mem 88 gpu false, M05 cpu 42 mem 96 gpu true, M06 cpu 46 mem 72 gpu false, M07 cpu 34 mem 80 gpu false, M08 cpu 38 mem 88 gpu true. Jobs: J001 cpu 8 mem 13 deadline 84 priority 3, J002 cpu 4 mem 20 deadline 95 priority 4, J003 cpu 9 mem 9 deadline 106 priority 1, J004 cpu 5 mem 16 deadline 117 priority 2, J005 cpu 10 mem 5 deadline 128 priority 3, J006 cpu 6 mem 12 deadline 139 priority 4, J007 cpu 2 mem 19 deadline 150 priority 1, J008 cpu 7 mem 8 deadline 161 priority 2, J009 cpu 3 mem 15 deadline 172 priority 3, J010 cpu 8 mem 4 deadline 183 priority 4, J011 cpu 4 mem 11 deadline 74 priority 1, J012 cpu 9 mem 18 deadline 85 priority 2, J013 cpu 5 mem 7 deadline 96 priority 3, J014 cpu 10 mem 14 deadline 107 priority 4, J015 cpu 6 mem 21 deadline 118 priority 1, J016 cpu 2 mem 10 deadline 129 priority 2, J017 cpu 7 mem 17 deadline 140 priority 3, J018 cpu 3 mem 6 deadline 151 priority 4, J019 cpu 8 mem 13 deadline 162 priority 1, J020 cpu 4 mem 20 deadline 173 priority 2, J021 cpu 9 mem 9 deadline 184 priority 3, J022 cpu 5 mem 16 deadline 75 priority 4, J023 cpu 10 mem 5 deadline 86 priority 1, J024 cpu 6 mem 12 deadline 97 priority 2, B21\_01 cpu 4 mem 10 deadline 70 priority 4, B21\_02 cpu 5 mem 13 deadline 74 priority 4, B21\_03 cpu 2 mem 6 deadline 78 priority 4, B21\_04 cpu 3 mem 9 deadline 82 priority 4, B21\_05 cpu 4 mem 12 deadline 86 priority 4, B21\_06 cpu 5 mem 5 deadline 90 priority 4, L001 cpu 9 mem 19 deadline 92 priority 3, L002 cpu 14 mem 28 deadline 103 priority 4, L003 cpu 7 mem 9 deadline 114 priority 1, L004 cpu 12 mem 18 deadline 125 priority 2, L005 cpu 5 mem 27 deadline 136 priority 3, L006 cpu 10 mem 8 deadline 147 priority 4, L007 cpu 3 mem 17 deadline 158 priority 1, L008 cpu 8 mem 26 deadline 169 priority 2, L009 cpu 13 mem 7 deadline 180 priority 3, L010 cpu 6 mem 16 deadline 191 priority 4, L011 cpu 11 mem 25 deadline 202 priority 1, L012 cpu 4 mem 34 deadline 213 priority 2, L013 cpu 9 mem 15 deadline 224 priority 3, L014 cpu 14 mem 24 deadline 235 priority 4, L015 cpu 7 mem 33 deadline 86 priority 1, L016 cpu 12 mem 14 deadline 97 priority 2, L017 cpu 5 mem 23 deadline 108 priority 3, L018 cpu 10 mem 32 deadline 119 priority 4, L019 cpu 3 mem 13 deadline 130 priority 1, L020 cpu 8 mem 22 deadline 141 priority 2, L021 cpu 13 mem 31 deadline 152 priority 3, L022 cpu 6 mem 12 deadline 163 priority 4, L023 cpu 11 mem 21 deadline 174 priority 1, L024 cpu 4 mem 30 deadline 185 priority 2, L025 cpu 9 mem 11 deadline 196 priority 3, L026 cpu 14 mem 20 deadline 207 priority 4, L027 cpu 7 mem 29 deadline 218 priority 1, L028 cpu 12 mem 10 deadline 229 priority 2, L029 cpu 5 mem 19 deadline 80 priority 3, L030 cpu 10 mem 28 deadline 91 priority 4, L031 cpu 3 mem 9 deadline 102 priority 1, L032 cpu 8 mem 18 deadline 113 priority 2, L033 cpu 13 mem 27 deadline 124 priority 3, L034 cpu 6 mem 8 deadline 135 priority 4, L035 cpu 11 mem 17 deadline 146 priority 1, L036 cpu 4 mem 26 deadline 157 priority 2, L037 cpu 9 mem 7 deadline 168 priority 3, L038 cpu 14 mem 16 deadline 179 priority 4, L039 cpu 7 mem 25 deadline 190 priority 1, L040 cpu 12 mem 34 deadline 201 priority 2. Jobs with active=false are public candidate jobs that are not scheduled until a later natural-language update explicitly activates them. Respect CPU, memory, GPU needs, and machine availability. Optimize two main criteria: energy use and load imbalance. Also report SLA violations, latency, and migrations as diagnostics; these diagnostics do not define feasibility.

\noindent\textbf{Public updates.}
\begin{NLDOUpdateList}
\item[\NLDOUpdateLabel{t01}{D}] A latency-sensitive workload group burst-1 is identified among existing jobs J002, J003, J004, J005, J006. Increase their public priority by 1 and remember this group name for later updates. The job table remains fixed.
\item[\NLDOUpdateLabel{t02}{O}] Machine M03's energy\_per\_cpu increases by 0.055 because of thermal throttling, but its CPU and memory capacity stay unchanged.
\item[\NLDOUpdateLabel{t03}{D}] Soft SLA target for job J006 is tightened by 35 minutes after an escalation. This affects latency penalty, not hard feasibility.
\item[\NLDOUpdateLabel{t04}{R}] Energy price doubles for the next control window, so energy cost becomes much more important.
\item[\NLDOUpdateLabel{t05}{D,M}] The workload burst mentioned at the beginning now gets premium priority 5; infer the affected jobs from memory.
\item[\NLDOUpdateLabel{t06}{O}] A carbon-aware operating window starts: machine M07's energy\_idle increases by 1.6, carbon\_intensity rises to 1.80, and energy\_price is set to 1.25. This changes the energy-versus-balance trade-off without changing the machine set or capacities.
\item[\NLDOUpdateLabel{t07}{D}] Existing analytics-like job J019 becomes less urgent: reduce its priority by 1 and extend its soft SLA target by 25 minutes. The active job set remains fixed.
\item[\NLDOUpdateLabel{t08}{R}] A balancing directive follows the scarcity window: energy\_price drops to 0.70 while carbon\_intensity remains elevated at 1.35. The plan should now expose better load balance instead of simply minimizing energy.
\item[\NLDOUpdateLabel{t09}{R}] Operations relaxes the analytics job J019 by extending its soft SLA target by 30 minutes, and carbon\_intensity is set to 1.10.
\item[\NLDOUpdateLabel{t10}{O}] A late reset changes energy and service priorities over the same jobs. Machine M03's energy\_per\_cpu is reduced by 0.04; machine M06's energy\_idle increases by 2.2; energy\_price is set to 0.60 and carbon\_intensity to 0.85. Existing jobs J011, J012, J013, J014, J015 receive priority +1 and soft SLA targets tightened by 20 minutes. The objective trade-off has moved, but current feasibility is still judged over the same fixed public resource and job set.
\item[\NLDOUpdateLabel{t11}{F,O}] A cluster-wide hardware and workload reset activates public regime A. Activate the prelisted late-window jobs, replace every machine's CPU, memory, GPU, availability, idle-energy, and per-CPU energy fields and replace every active job's CPU, memory, GPU requirement, deadline, latency sensitivity, and priority by the supplied rows. Job and machine IDs remain unchanged, and the placement encoding and two objective definitions remain unchanged. Also replace energy\_price and carbon\_intensity by the supplied current values.
\item[\NLDOUpdateLabel{t12}{F,O}] A cluster-wide hardware and workload reset activates public regime B. Activate the prelisted late-window jobs, replace every machine's CPU, memory, GPU, availability, idle-energy, and per-CPU energy fields and replace every active job's CPU, memory, GPU requirement, deadline, latency sensitivity, and priority by the supplied rows. Job and machine IDs remain unchanged, and the placement encoding and two objective definitions remain unchanged. Also replace energy\_price and carbon\_intensity by the supplied current values.
\end{NLDOUpdateList}

\NLDOEpisodeHead{NLDO-P015 (NLDO; cloud\_scheduling\_multiobjective)}
\noindent\textbf{Initial public request.} The instance data is provided in attached tables; treat the listed rows and columns as the source of truth. Assign active tasks to available compute resources. Rows with active=false are public candidate jobs that are not scheduled until a later update explicitly activates them. Respect current public CPU, memory, accelerator needs, and machine availability. Optimize energy and imbalance as main objectives and report SLA/latency/migration diagnostics without treating those diagnostics as hard feasibility constraints. Later updates may replace public workload requirements, machine capacity/accelerator profiles, availability, soft SLA, priority, energy, or carbon parameters. Place active cloud jobs onto available machines for the next scheduling window. Machines: M01 cpu 46 mem 80 gpu false, M02 cpu 34 mem 88 gpu true, M03 cpu 38 mem 96 gpu false, M04 cpu 42 mem 72 gpu false, M05 cpu 46 mem 80 gpu true, M06 cpu 34 mem 88 gpu false, M07 cpu 38 mem 96 gpu false, M08 cpu 42 mem 72 gpu true. Jobs: J001 cpu 9 mem 15 deadline 87 priority 4, J002 cpu 5 mem 4 deadline 98 priority 1, J003 cpu 10 mem 11 deadline 109 priority 2, J004 cpu 6 mem 18 deadline 120 priority 3, J005 cpu 2 mem 7 deadline 131 priority 4, J006 cpu 7 mem 14 deadline 142 priority 1, J007 cpu 3 mem 21 deadline 153 priority 2, J008 cpu 8 mem 10 deadline 164 priority 3, J009 cpu 4 mem 17 deadline 175 priority 4, J010 cpu 9 mem 6 deadline 186 priority 1, J011 cpu 5 mem 13 deadline 77 priority 2, J012 cpu 10 mem 20 deadline 88 priority 3, J013 cpu 6 mem 9 deadline 99 priority 4, J014 cpu 2 mem 16 deadline 110 priority 1, J015 cpu 7 mem 5 deadline 121 priority 2, J016 cpu 3 mem 12 deadline 132 priority 3, J017 cpu 8 mem 19 deadline 143 priority 4, J018 cpu 4 mem 8 deadline 154 priority 1, J019 cpu 9 mem 15 deadline 165 priority 2, J020 cpu 5 mem 4 deadline 176 priority 3, J021 cpu 10 mem 11 deadline 187 priority 4, J022 cpu 6 mem 18 deadline 78 priority 1, J023 cpu 2 mem 7 deadline 89 priority 2, J024 cpu 7 mem 14 deadline 100 priority 3, B22\_01 cpu 5 mem 12 deadline 71 priority 4, B22\_02 cpu 2 mem 5 deadline 75 priority 4, B22\_03 cpu 3 mem 8 deadline 79 priority 4, B22\_04 cpu 4 mem 11 deadline 83 priority 4, B22\_05 cpu 5 mem 14 deadline 87 priority 4, B22\_06 cpu 2 mem 7 deadline 91 priority 4, L001 cpu 10 mem 22 deadline 93 priority 4, L002 cpu 3 mem 31 deadline 104 priority 1, L003 cpu 8 mem 12 deadline 115 priority 2, L004 cpu 13 mem 21 deadline 126 priority 3, L005 cpu 6 mem 30 deadline 137 priority 4, L006 cpu 11 mem 11 deadline 148 priority 1, L007 cpu 4 mem 20 deadline 159 priority 2, L008 cpu 9 mem 29 deadline 170 priority 3, L009 cpu 14 mem 10 deadline 181 priority 4, L010 cpu 7 mem 19 deadline 192 priority 1, L011 cpu 12 mem 28 deadline 203 priority 2, L012 cpu 5 mem 9 deadline 214 priority 3, L013 cpu 10 mem 18 deadline 225 priority 4, L014 cpu 3 mem 27 deadline 236 priority 1, L015 cpu 8 mem 8 deadline 87 priority 2, L016 cpu 13 mem 17 deadline 98 priority 3, L017 cpu 6 mem 26 deadline 109 priority 4, L018 cpu 11 mem 7 deadline 120 priority 1, L019 cpu 4 mem 16 deadline 131 priority 2, L020 cpu 9 mem 25 deadline 142 priority 3, L021 cpu 14 mem 34 deadline 153 priority 4, L022 cpu 7 mem 15 deadline 164 priority 1, L023 cpu 12 mem 24 deadline 175 priority 2, L024 cpu 5 mem 33 deadline 186 priority 3, L025 cpu 10 mem 14 deadline 197 priority 4, L026 cpu 3 mem 23 deadline 208 priority 1, L027 cpu 8 mem 32 deadline 219 priority 2, L028 cpu 13 mem 13 deadline 230 priority 3, L029 cpu 6 mem 22 deadline 81 priority 4, L030 cpu 11 mem 31 deadline 92 priority 1, L031 cpu 4 mem 12 deadline 103 priority 2, L032 cpu 9 mem 21 deadline 114 priority 3, L033 cpu 14 mem 30 deadline 125 priority 4, L034 cpu 7 mem 11 deadline 136 priority 1, L035 cpu 12 mem 20 deadline 147 priority 2, L036 cpu 5 mem 29 deadline 158 priority 3, L037 cpu 10 mem 10 deadline 169 priority 4, L038 cpu 3 mem 19 deadline 180 priority 1, L039 cpu 8 mem 28 deadline 191 priority 2, L040 cpu 13 mem 9 deadline 202 priority 3. Jobs with active=false are public candidate jobs that are not scheduled until a later natural-language update explicitly activates them. Respect CPU, memory, GPU needs, and machine availability. Optimize two main criteria: energy use and load imbalance. Also report SLA violations, latency, and migrations as diagnostics; these diagnostics do not define feasibility.

\noindent\textbf{Public updates.}
\begin{NLDOUpdateList}
\item[\NLDOUpdateLabel{t01}{D}] A latency-sensitive workload group burst-2 is identified among existing jobs J003, J004, J005, J006, J007. Increase their public priority by 1 and remember this group name for later updates. The job table remains fixed.
\item[\NLDOUpdateLabel{t02}{O}] Machine M04's energy\_per\_cpu increases by 0.055 because of thermal throttling, but its CPU and memory capacity stay unchanged.
\item[\NLDOUpdateLabel{t03}{D}] Soft SLA target for job J007 is tightened by 35 minutes after an escalation. This affects latency penalty, not hard feasibility.
\item[\NLDOUpdateLabel{t04}{R}] Energy price doubles for the next control window, so energy cost becomes much more important.
\item[\NLDOUpdateLabel{t05}{D,M}] The workload burst mentioned at the beginning now gets premium priority 5; infer the affected jobs from memory.
\item[\NLDOUpdateLabel{t06}{O}] A carbon-aware operating window starts: machine M06's energy\_idle increases by 1.6, carbon\_intensity rises to 1.80, and energy\_price is set to 1.25. This changes the energy-versus-balance trade-off without changing the machine set or capacities.
\item[\NLDOUpdateLabel{t07}{D}] Existing analytics-like job J020 becomes less urgent: reduce its priority by 1 and extend its soft SLA target by 25 minutes. The active job set remains fixed.
\item[\NLDOUpdateLabel{t08}{R}] A balancing directive follows the scarcity window: energy\_price drops to 0.70 while carbon\_intensity remains elevated at 1.35. The plan should now expose better load balance instead of simply minimizing energy.
\item[\NLDOUpdateLabel{t09}{R}] Operations relaxes the analytics job J020 by extending its soft SLA target by 30 minutes, and carbon\_intensity is set to 1.10.
\item[\NLDOUpdateLabel{t10}{O}] A late reset changes energy and service priorities over the same jobs. Machine M04's energy\_per\_cpu is reduced by 0.04; machine M07's energy\_idle increases by 2.2; energy\_price is set to 0.60 and carbon\_intensity to 0.85. Existing jobs J012, J013, J014, J015, J016 receive priority +1 and soft SLA targets tightened by 20 minutes. The objective trade-off has moved, but current feasibility is still judged over the same fixed public resource and job set.
\item[\NLDOUpdateLabel{t11}{F,O}] A cluster-wide hardware and workload reset activates public regime A. Activate the prelisted late-window jobs, replace every machine's CPU, memory, GPU, availability, idle-energy, and per-CPU energy fields and replace every active job's CPU, memory, GPU requirement, deadline, latency sensitivity, and priority by the supplied rows. Job and machine IDs remain unchanged, and the placement encoding and two objective definitions remain unchanged. Also replace energy\_price and carbon\_intensity by the supplied current values.
\item[\NLDOUpdateLabel{t12}{F,O}] A cluster-wide hardware and workload reset activates public regime B. Activate the prelisted late-window jobs, replace every machine's CPU, memory, GPU, availability, idle-energy, and per-CPU energy fields and replace every active job's CPU, memory, GPU requirement, deadline, latency sensitivity, and priority by the supplied rows. Job and machine IDs remain unchanged, and the placement encoding and two objective definitions remain unchanged. Also replace energy\_price and carbon\_intensity by the supplied current values.
\end{NLDOUpdateList}
\endgroup
\endgroup

\end{document}